\documentclass[final,journal,letterpaper,twocolumn]{IEEEtran}
\usepackage[utf8]{inputenc}
\usepackage{tabularx}
\usepackage{amsmath}
\usepackage{bbm}
\usepackage{amsfonts} 
\usepackage{amssymb}
\usepackage{graphicx}
\usepackage{float}
\usepackage{authblk}
\usepackage{subcaption}
\usepackage{xcolor}
\usepackage{siunitx}
\usepackage{makecell}
\usepackage{multirow}
\usepackage[normalem]{ulem}
\usepackage{booktabs}
\usepackage{mdframed}

\usepackage[colorlinks=true,linkcolor=black,anchorcolor=black,citecolor=black,menucolor=black,runcolor=black,urlcolor=black,bookmarks=true]{hyperref}

\usepackage{tikz}
\usetikzlibrary{mindmap,shadows,backgrounds}
\newcommand*{\info}[4][12]{%
  \node [ annotation, #3, scale=0.65, text width = #1em,
          inner sep = 2mm ] at (#2) {%
  \list{$\bullet$}{\topsep=0pt\itemsep=0pt\parsep=0pt
    \parskip=0pt\labelwidth=8pt\leftmargin=8pt
    \itemindent=0pt\labelsep=2pt}%
    #4
  \endlist
  };
}
\newcommand{\X}{\mathcal{X}}
\newcommand{\Y}{\mathcal{Y}}
\newcommand{\D}{\mathcal{D}}
\newcommand{\W}{\mathcal{W}}

\newcommand{\Dspace}{\mathbb{D}}
\newcommand{\Xspace}{\mathbb{X}}
\newcommand{\Yspace}{\mathbb{Y}}

\newcommand{\R}{\mathbb{R}}

\newcolumntype{L}{>{\arraybackslash}l}

\begin{document}

\title{A Survey of Uncertainty in Deep Neural Networks}

\author{Jakob Gawlikowski (Student Member, IEEE),
        Cedrique Rovile Njieutcheu Tassi,
        Mohsin Ali,
        Jongseok Lee,
        Matthias Humt,
        Jianxiang Feng,
        Anna Kruspe,
        Rudolph Triebel,
        Peter Jung (Member, IEEE),
        Ribana Roscher (Member, IEEE),
        Muhammad Shahzad (Member IEEE),
        Wen Yang (Senior Member, IEEE),
        Richard Bamler (Fellow, IEEE),
        Xiao Xiang Zhu (Fellow, IEEE)
\thanks{This work is in part supported by the German Federal Ministry of Education and Research (BMBF) in the framework of the international future AI lab "AI4EO -- Artificial Intelligence for Earth Observation: Reasoning, Uncertainties, Ethics and Beyond" (Grant number: 01DD20001).}
}


\maketitle

\begin{abstract}
Over the last decade, neural networks have reached almost every field of science and became a crucial part of various real world applications. Due to the increasing spread, confidence in neural network predictions became more and more important. However, basic neural networks do not deliver certainty estimates or suffer from over or under confidence, i.e. are badly calibrated. To overcome this, many researchers have been working on understanding and quantifying uncertainty in a neural network's prediction. As a result, different types and sources of uncertainty have been identified and a variety of approaches to measure and quantify uncertainty in neural networks have been proposed. This work gives a comprehensive overview of uncertainty estimation in neural networks, reviews recent advances in the field, highlights current challenges, and identifies potential research opportunities. It is intended to give anyone interested in uncertainty estimation in neural networks a broad overview and introduction, without presupposing prior knowledge in this field.
For that, a comprehensive introduction to the most crucial sources of uncertainty is given and their separation into reducible model uncertainty and not reducible data uncertainty is presented. The modeling of these uncertainties based on deterministic neural networks, Bayesian neural networks, ensemble of neural networks, and test-time data augmentation approaches is introduced and different branches of these fields as well as the latest developments are discussed. For a practical application, we discuss different measures of uncertainty, approaches for the calibration of neural networks and give an overview of existing baselines and available implementations. Different examples from the wide spectrum of challenges in the fields of medical image analysis, robotic and earth observation give an idea of the needs and challenges regarding uncertainties in practical applications of neural networks. Additionally, the practical limitations of uncertainty quantification methods in neural networks for mission- and safety-critical real world applications are discussed and an outlook on the next steps towards a broader usage of such methods is given. 
\end{abstract}

\begin{IEEEkeywords} Bayesian deep neural networks, Ensembles, Test-time augmentation, Calibration, Uncertainty
\end{IEEEkeywords}

\section{Introduction}
Within the last decade enormous advances on deep neural networks (DNNs) have been realized, encouraging their adaptation in a variety of research fields, where complex systems have to be modeled or understood, as for example earth observation, medical image analysis or robotics. Although DNNs have become attractive in high risk fields such as medical image analysis \cite{nair2020exploring,roy2019bayesian,seebock2019exploiting,labonte2019we,reinhold2020validating,eggenreich2020variational} or autonomous vehicle control \cite{feng2018towards, choi2019gaussian, alex2018spatial, loquercio2020general}, their deployment in mission- and safety-critical real world applications remains limited. The main factors responsible for this limitation are
\begin{itemize}
  \setlength\itemsep{0.5em}
    \item the lack of expressiveness and transparency of a deep neural network's inference model, which makes it difficult to trust their outcomes \cite{roy2019bayesian},
    \item the inability to distinguish between in-domain and out-of-domain samples \cite{lee2017training,mitros2019validity} and the sensitivity to domain shifts \cite{ovadia2019can},
    \item the inability to provide reliable uncertainty estimates for a deep neural network's decision \cite{Ayhan.2018} and frequently occurring overconfident predictions \cite{guo2017calibration,wilson2020bayesian},
    \item the sensitivity to adversarial attacks that make deep neural networks vulnerable for sabotage \cite{rawat2017harnessing,serban2018adversarial,smith2018understanding}.
\end{itemize}
These factors are mainly based on an uncertainty already included in the data (data uncertainty) or a lack of knowledge of the neural network (model uncertainty). To overcome these limitations, it is essential to provide uncertainty estimates, such that uncertain predictions can be ignored or passed to human experts \cite{gal2016dropout}. Providing uncertainty estimates is not only important for a safe decision-making in high-risks fields, but also crucial in fields where the data sources are highly inhomogeneous and labeled data is rare, such as in remote sensing \cite{marcmohsin2020uncertainty,gawlikowski2021out}. Also for fields where uncertainties form a crucial part of the learning techniques, such as for active learning \cite{gal2017deep, chitta2018large, zeng2018relevance, nguyen2019epistemic} or reinforcement learning \cite{gal2016dropout, huang2019bootstrap, kahn2017uncertainty, lotjens2019safe}, uncertainty estimates are highly important.

In recent years, researchers have shown an increased interest in estimating uncertainty in DNNs \cite{blundell2015weight, gal2016dropout, deep.ensembles, prior.network, regularized.evidential.networks, mixture.of.dirichlet, kernel.network, density.estimation.in.representation.space}. The most common way to estimate the uncertainty on a prediction (the predictive uncertainty) is based on separately modelling the uncertainty caused by the model (epistemic or model uncertainty) and the uncertainty caused by the data (aleatoric or data uncertainty). While the first one is reducible by improving the model which is learned by the DNN, the latter one is not reducible. The most important approaches for modeling this separation are Bayesian inference \cite{blundell2015weight, gal2016dropout,  mobiny2019dropconnect, alex2018spatial, krueger2017bayesian}, ensemble approaches \cite{deep.ensembles, sub.ensembles, batch.ensembles}, test time data augmentation approaches \cite{shorten2019survey, wen2020time}, or single deterministic networks containing explicit components to represent the model and the data uncertainty \cite{dirichlet.networks,evidential.neural.networks,distribution.distillation,prior.network,second.opinion.medical}. Estimating the predictive uncertainty is not sufficient for safe decision-making. Furthermore, it is crucial to assure that the uncertainty estimates are reliable. To this end, the calibration property (the degree of reliability) of DNNs has been investigated and re-calibration methods have been proposed \cite{guo2017calibration, wenger2020non, zhang2020mix} to obtain reliable (well-calibrated) uncertainty estimates.

There are several works that give an introduction and overview of uncertainty in statistical modelling. Ghanem et al. \cite{ghanem2017handbook} published a handbook about uncertainty quantification, which includes a detailed and broad description of different concepts of uncertainty quantification, but without explicitly focusing on the application of neural networks. The theses of Gal \cite{gal2016uncertainty} and Kendall \cite{kendall2019geometry} contain a good overview about Bayesian neural networks, especially with focus on the Monte Carlo (MC) Dropout approach and its application in computer vision tasks. The thesis of Malinin \cite{malinin2019uncertainty} also contains a very good introduction and additional insights into Prior networks. Wang et al. contributed two surveys on Bayesian deep learning \cite{wang2016towards, wang2020survey}. They introduced a general framework and the conceptual description of the Bayesian neural networks (BNNs), followed by  an updated presentation of Bayesian approaches for uncertainty quantification in neural networks with special focus on recommender systems, topic models, and control. In \cite{Staahl_UncertaintyReview} an evaluation of uncertainty quantification in deep learning is given by presenting and comparing the uncertainty quantification based on the softmax output, the ensemble of networks, Bayesian neural networks, and autoencoders on the MNIST data set. Regarding the practicability of uncertainty quantification approaches for real life mission- and safety-critical applications, Gustafsson et al. \cite{gustafsson2020evaluating} introduced a framework to test the robustness required in real-world computer vision applications and delivered a comparison of two popular approaches, namely MC Dropout and Ensemble methods.
Hüllermeier et al. \cite{hullermeier2019aleatoric} presented the concepts of aleatoric and epistemic uncertainty in neural networks and discussed different concepts to model and quantify them. In contrast to this, Abdar et al. \cite{abdar2021review} presented an overview on uncertainty quantification methodologies in neural networks and provide an extensive list of references for different application fields and a discussion of open challenges.\\

In this work, we present an extensive overview over all concepts that have to be taken into account when working with uncertainty in neural networks while keeping the applicability on real world applications in mind. Our goal is to provide the reader with a clear thread from the sources of uncertainty to applications, where uncertainty estimations are needed. Furthermore, we point out the limitations of current approaches and discuss further challenges to be tackled in the future. For that, we provide a broad introduction and comparison of different approaches and fundamental concepts. The survey is mainly designed for people already familiar with deep learning concepts and who are planning to incorporate uncertainty estimation into their predictions. But also for people already familiar with the topic, this review provides a useful overview of the whole concept of uncertainty in neural networks and their applications in different fields.\\
In summary, we comprehensively discuss
\begin{itemize}
  \setlength\itemsep{0.5em}
    \item Sources and types of uncertainty (Section \ref{sec:uncertainty_types_and_sources}),
    \item Recent studies and approaches for estimating uncertainty in DNNs (Section \ref{sec:uncertainty_quantification_methods}),
    \item Uncertainty measures and methods for assessing the quality and impact of uncertainty estimates (Section \ref{sec:uncertainty_measures}),
    \item Recent studies and approaches for calibrating DNNs (Section \ref{sec:calibration}),
    \item An overview over frequently used evaluation data sets, available benchmarks and implementations\footnote{The list of available implementations can be found in Section \ref{sec:data_sets_and_baselines} as well as within an additional GitHub repository under \href{https://github.com/JakobCode/UncertaintyInNeuralNetworks\_Resources}{github.com/JakobCode/UncertaintyInNeuralNetworks\_Resources}.} (Section \ref{sec:data_sets_and_baselines}),
    \item An overview over real-world applications using uncertainty estimates (Section \ref{sec:application_fields}),
    \item A discussion on current challenges and further directions of research in the future (Section \ref{sec:con}).
\end{itemize}
In general, the principles and methods for estimating uncertainty and calibrating DNNs can be applied to all regression, classification, and segmentation problems, if not stated differently. In order to get a deeper dive into explicit applications of the methods, we refer to the section on applications and to further readings in the referenced literature. 

\section{Uncertainty in Deep Neural Networks}\label{sec:uncertainty_types_and_sources}
A neural network is a non-linear function $f_\theta$ parameterized by model parameters $\theta$ (i.e. the network weights) that maps from a measurable input set $\Xspace$ to a measurable output set $\Yspace$, i.e.
\begin{equation}\label{eq:inference_model}
f_\theta: \Xspace\rightarrow \Yspace \qquad f_\theta(x)=y~.
\end{equation}
For a supervised setting, we further have a finite set of training data $\D\subseteq\Dspace=\Xspace\times\Yspace$ containing $N$ data samples and corresponding targets, i.e.
\begin{align}
    \D=(\X,\Y)=\{x_n,y_n\}_{n=1}^N\subseteq\Dspace~.
\end{align}
For a new data sample $x^*\in\Xspace$, a neural network trained on $\D$ can be used to predict a corresponding target $f_\theta(x^*) = y^*$.
We consider four different steps from the raw information in the environment to a prediction by a neural network with quantified uncertainties, namely
\begin{enumerate}
    \vspace{0.1em}
    \setlength\itemsep{0.5em}
    \item the \textit{data acquisition} process: \\
    The occurrence of some information in the environment (e.g. a bird's singing) and a measured observation of this information (e.g. an audio record).
    \item the \textit{DNN building} process: \\
    The design and training of a neural network. 
    \item the \textit{applied inference} model:
    The model which is applied for inference (e.g. a Bayesian neural network or an ensemble of neural networks).
    \item the \textit{prediction's uncertainty} model:
    The modelling of the uncertainties caused by the neural network and by the data. 
\end{enumerate} 
In practice, these four steps contain several potential sources of uncertainty and errors, which again affect the final prediction of a neural network. The five factors that we think are the most vital for the cause of uncertainty in a DNN's predictions are
\begin{itemize}
    \vspace{0.1em}
    \setlength\itemsep{0.5em}
    \item the variability in real world situations,
    \item the errors inherent to the measurement systems,
    \item the errors in the architecture specification of the DNN,
    \item the errors in the training procedure of the DNN,
    \item the errors caused by unknown data. \\
\end{itemize}
In the following, the four steps leading from raw information to uncertainty quantification on a DNN's prediction are described in more detail. Within this, we highlight the sources of uncertainty that are related to the single steps and explain how the uncertainties are propagated through the process. Finally, we introduce a model for the uncertainty on a neural network's prediction and introduce the main types of uncertainty considered in neural networks.  \\
The goal of this section is to give an accountable idea of the uncertainties in neural networks. Hence, for the sake of simplicity we only describe and discuss the mathematical properties, which are relevant for understanding the approaches and applying the methodology in different fields.
\subsection{Data Acquisition}
    In the context of supervised learning, the data acquisition describes the process where measurements $x$ and target variables $y$ are generated in order to represent a (real world) situation $\omega$ from some space $\Omega$. In the real world, a realization of $\omega$ could for example be a bird, $x$ a picture of this bird, and $y$ a label stating \textit{'bird'}. During the measurement, random noise can occur and information may get lost. We model this randomness in $x$ by
    \begin{equation}
        x\vert\omega \sim p_{x\vert \omega}~.
    \end{equation}
    Equivalently, the corresponding target variable $y$ is derived, where the description is either based on another measurement or is the result of a labeling process\footnote{In many cases one can model the labeling process as a mapping from $\Xspace$ to $\Yspace$, e.g. for speech recognition or various computer vision tasks. For other tasks, as for example earth observation, this is not always the case. Data is often labeled based on high resolutional data while low resolutional data is utilized for the prediction task.}. For both cases the description can be affected by noise and errors and we state it as
    \begin{equation}
        y\vert\omega \sim p_{y\vert \omega}~.
    \end{equation}
    A neural network is trained on a finite data set of realizations of $x|\omega_i$ and $y|\omega_i$ based on $N$ real world situations $\omega_1,...,\omega_N$,
    \begin{align}
        \D=\{x_i, y_i\}_{i=1}^N~.
    \end{align}
    When collecting the training data, two factors can cause uncertainty in a neural network trained on this data. First, the sample space $\Omega$ should be sufficiently covered by the training data $x_1,...,x_N$ for $\omega_1,...,\omega_N$. For that, one has to take into account that for a new sample $x^*$ it in general holds that $x^*\neq x_i$ for all training situations $x_i$. Following, the target has to be estimated based on the trained neural network model, which directly leads to the first factor of uncertainty:
    \vspace{0.1em}
    \begin{mdframed}
        \textbf{Factor I: Variability in Real World Situations}\\
        Most real world environments are highly variable and almost constantly affected by changes. These changes affect parameters as for example temperature, illumination, clutter, and physical objects' size and shape. Changes in the environment can also affect the expression of objects, as for example plants after rain look very different to plants after a drought. 
        When real world situations change compared to the training set, this is called a distribution shift. Neural networks are sensitive to distribution shifts, which can lead to significant changes in the performance of a neural network.
    \end{mdframed}\vspace{0.8em}
    The second case is based on the measurement system, which has a direct effect on the correlation between the samples and the corresponding targets. The measurement system generates information $x_i$ and $y_i$ that describe $\omega_i$ but might not contain enough information to learn a direct mapping from $x_i$ to $y_i$. This means that there might be highly different real world information $\omega_i$ and $\omega_j$ (e.g. city and forest) resulting in very similar corresponding measurements $x_i$ and $x_j$ (e.g. temperature) or similar corresponding targets $y_i$ and $y_j$ (e.g. label noise that labels both samples as forest). This directly leads to our second factor of uncertainty:
    \vspace{0.8em}
    \begin{mdframed}
    \textbf{Factor II: Error and Noise in Measurement Systems}\\
    The measurements themselves can be a source of uncertainty on the neural network's prediction. This can be caused by limited information in the measurements, as for example the image resolution, or by not measuring false or insufficiently available information modalities. Moreover, it can be caused by noise, for example sensor noise,  by motion, or mechanical stress leading to imprecise measures. Furthermore, false labeling is also a source of uncertainty that can be seen as error and noise in the measurement system. It is referenced as label noise and affects the model by reducing the confidence on the true class prediction during training.
    \end{mdframed}    
\subsection{Deep Neural Network Design and Training}
    The design of a DNN covers the explicit modeling of the neural network and its stochastic training process. The assumptions on the problem structure induced by the design and training of the neural network are called inductive bias \cite{battaglia2018relational}. We summarize all decisions of the modeler on the network's structure (e.g. the number of parameters, the layers, the activation functions, etc.) and training process (e.g. optimization algorithm, regularization, augmentation, etc.) in a structure configuration $s$.
    The defined network structure gives the third factor of uncertainty in a neural network's predictions:
    \vspace{0.8em}
    \begin{mdframed}
        \textbf{Factor III: Errors in the Model Structure}\\
        The structure of a neural network has a direct effect on its performance and therefore also on the uncertainty of its prediction. For instance, the number of parameters affects the memorization capacity, which can lead to under- or over-fitting on the training data. Regarding uncertainty in neural networks, it is known that deeper networks tend to be overconfident in their softmax output, meaning that they predict too much probability on the class with highest probability score \cite{guo2017calibration}.
    \end{mdframed}
    \vspace{0.8em}
    For a given network structure $s$ and a training data set $\D$, the training of a neural network is a stochastic process and therefore the resulting neural network $f_\theta$ is based on a random variable,
    \begin{align}
        \theta\vert D, s \sim p_{\theta|D,s}.
    \end{align}
    The process is stochastic due to random decisions as the order of the data, random initialization or random regularization as augmentation or dropout. The loss landscape of a neural network is highly non-linear and the randomness in the training process in general leads to different local optima $\theta^*$ resulting in different models \cite{deep.ensembles}. Also, parameters as batch size, learning rate, and the number of training epochs affect the training and result in different models. Depending on the underlying task these models can significantly differ in their predictions for single samples, even leading to a difference in the overall model performance. This sensitivity to the training process directly leads to the fourth factor for uncertainties in neural network predictions: 
    \vspace{0.8em}
    \begin{mdframed}
    \textbf{Factor IV: Errors in the Training Procedure} \\
    The training process of a neural network includes many parameters that have to be defined (batch size, optimizer, learning rate, stopping criteria, regularization, etc.) and also stochastic decisions within the training process (batch generation and weight initialization) take place. All these decisions affect the local optima and it is therefore very unlikely that two training processes deliver the same model parameterization. A training data set that suffers from imbalance or low coverage of single regions in the data distribution also introduces uncertainties on the network's learned parameters, as already described in the data acquisition. This might be softened by applying augmentation to increase the variety or by balancing the impact of single classes or regions on the loss function.
    \end{mdframed}
    Since the training process is based on the given training data set $\D$, errors in the data acquisition process (e.g. label noise) can result in errors in the  training  process.
\subsection{Inference}
    The inference describes the prediction of an output $y^*$ for a new data sample $x^*$ by the neural network. At this time, the network is trained for a specific task. Thus, samples which are not inputs for this task cause errors and are therefore also a source of uncertainty:  
    \vspace{0.8em}
    \begin{mdframed}
        \textbf{Factor V: Errors Caused by Unknown Data}\\
        Especially in classification tasks, a neural network that is trained on samples derived from a world $\W_1$ can also be capable of processing samples derived from a completely different world $\W_2$. This is for example the case, when a network trained on images of cats and dogs receives a sample showing a bird. Here, the source of uncertainty does not lie in the data acquisition process, since we assume a world to contain only feasible inputs for a prediction task. Even though the practical result might be equal to too much noise on a sensor or complete failure of a sensor, the data considered here represents a valid sample, but for a different task or domain.
    \end{mdframed}
\subsection{Predictive Uncertainty Model}
As a modeller, one is mainly interested in the uncertainty that is propagated onto a prediction $y^*$, the so-called \textit{predictive uncertainty}. Within the data acquisition model, the probability distribution for a prediction $y^*$ based on some sample $x^*$ is given by
\begin{equation}\label{eq:real_world_distribution}
    p(y^*|x^*) = \int_\Omega p(y^*|\omega)p(\omega|x^*)d\omega
\end{equation}
and a maximum a posteriori (MAP) estimation is given by
\begin{equation}\label{eq:real_world_max_likelihood}
    y^* = \arg \max_y p(y | x^*)~.
\end{equation}
Since the modeling is based on the unavailable latent variable $\omega$, one takes an approximative representation based on a sampled training data set $\mathcal{D}=\{x_i,y_i\}_{i=1}^N$ containing $N$ samples and corresponding targets. The distribution and MAP estimator in \eqref{eq:real_world_distribution} and \eqref{eq:real_world_max_likelihood} for a new sample $x^*$ are then predicted based on the known examples by
\begin{equation}\label{eq:data_set_distribution}
    p(y^*\vert x^*) = \int_D p(y^*\vert \D,x^*)
\end{equation}
and
\begin{equation}\label{eq:data_set_max_likelihood}
    y^* = \arg \max_y p(y | \D,x^*)~.
\end{equation}
In general, the distribution given in \eqref{eq:data_set_distribution} is unknown and can only be estimated based on the given data in $D$. For this estimation, neural networks form a very powerful tool for many tasks and applications. 
\par
The prediction of a neural network is subject to both model-dependent and input data-dependent errors, and therefore the predictive uncertainty associated with $y^*$ is in general separated into \textit{data uncertainty} (also statistical or aleatoric uncertainty \cite{hullermeier2019aleatoric}) and \textit{model uncertainty} (also systemic or epistemic uncertainty \cite{hullermeier2019aleatoric}). Depending on the underlying approach, an additional explicit modeling of \textit{distributional uncertainty} \cite{prior.network} is used to model the uncertainty, which is caused by examples from a region not covered by the training data.
\subsubsection{Model- and Data Uncertainty}
The model uncertainty covers the uncertainty that is caused by shortcomings in the model, either by errors in the training procedure, an insufficient model structure, or lack of knowledge due to unknown samples or a bad coverage of the training data set. 

In contrast to this, data uncertainty is related to uncertainty that directly stems from the data. Data uncertainty is caused by information loss when representing the real world within a data sample and represents the distribution stated in \eqref{eq:real_world_distribution}. For example, in regression tasks noise in the input and target measurements causes data uncertainty that the network cannot learn to correct. In classification tasks, samples which do not contain enough information in order to identify one class with 100\% certainty cause data uncertainty on the prediction. The information loss is a result of the measurement system, e.g. by representing real world information by image pixels with a specific resolution, or by errors in the labelling process.  \\
Considering the five presented factors for uncertainties on a neural network's prediction, model uncertainty covers Factors I, III, IV, and V and data uncertainty is related to Factor II. While model uncertainty can be (theoretically) reduced by improving the architecture, the learning process, or the training data set, the data uncertainties cannot be explained away \cite{kendall2017uncertainties}. Therefore, DNNs that are capable of handling uncertain inputs and that are able to remove or quantify the model uncertainty and give a correct prediction of the data uncertainty are of paramount importance for a variety of real world mission- and safety-critical applications.
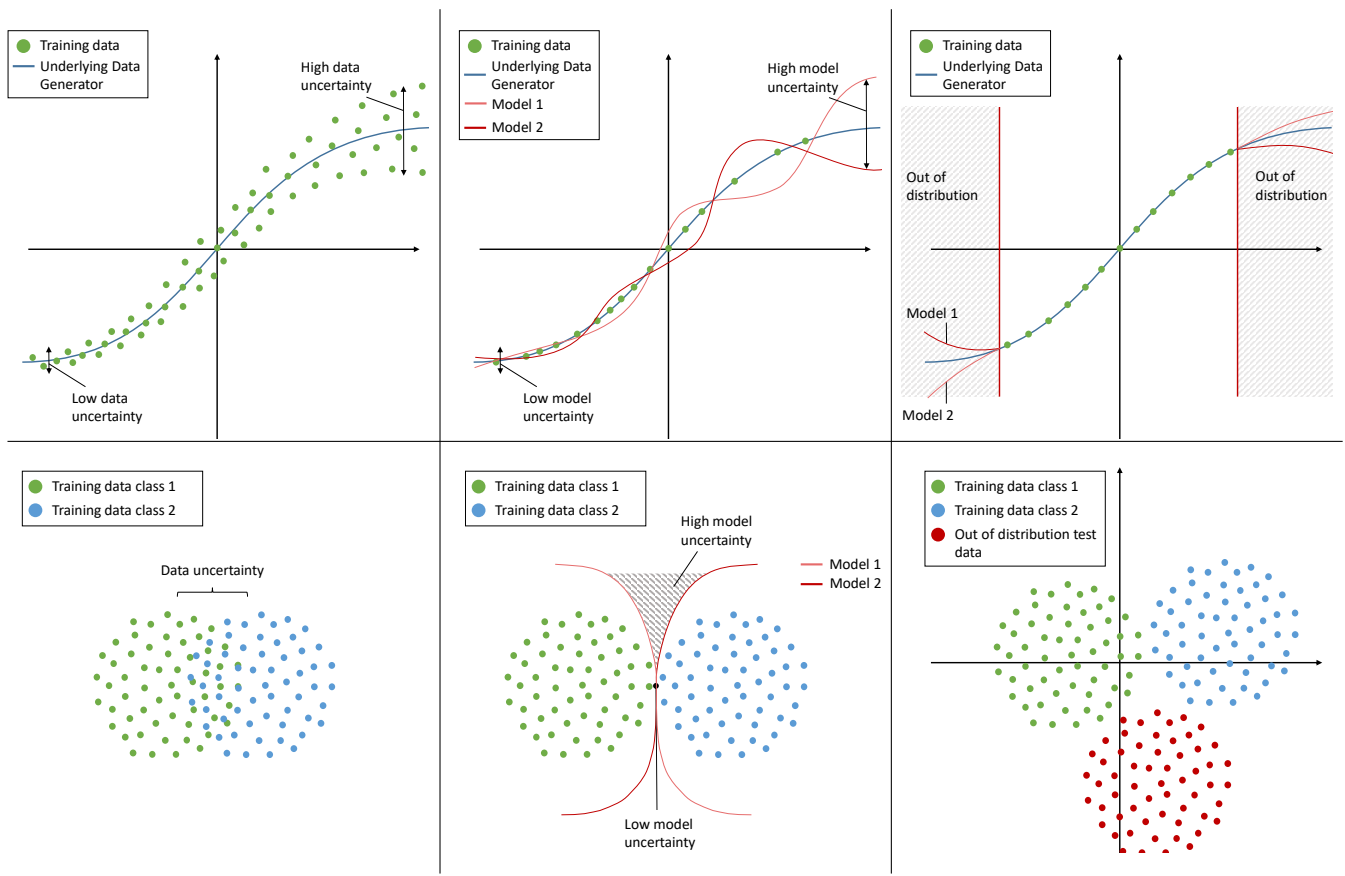
\begin{figure*}
	\centering
	\begin{tikzpicture}
		\node[inner sep=0pt] at (0,0)
		{\includegraphics[clip, trim=7.5cm 2.8cm 10.5cm 1.2cm, width=.316\textwidth]{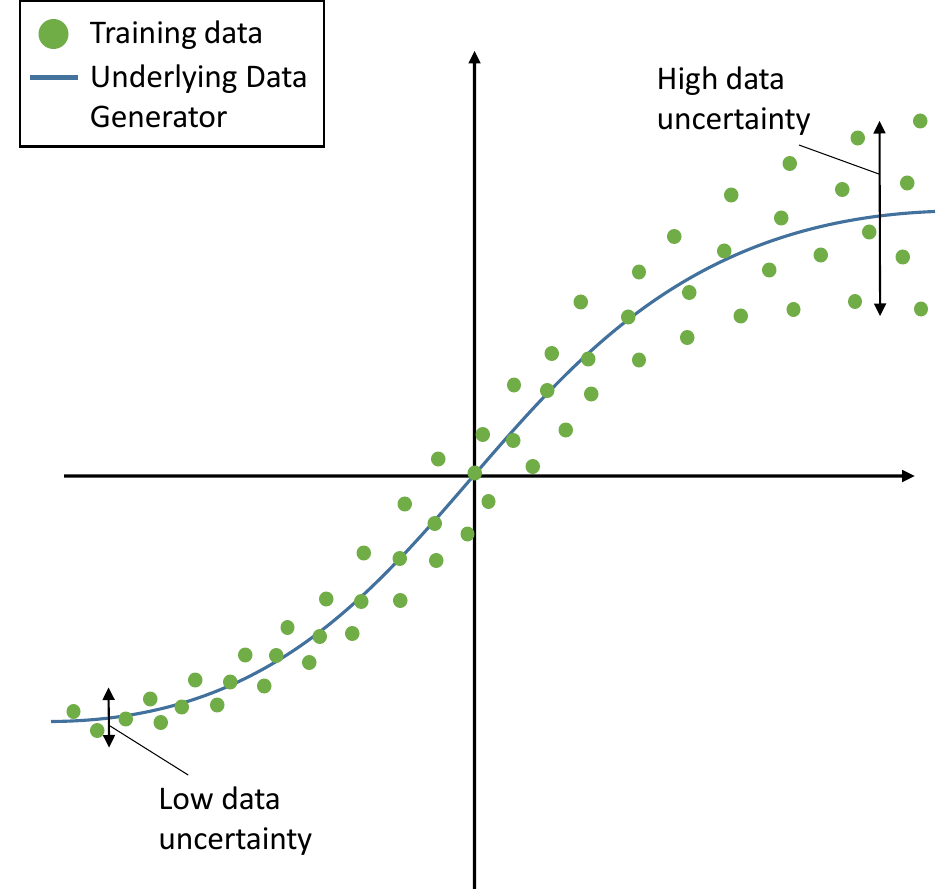}};
		\node[inner sep=0pt] at (6,0)
		{\includegraphics[clip, trim=7.5cm 2.8cm 10.5cm 1.2cm, width=.316\textwidth]{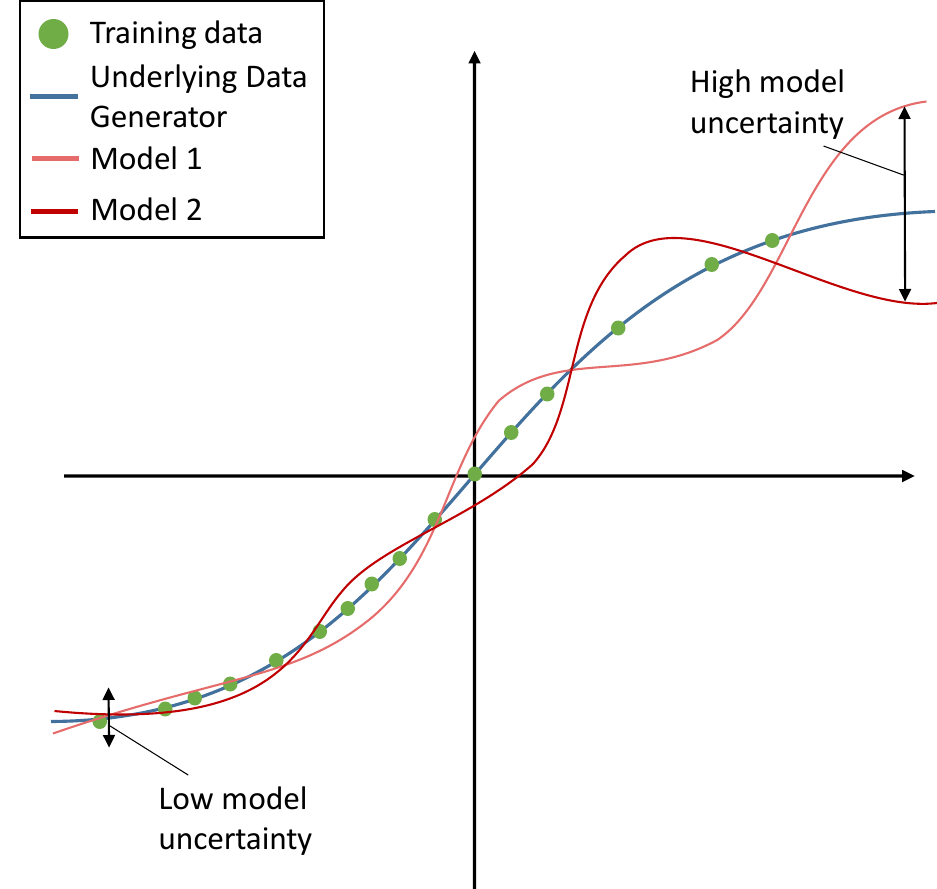}};
		\node[inner sep=0pt] at (12,0)
		{\includegraphics[clip, trim=7.5cm 2.8cm 10.5cm 1.2cm, width=.316\textwidth]{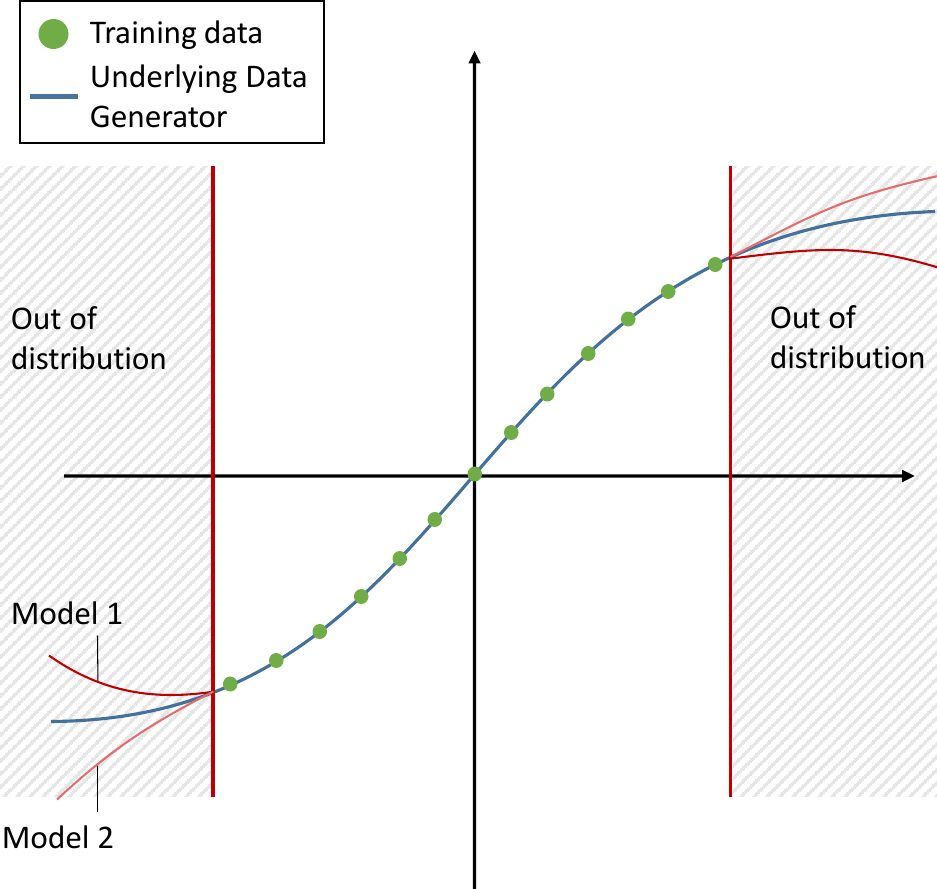}};
		\node[inner sep=0pt]  at (0,-5.5)
		{\includegraphics[clip, trim=7.5cm 2.8cm 10.5cm 1.2cm, width=.316\textwidth]{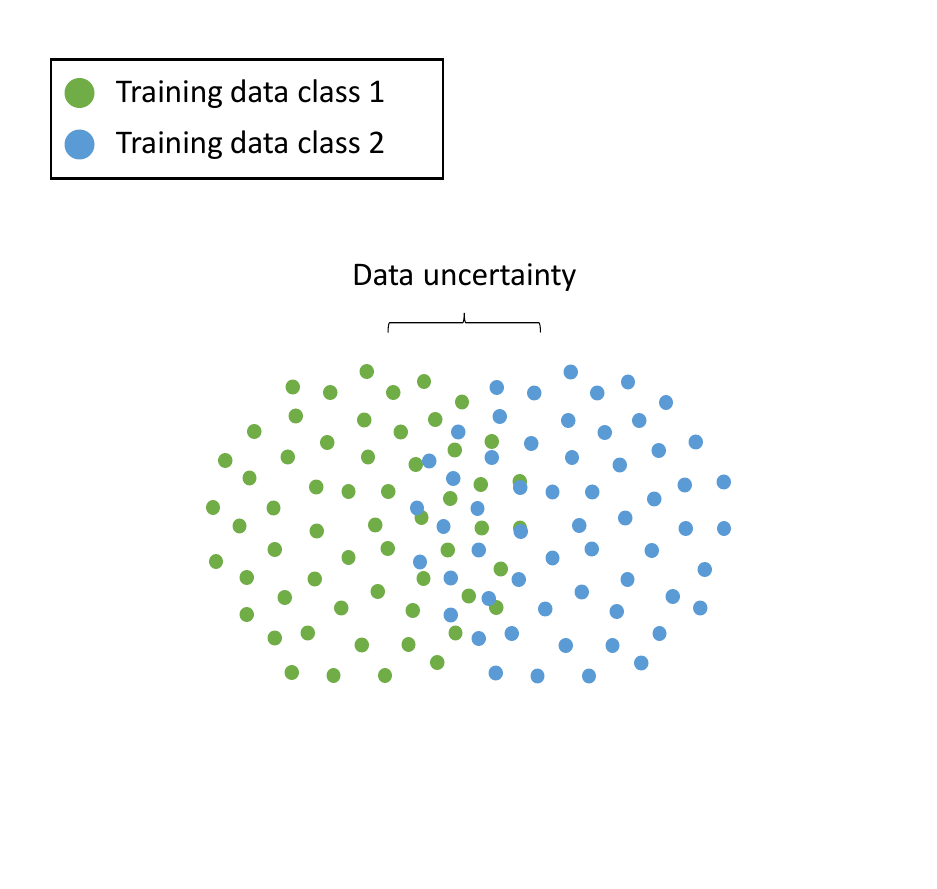}};
		\node[inner sep=0pt] at (6,-5.5)
		{\includegraphics[clip, trim=7.8cm 2.8cm 10.2cm 1.2cm, width=.316\textwidth]{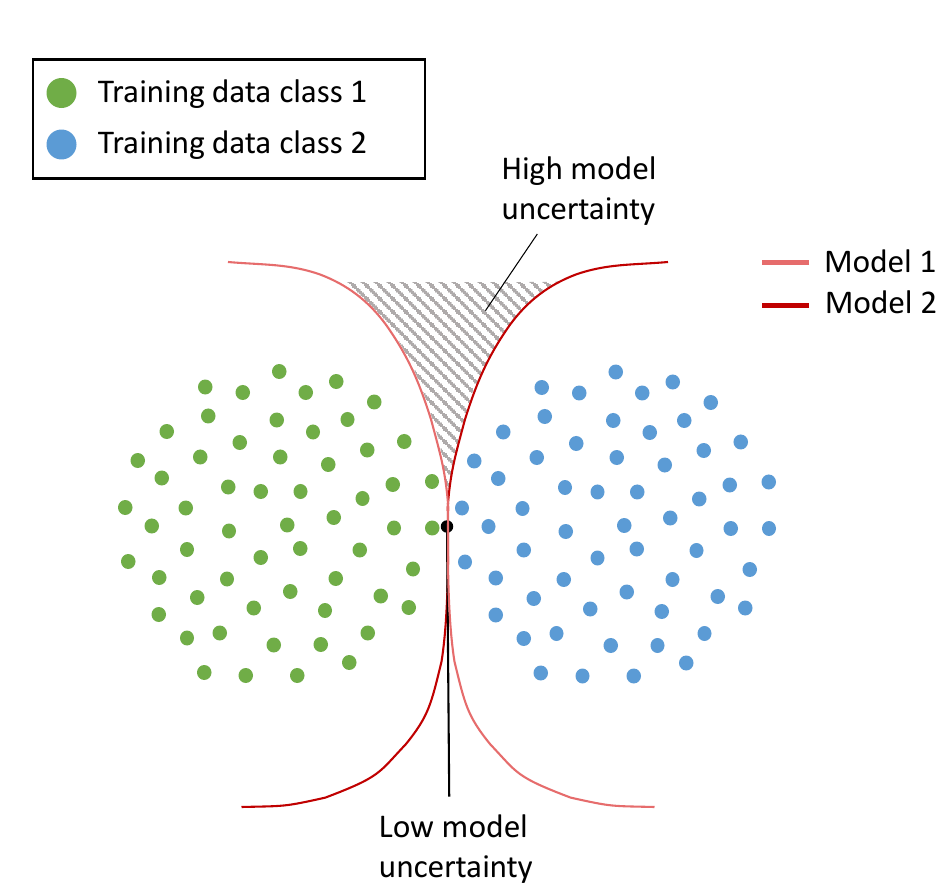}};
		\node[inner sep=0pt] at (12,-5.5)
		{\includegraphics[clip, trim=7.5cm 2.8cm 10.5cm 1.2cm, width=.316\textwidth]{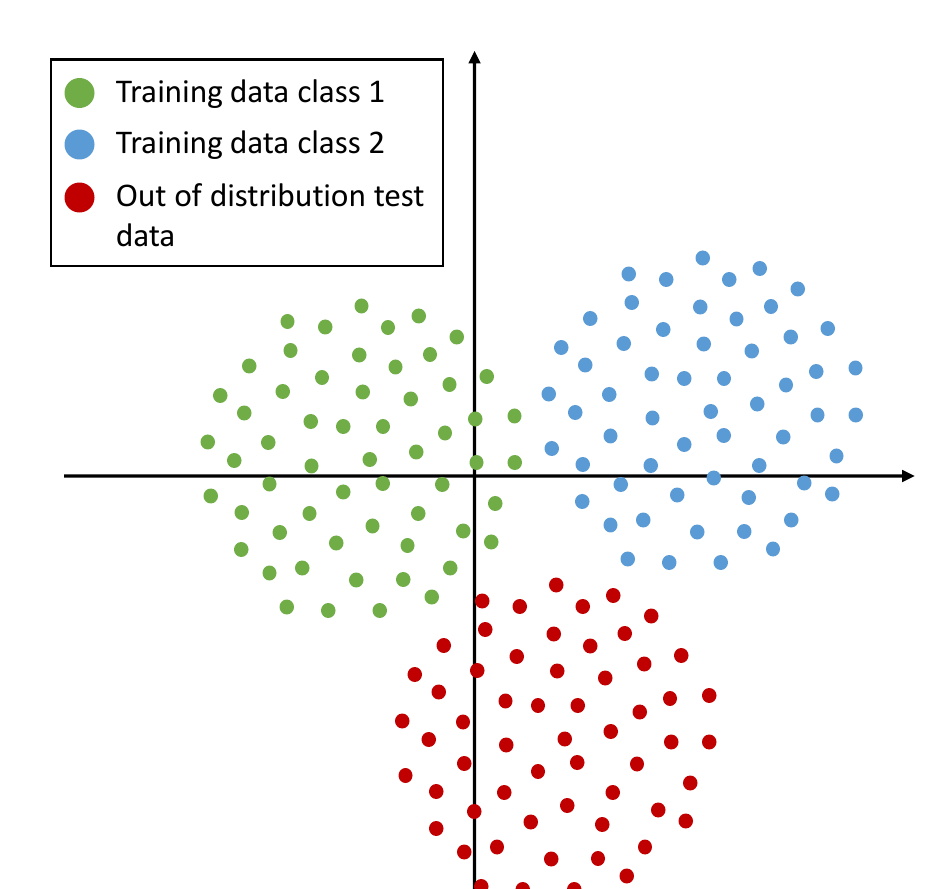}};
		\draw (-2.75,-2.75) -- (15,-2.75);
		\draw (3,3) -- (3,-8.5);
		\draw (9,3) -- (9,-8.5);
	\end{tikzpicture}
	\caption{Visualization of the data, the model, and the distributional uncertainty for classification and regression models.}
	\label{fig:uncertainty_types}
\end{figure*}

The Bayesian framework offers a practical tool to reason about uncertainty in deep learning \cite{gal2015bayesian}. In Bayesian modeling, the model uncertainty is formalized as a probability distribution over
the model parameters $\theta$, while the data uncertainty is formalized as a probability distribution over the model outputs $y^*$, given a parameterized model $f_\theta$. The distribution over a prediction $y^*$, the predictive distribution, is then given by 
\begin{align}\label{eq:predictive_uncertainty_intro} 
    	    p(y^*|x^*, D)&=\int\underbrace{p(y^*\vert x^*,\theta)}_{\text{Data}}\underbrace{p(\theta\vert D)}_{\text{Model}}d\theta~.
\end{align}
The term $p(\theta | D)$ is referenced as posterior distribution on the model parameters and describes the uncertainty on the model parameters given a training data set $D$. The posterior distribution is in general not tractable. While ensemble approaches seek to approximate it by learning several different parameter settings and averaging over the resulting models \cite{deep.ensembles}, Bayesian inference reformulates it using Bayes Theorem \cite{bishop2006pattern} 
\begin{equation}\label{eq:bayes_posterior_intro}
    p(\theta|D) = \frac{p(D|\theta)p(\theta)}{p(D)}~.
\end{equation}
The term $p(\theta)$ is called the prior distribution on the model parameters, since it does not take any information but the general knowledge on $\theta$ into account. The term $p(D\vert\theta)$ represents the likelihood that the data in $D$ is a realization of the distribution predicted by a model parameterized with $\theta$. Many loss functions are motivated by or can be related to the likelihood function. Loss functions that seek to maximize the log-likelihood (for an assumed distribution) are for example the cross-entropy or the mean squared error \cite{ritter2018scalable}. 

Even with the reformulation given in \eqref{eq:bayes_posterior_intro}, the predictive distribution given in \eqref{eq:predictive_uncertainty_intro} is still intractable. To overcome this, several different ways to approximate the predictive distribution were proposed. A broad overview on the different concepts and some specific approaches is presented in Section \ref{sec:uncertainty_quantification_methods}.
\subsubsection{Distributional Uncertainty}
Depending on the approaches that are used to quantify the uncertainty in $y^*$, the formulation of the predictive distribution might be further separated into data, distributional, and model parts \cite{prior.network}:
\begin{equation}\label{eq:predictive_uncertainty_split_2_intro}
p(y^*|x^*, D)=\int\int \underbrace{p(y\vert \mu)}_{\text{Data}}\underbrace{p(\mu\vert x^*,\theta)}_{\text{Distributional}}\underbrace{p(\theta\vert D)}_{\text{Model}}d\mu d\theta~.
\end{equation}
The distributional part in \eqref{eq:predictive_uncertainty_split_2_intro} represents the uncertainty on the actual network output, e.g. for classification tasks this might be a Dirichlet distribution, which is a distribution over the categorical distribution given by the softmax output.
Modeled this way, distributional uncertainty refers to uncertainty that is caused by a change in the input-data distribution, while model uncertainty refers to uncertainty that is caused by the process of building and training the DNN. As modeled in \eqref{eq:predictive_uncertainty_split_2_intro}, the model uncertainty affects the estimation of the distributional uncertainty, which affects the estimation of the data uncertainty.

While most methods presented in this paper only distinguish between model and data uncertainty, approaches specialized on out-of-distribution detection often explicitly aim at representing the distributional uncertainty \cite{prior.network,nandy2020towards}. A more detailed presentation of different approaches for quantifying uncertainties in neural networks is given in Section \ref{sec:uncertainty_quantification_methods}. In Section \ref{sec:uncertainty_measures}, different measures for measuring the different types of uncertainty are presented.

\subsection{Uncertainty Classification}
On the basis of the input data domain, the predictive uncertainty can also be classified into three main classes: 
\begin{itemize}
  \setlength\itemsep{0.5em}
    \item \textit{In-domain uncertainty} \cite{ashukha2020pitfalls}\\
    In-domain uncertainty represents the uncertainty related to an input drawn from a data distribution assumed to be equal to the training data distribution. The in-domain uncertainty stems from the inability of the deep neural network to explain an in-domain sample due to lack of in-domain knowledge. From a modeler point of view, in-domain uncertainty is caused by design errors (model uncertainty) and the complexity of the problem at hand (data uncertainty). Depending on the source of the in-domain uncertainty, it might be reduced by increasing the quality of the training data (set) or the training process \cite{hullermeier2019aleatoric}.   
    \item \textit{Domain-shift uncertainty} \cite{ovadia2019can} \\
    Domain-shift uncertainty denotes the uncertainty related to an input drawn from a shifted version of the training distribution. The distribution shift results from insufficient coverage by the training data and the variability inherent to real world situations. A domain-shift might increase the uncertainty due to the inability of the DNN to explain the domain shift sample on the basis of the seen samples at training time. Some errors causing domain shift uncertainty can be modeled and can therefore be reduced. For example, occluded samples can be learned by the deep neural network to reduce domain shift uncertainty caused by occlusions \cite{devries2017improved}. However, it is difficult if not impossible to model all errors causing domain shift uncertainty, e.g., motion noise \cite{kendall2017uncertainties}. From a modeler point of view, domain-shift uncertainty is caused by external or environmental factors but can be reduced by covering the shifted domain in the training data set. 
    \item \textit{Out-of-domain uncertainty} \cite{hendrycks2016baseline,liang2017enhancing,shafaei2018less,mundt2019open}\\
    Out-of-domain uncertainty represents the uncertainty related to an input drawn from the subspace of unknown data. The distribution of unknown data is different and far from the training distribution. While a DNN can extract in-domain knowledge from domain-shift samples, it cannot extract in-domain knowledge from out-of-domain samples. For example, when domain-shift uncertainty describes phenomena like a blurred picture of a dog, out-of-domain uncertainty describes the case when a network that learned to classify cats and dogs is asked to predict a bird.
    The out-of-domain uncertainty stems from the inability of the DNN to explain an out-of-domain sample due to its lack of out-of-domain knowledge. From a modeler point of view, out-of-domain uncertainty is caused by input samples, where the network is not meant to give a prediction for or by insufficient training data.
\end{itemize}
Since the model uncertainty captures what the DNN does not know due to lack of in-domain or out-of-domain knowledge, it captures all, in-domain, domain-shift, and out-of-domain uncertainties. In contrast, the data uncertainty captures in-domain uncertainty that is caused by the nature of the data the network is trained on, as for example overlapping samples and systematic label noise. 

\begin{figure*}[t]
\resizebox{\textwidth}{!}{
   \includegraphics{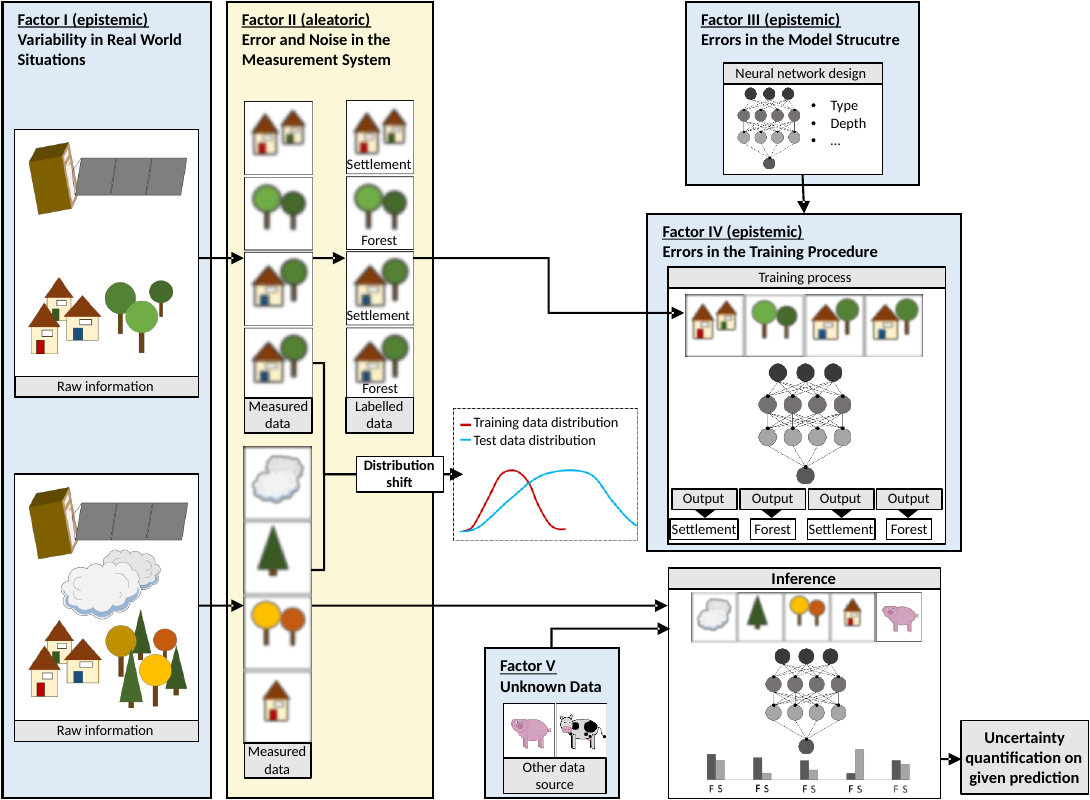}}
    \caption{The illustration shows the different steps of a neural network pipeline, based on the earth observation example of land cover classification (here settlement and forest) based on optical images. The different factors that affect the predictive uncertainty are highlighted in the boxes. Factor I is shown as changing environments by cloud covered trees, different types and colors of trees. Factor II is shown by insufficient measurements, that can not directly be used to separate between settlement and forest and by label noise. In practice, the resolution of such images can be low and which would also be part of Factor II. Factor III and Factor IV represent the uncertainties caused by the network structure and the stochastic training process, respectively. Factor V in contrast is represented by feeding the trained network with unknown types of images, namely cows and pigs.} 
\label{fig:uncertainty_sources}
\end{figure*}

\section{Uncertainty Estimation}\label{sec:uncertainty_quantification_methods}

As described in Section \ref{sec:uncertainty_types_and_sources}, several factors may cause model and data uncertainty and affect a DNN's prediction. This variety of sources of uncertainty makes the complete exclusion of uncertainties in a neural network impossible for almost all applications. Especially in practical applications employing real world data, the training data is only a subset of all possible input data, which means that a miss-match between the DNN domain and the unknown actual data domain is often unavoidable. However, an exact representation of the uncertainty of a DNN prediction is also not possible to compute, since the different uncertainties can in general not be modeled accurately and are most often even unknown.\\
Therefore, methods for estimating uncertainty in a DNN prediction is a popular and vital field of research. The data uncertainty part is normally represented in the prediction, e.g. in the softmax output of a classification network or in the explicit prediction of a standard deviation in a regression network \cite{kendall2017uncertainties}. In contrast to this, several different approaches which model the model uncertainty and seek to separate it from the data uncertainty in order to receive an accurate representation of the data uncertainty were introduced \cite{kendall2017uncertainties, prior.network, deep.ensembles}. \\
In general, the methods for estimating the uncertainty can be split in four different types based on the number (single or multiple) and the nature (deterministic or stochastic) of the used DNNs. 
\begin{itemize}
  \setlength\itemsep{0.5em}
    \item \textbf{Single deterministic methods} give the prediction based on one single forward pass within a deterministic network. The uncertainty quantification is either derived by using additional (external) methods or is directly predicted by the network. 
    \item \textbf{Bayesian methods} cover all kinds of stochastic DNNs, i.e. DNNs where two forward passes of the same sample generally lead to different results. 
    \item \textbf{Ensemble methods} combine the predictions of several different deterministic networks at inference. 
    \item \textbf{Test-time augmentation methods} give the prediction based on one single deterministic network but augment the input data at test-time in order to generate several predictions that are used to evaluate the certainty of the prediction.
\end{itemize}
\begin{figure*}[ht]
\centering
\caption{Visualization of the four different types of uncertainty quantification methods presented in this paper.}
\resizebox{\textwidth}{!}{%
\begin{tikzpicture}[framed, every annotation/.style = {draw,
                     fill = white, font = \large}]
  \path[mindmap,concept color=black!40,text=black,
    every node/.style={concept,circular drop shadow},
    root/.style    = {concept color=black!40,
      font=\Large\bfseries,text width=10em},
    level 1 concept/.append style={font=\normalsize\bfseries, clockwise from=0,
      sibling angle=60,text width=7em,
    level distance=15em,inner sep=0pt},
    level 2 concept/.append style={font=\small\bfseries,level distance=9em},
  ]
  
  node[root] {\hyperref[sec:uncertainty_quantification_methods]Neural Network Uncertainty Quantification Methods} [clockwise from=0]
    child[concept color=yellow!60!black] {
      node[concept](test_time_augmentation) {\hyperref[sec:test_time_augmentation]{Test-Time Augmentation Methods}}
    }
    child[concept color=orange] {
      node[concept] {\hyperref[sec:ensemble.methods]{Ensemble Methods}}
        [clockwise from=30]
      child { node[concept] (Weight_Sharing)
        {Weight Sharing}}
      child { node[concept] (Reduce_Member)
        {Reduce Members} }
      child { node[concept] (Training_Strategies)
        {Training Strategies}}
    }
    child[concept color=blue!60] {
      node {\hyperref[sec:bayesian_methods]{Bayesian Methods}} [clockwise from=300]
        child { node (laplace_approx){Laplace Approximation} }
        child { node (sampling_methods){Sampling Methods} }
        child { node (var_inference) {Variational Inference} }
    }
    child[concept color=green!40!black] {
      node[concept] {\hyperref[sec:deterministic_methods]{Single Network Deterministic Methods}}
        [clockwise from=200]
      child { node[concept] (internal) 
        {Internal Methods}}
      child { node[concept] (external)
        {External Methods}}
    };
    
    \info{test_time_augmentation.north east}{above,anchor=west}{%
      \item Augmentation Policies\textsuperscript{25}
    }    
    
    \info{Weight_Sharing.south east}{above,anchor=west}{%
      \item Sub-Ensembles\textsuperscript{24} 
      \item Batch-Ensembles\textsuperscript{23} 
    }
    \info{Reduce_Member.south east}{above,anchor=west}{%
      \item Model Pruning\textsuperscript{21}
      \item Distillation\textsuperscript{22}
    }
    \info{Training_Strategies.south east}{above,anchor=north}{%
      \item Random Initialization/ Data Shuffling\textsuperscript{18}
      \item Bagging/ Boosting\textsuperscript{19} 
      \item Single Training Run\textsuperscript{20} 
    }
    
    \info{laplace_approx.south east}{above,anchor=north}{%
      \item Diagonal Information Matrix\textsuperscript{15}
      \item Kronecker-Factorization\textsuperscript{16}
      \item Sparse Information Matrix\textsuperscript{17}
    }
    \info{sampling_methods.south west}{above,anchor=north}{%
      \item Original works\textsuperscript{12}
      \item Stochastic MCMC\textsuperscript{13}
      \item Theoretic Advances\textsuperscript{14}
    }
    \info{[xshift=-3pt,yshift=-5pt]var_inference.south west}{above,anchor=east}{%
      \item Application of Variational Inference\textsuperscript{8}
      \item Stochastic Variational Inference\textsuperscript{9}
      \item Normalizing flows\textsuperscript{10}
      \item Monte-Carlo Dropout\textsuperscript{11}
    }
    
    \info{[yshift=3pt]internal.south east}{above,anchor=north}{%
      \item Prior Networks\textsuperscript{4} 
      \item Evidential Neural Networks\textsuperscript{5} 
      \item Gradient penalties\textsuperscript{7} 
    }
    \info{external.north east}{above,anchor=west}{%
      \item Gradient Metrices\textsuperscript{1} 
      \item Additional Network for Uncertainty\textsuperscript{2} 
      \item Distance to Training Data\textsuperscript{3}
    }

    \node[draw,text width=18cm, align=left] at (0,-12.5){
    \textsuperscript{1}~\cite{oberdiek2018classification,lee2020gradients}
    ~~~\textsuperscript{2}~\cite{second.opinion.medical}
    ~~~\textsuperscript{3}~\cite{density.estimation.in.representation.space,kernel.network}
    ~~~\textsuperscript{4}~\cite{prior.network}
    ~~~\textsuperscript{5}~\cite{evidential.neural.networks}
    ~~~\textsuperscript{7}~\cite{kernel.network}
    \textsuperscript{8}~\cite{hinton1993keeping,barber1998ensemble},
    ~~~\textsuperscript{9}~\cite{graves2011practical,blundell2015weight}
    ~~~\textsuperscript{10}~\cite{louizos2017bayesian,rezende2015variational}
    ~~~\textsuperscript{11}~\cite{gal2016dropout}
    ~~~\textsuperscript{12}~\cite{neal1992bayesian,neal1994improved,neal1995bayesian}
    ~~~\textsuperscript{13}~\cite{welling2011bayesian}
    ~~~\textsuperscript{14}~\cite{nemeth2019stochastic}
    ~~~\textsuperscript{15}~\cite{Salimans2016}
    ~~~\textsuperscript{16}~\cite{ritter2018scalable}
    ~~~\textsuperscript{17}~\cite{lee2020estimating}
    \textsuperscript{18}~\cite{deep.ensembles}
    ~~~\textsuperscript{19} \cite{achrack2020multi}
    ~~~\textsuperscript{20}~\cite{huang2017snapshot}
    ~~~\textsuperscript{21}~\cite{cavalcanti2016combining,guo2018margin,martinez2019ensemble}
    ~~~\textsuperscript{22} \cite{distribution.distillation.general.framework,distribution.distillation}
    ~~~\textsuperscript{23}~\cite{sub.ensembles}
    ~~~\textsuperscript{24}~\cite{batch.ensembles} 
    \textsuperscript{25}~\cite{molchanov2020greedy}
    };
\end{tikzpicture}
}
\label{fig:methods_diagram}
\end{figure*}
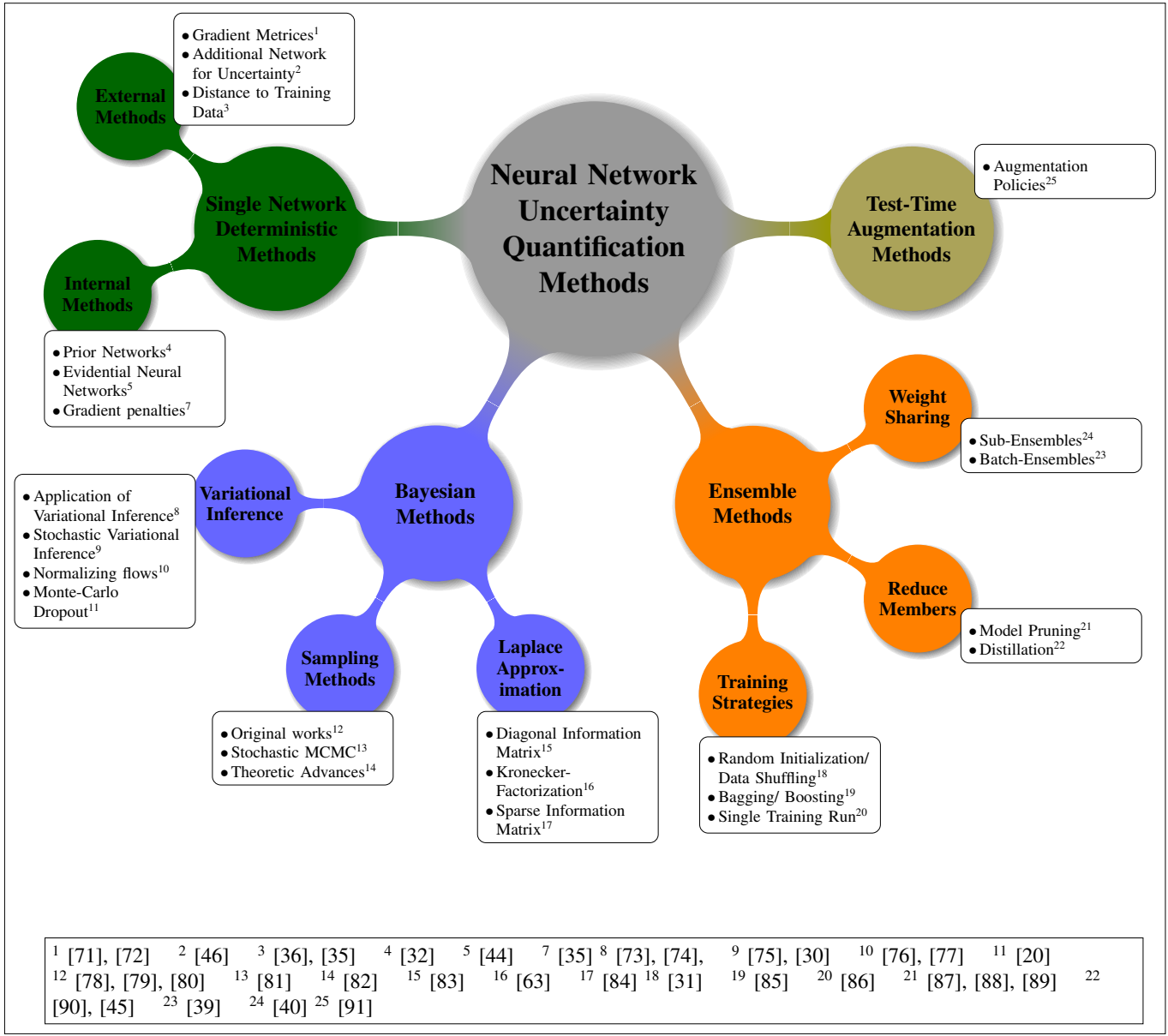
\begin{table*}[t]
\centering
\caption{An overview about the four general methods presented in this paper, namely Bayesian Neural Networks, Ensembles, Single Deterministic Neural Networks, and Test-Time Data Augmentation. The labels \text{high} and \text{low} are given relative to the other approaches and based on the general idea behind them.}
\begin{tabular}{>{\raggedright\arraybackslash}p{3cm}p{3.2cm}p{3.2cm}p{3.2cm}p{3.2cm}}


&\centering\textbf{Single Deterministic Networks} 
&\centering\textbf{Bayesian Methods}
&\centering\textbf{Ensemble  Methods} 
&\multicolumn{1}{>{\centering\arraybackslash}p{3.2cm}}{\textbf{Test-Time Data Augmentation}}
\\ \hline

\addlinespace[2ex]
\textbf{Description}
&Approaches that receive an uncertainty quantification on a prediction of a deterministic neural network. 
&Model parameters are explicitly modeled as random variables. For a single forward pass the parameters are sampled from this distribution. Therefore, the prediction is stochastic and each prediction is based on different model weights.  
&The predictions of several models are combined into one prediction. A variety among the single models is crucial. 
&The prediction and uncertainty quantification at inference is based on several predictions resulting from different augmentations of the original input sample.   \\ 

\addlinespace[2ex]
\textbf{Description of Model Uncertainties}
&No
&Yes
&No
&No
\\ 

\addlinespace[2ex]
\textbf{Need changes on existing networks} 
&Depends on method 
&Yes 
&Yes (retrain several times) 
&No 
\\ 

\addlinespace[2ex]
\textbf{Sensitivity to initialization and parameters of training process} 
&High (in general) 
&Low  \newline (Usage of uninformative priors possible) 
&Low 
&Low  
\\ 


\addlinespace[2ex]
\textbf{Number of networks trained} 
&1 
&1 
&Several 
&1 
\\ 

\addlinespace[2ex]
\textbf{Computational effort during training} 
&Low 
&High 
&High 
&Low 
\\ 

\addlinespace[2ex]
\textbf{Memory consumption during training} 
&Low 
&Low 
&High 
&Low 
\\ 

\addlinespace[2ex]
\textbf{Number of inputs per prediction} 
&1 
&1 
&1 
&Several 
\\ 

\addlinespace[2ex]
\textbf{Forward passes per prediction} 
&1
&Several 
&Several 
&Several 
\\

\addlinespace[2ex]
\textbf{Evaluated modes} 
&Single 
&Single 
&Multiple 
&Single 
\\

\addlinespace[2ex]
\textbf{Computational effort during inference} 
&Low \newline
(One forward pass, possibly some minor additional effort for uncertainty quantification)
&High\newline
(sampling is either needed for explicit approach or for the approximation of intractable formulas)
&High \newline
(Several models need  to be evaluated) 
&High \newline (Several augmentations and forward passes are performed)
\\
 
\addlinespace[2ex]
\textbf{Memory Consumption - Inference} 
&Low 
&Low 
&High 
&Low \\ \hline

\end{tabular}
\label{tab:types_compare}
\end{table*}
\begin{figure*}[t]
\resizebox{\textwidth}{!}{
   \includegraphics{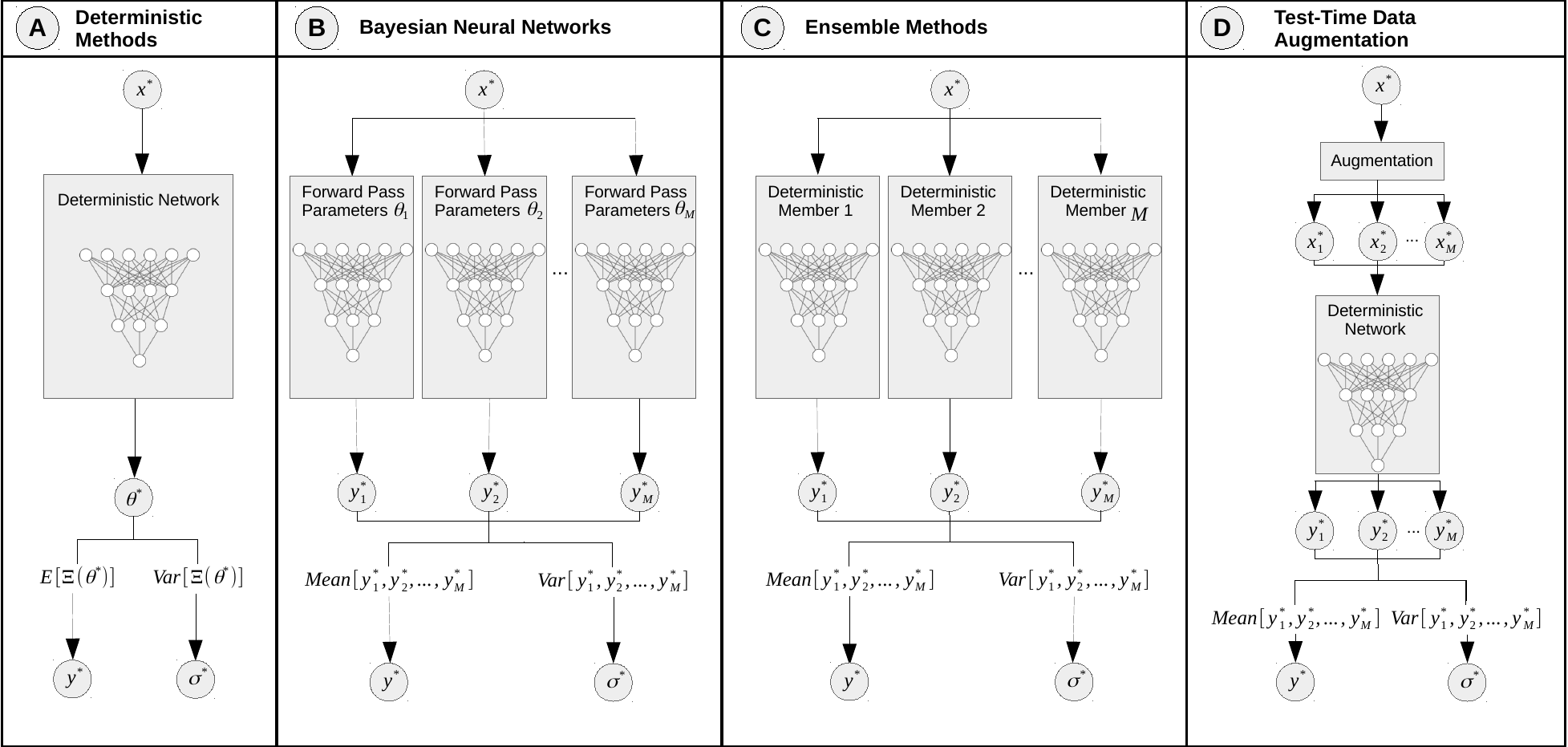}}
    \caption{A visualization of the basic principles of uncertainty modeling of the four presented general types of uncertainty prediction in neural networks. For a given input sample $x^*$ each approach delivers a prediction $y^*$, a representation of model uncertainty $\sigma_{\text{model}}$ and a value of data uncertainty $\sigma_{\text{data}}$. \textbf{A) single deterministic model}, \textbf{B) Bayesian neural network}, \textbf{B) ensemble approach}, and \textbf{D) test-time data augmentation}. The mean and the standard deviation are only used to keep the visualization simple. In practice other methods, could be utilized. For the deterministic approaches the idea of predicting the parameters of an probability distribution $\Xi$ is visualized, other approaches which base on tools additional to the prediction network are not visualized here.
    \label{fig:methods_overview}} 
\end{figure*}
In the following, the main ideas and further extensions of the four types are presented and their main properties are discussed. In Figure \ref{fig:methods_diagram}, an overview of the different types and methods is given. In Figure \ref{fig:methods_overview}, the different underlying principles that are used to differentiate between the different types of methods are presented. 
Table \ref{tab:types_compare} summarizes the main properties of the methods presented in this work, such as complexity, computational effort, memory consumption, flexibility, and others. 
\begin{table*}[t]
\centering
\caption{Overview over the properties of internal and external deterministic single network methods. For a comparison of single deterministic network approaches with Bayesian, ensemble, and test-time augmentation methods, see Table \ref{tab:types_compare}.}
\begin{tabular}{>{\raggedright\arraybackslash}p{3.1cm}p{4cm}p{4cm}}

& \textbf{Internal Methods} & \textbf{External Methods} 
\\ \hline

\addlinespace[2ex]
\textbf{Description} 
& Estimate uncertainty using one evaluation of a single network without external components. 
&Estimate uncertainty using one evaluation of the network while relying on additional external components. 
\\

\addlinespace[2ex]
\textbf{Implementation effort} 
&Relatively low, but depends on explicit approach, often only loss and network output has to be fixed. 
&Relatively low, but depends on explicit approach.
\\ 

\addlinespace[2ex]
\textbf{Application on already trained networks possible} &No &Yes \\ 

\addlinespace[2ex]
\textbf{Separated prediction and uncertainty estimation}
&No 
&Yes 
\\ \hline

\end{tabular}
\label{tab:deterministic_compare}
\end{table*}
\subsection{\textbf{Single Deterministic Methods}}\label{sec:deterministic_methods}
For deterministic neural networks the parameters are deterministic and each repetition of a forward pass delivers the same result. With single deterministic network methods for uncertainty quantification, we summarize all approaches where the uncertainty on a prediction $y^*$ is computed based on one single forward pass within a deterministic network. In the literature, several such approaches can be found. They can be roughly categorized into approaches where one single network is explicitly modeled and trained in order to quantify uncertainties \cite{evidential.neural.networks, prior.network,inhibited.softmax,nandy2020towards,interval.neural.networks.uncertainty.score} and approaches that use additional components in order to give an uncertainty estimate on the prediction of a network \cite{second.opinion.medical,density.estimation.in.representation.space,oberdiek2018classification,lee2020gradients}. While for the first type, the uncertainty quantification affects the training procedure and the predictions of the network, the latter type is in general applied on already trained networks. Since trained networks are not modified by such methods, they have no effect on the network's predictions. In the following, we call these two types \textit{internal} and \textit{external} uncertainty quantification approaches.
\subsubsection{Internal Uncertainty Quantification Approaches}
Many of the internal uncertainty quantification approaches followed the idea of predicting the parameters of a distribution over the predictions instead of a direct pointwise maximum-a-posteriori estimation. Often, the loss function of such networks takes the expected divergence between the true distribution and the predicted distribution into account as e.g., in \cite{prior.network,malinin2019reverse}. The distribution over the outputs can be interpreted as a quantification of the model uncertainty (see Section \ref{sec:uncertainty_types_and_sources}), trying to emulate the behavior of a Bayesian modeling of the network parameters \cite{nandy2020towards}. The prediction is then given as the expected value of the predicted distribution. 

\begin{figure}[ht]
    \begin{tikzpicture}
    	\node[text width=0.8cm] at (-1.15,-0.7){Pred:};
    	\node[text width=0.8cm] at (-1.15,-1.1){Conf:};
    	\node[inner sep=0pt] at (0,0)
    	{\includegraphics{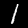}};
    	\node[text width=0.8cm, align=center] at (0,-0.7){1};
    	\node[text width=0.8cm, align=center] at (0,-1.1){1};
    	\node[inner sep=0pt] at (1,0)
    	{\includegraphics{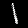}};
    	\node[text width=0.8cm, align=center] at (1,-0.7){1};
    	\node[text width=0.8cm, align=center] at (1,-1.1){0.8};
    	\node[inner sep=0pt] at (2,0)
    	{\includegraphics{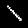}};
    	\node[text width=0.8cm, align=center, red] at (2,-0.7){3};
    	\node[text width=0.8cm, align=center] at (2,-1.1){0.97};
    	\node[inner sep=0pt] at (3,0)
    	{\includegraphics{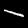}};
    	\node[text width=0.8cm, align=center, red] at (3,-0.7){7};
    	\node[text width=0.8cm, align=center] at (3,-1.1){1};
    	\node[inner sep=0pt] at (4,0)
    	{\includegraphics{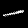}};
    	\node[text width=0.8cm, align=center, red] at (4,-0.7){7};
    	\node[text width=0.8cm, align=center] at (4,-1.1){0.98};
    	\node[inner sep=0pt] at (5,0)
    	{\includegraphics{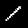}};
    	\node[text width=0.8cm, align=center] at (5,-0.7){1};
    	\node[text width=0.8cm, align=center] at (5,-1.1){0.94};
    	\node[inner sep=0pt] at (6,0)
    	{\includegraphics{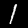}};
    	\node[text width=0.8cm, align=center] at (6,-0.7){1};
    	\node[text width=0.8cm, align=center] at (6,-1.1){1};
    	\node[text width=0.8cm] at (-1.15,-2.7){Pred:};
    	\node[text width=0.8cm] at (-1.15,-3.1){Conf:};
    	\node[inner sep=0pt] at (0,-2)
    	{\includegraphics{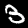}};
    	\node[text width=0.8cm, align=center] at (0,-2.7){3};
    	\node[text width=0.8cm, align=center] at (0,-3.1){1};
    	\node[inner sep=0pt] at (1,-2)
    	{\includegraphics{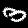}};
    	\node[text width=0.8cm, align=center] at (1,-2.7){8};
    	\node[text width=0.8cm, align=center] at (1,-3.1){0.62};
    	\node[inner sep=0pt] at (2,-2)
    	{\includegraphics{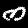}};
    	\node[text width=0.8cm, align=center, red] at (2,-2.7){1};
    	\node[text width=0.8cm, align=center] at (2,-3.1){1};
    	\node[inner sep=0pt] at (3,-2)
    	{\includegraphics{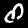}};
    	\node[text width=0.8cm, align=center, red] at (3,-2.7){1};
    	\node[text width=0.8cm, align=center] at (3,-3.1){0.84};
    	\node[inner sep=0pt] at (4,-2)
    	{\includegraphics{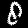}};
    	\node[text width=0.8cm, align=center, red] at (4,-2.7){8};
    	\node[text width=0.8cm, align=center] at (4,-3.1){0.99};
    	\node[inner sep=0pt] at (5,-2)
    	{\includegraphics{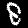}};
    	\node[text width=0.8cm, align=center, red] at (5,-2.7){8};
    	\node[text width=0.8cm, align=center] at (5,-3.1){1};
    	\node[inner sep=0pt] at (6,-2)
    	{\includegraphics{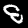}};
    	\node[text width=0.8cm, align=center, red] at (6,-2.7){8};
    	\node[text width=0.8cm, align=center] at (6,-3.1){0.99};
    	\node[text width=0.8cm] at (-1.15,-4.7){Pred:};
    	\node[text width=0.8cm] at (-1.15,-5.1){Conf:};
    	\node[inner sep=0pt] at (0,-4)
    	{\includegraphics{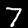}};
    	\node[text width=0.8cm, align=center] at (0,-4.7){7};
    	\node[text width=0.8cm, align=center] at (0,-5.1){1};
    	\node[inner sep=0pt] at (1,-4)
    	{\includegraphics{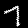}};
    	\node[text width=0.8cm, align=center] at (1,-4.7){7};
    	\node[text width=0.8cm, align=center] at (1,-5.1){1};
    	\node[inner sep=0pt] at (2,-4)
    	{\includegraphics{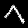}};
    	\node[text width=0.8cm, align=center, red] at (2,-4.7){7};
    	\node[text width=0.8cm, align=center] at (2,-5.1){0.88};
    	\node[inner sep=0pt] at (3,-4)
    	{\includegraphics{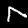}};
    	\node[text width=0.8cm, align=center, red] at (3,-4.7){1};
    	\node[text width=0.8cm, align=center] at (3,-5.1){0.86};
    	\node[inner sep=0pt] at (4,-4)
    	{\includegraphics{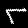}};
    	\node[text width=0.8cm, align=center, red] at (4,-4.7){5};
    	\node[text width=0.8cm, align=center] at (4,-5.1){0.87};
    	\node[inner sep=0pt] at (5,-4)
    	{\includegraphics{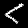}};
    	\node[text width=0.8cm, align=center, red] at (5,-4.7){6};
    	\node[text width=0.8cm, align=center] at (5,-5.1){0.97};
    	\node[inner sep=0pt] at (6,-4)
    	{\includegraphics{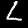}};
    	\node[text width=0.8cm, align=center, red] at (6,-4.7){6};
    	\node[text width=0.8cm, align=center] at (6,-5.1){1};
    \end{tikzpicture}
    \caption{Predictions received from a LeNet network trained on MNIST's handwritten digits from 0 to 9 and evaluated on different rotations of test samples. One can clearly see, that for some rotations the network gives a high confidence on the false class due to confusion (e.g.: 3 is confused with 8) or representations not seen at training. These examples represent a simple case of how a basic classification network can lead to overconfident wrong predictions under data distribution shifts.}%
\label{fig:softmax_probability}
\end{figure}

For classification tasks, the output in general represents class probabilities. These probabilities are a result of applying the softmax function 
\begin{align}
\begin{split}
&\text{softmax}:\mathbb{R}^K\rightarrow\left\{z\in  \mathbb{R}^K\vert z_i \geq 0, \sum_{k=1}^K z_k =1\right\} \\
&\text{softmax}(z)_j = \frac{\exp(z_j)}{\sum_{k=1}^K\exp(z_k)}
\end{split}
\end{align}
for multiclass settings and the sigmoid function 
\begin{align}
\begin{split}
&\text{sigmoid}:\mathbb{R}\rightarrow[0,1] \\
&\text{sigmoid}(z) = \frac{1}{1+\exp(-z)}
\end{split}
\end{align}
for binary classification tasks on the logits $z$. These probabilities can be already interpreted as a prediction of the data uncertainty. However, it is widely discussed that neural networks are often over-confident and the softmax output is often poorly calibrated, leading to inaccurate uncertainty estimates \cite{vasudevan2019towards,hendrycks2016baseline,evidential.neural.networks,inhibited.softmax}. Furthermore, the softmax output cannot be associated with model uncertainty. But without explicitly taking the model uncertainty into account, out-of-distribution samples could lead to outputs that certify a false confidence. For example, a network trained on cats and dogs will very likely not result in 50\% dog and 50\% cat when it is fed with the image of a bird. This is, because the network extracts features from the image and even though the features do not fit to the cat class, they might fit even less to the dog class. As a result, the network puts more probability on cat. Furthermore, it was shown that the combination of rectified linear unit (ReLu) networks and the softmax output leads to settings where the network becomes more and more confident as the distance between an out-of-distribution sample and the learned training set becomes larger \cite{hein2019relu}.
Figure \ref{fig:softmax_probability} shows an example where the rotation of a digit from MNIST leads to false predictions with high softmax values. 
This phenomenon is described and further investigated by Hein et al. \cite{hein2019relu} who proposed a method to avoid this behaviour, based on enforcing a uniform predictive distribution far away from the training data.

\begin{figure}
    \centering
     \begin{subfigure}[b]{0.15\textwidth}
         \centering
         \includegraphics[width=\textwidth]{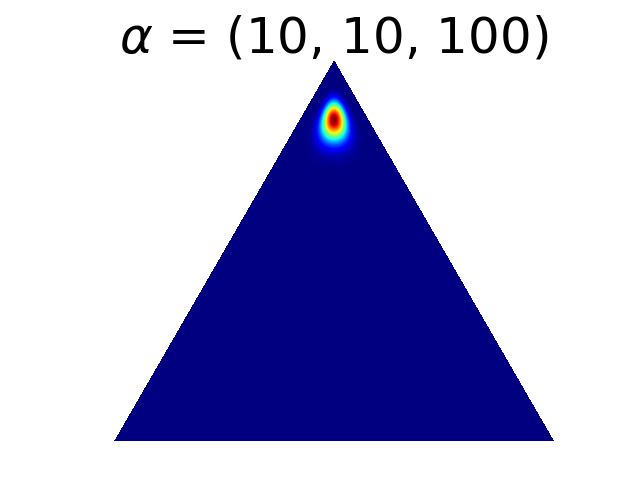}
         \caption{Low uncertainty}
         \label{fig:y equals x}
     \end{subfigure}
     \begin{subfigure}[b]{0.15\textwidth}
         \centering
         \includegraphics[width=\textwidth]{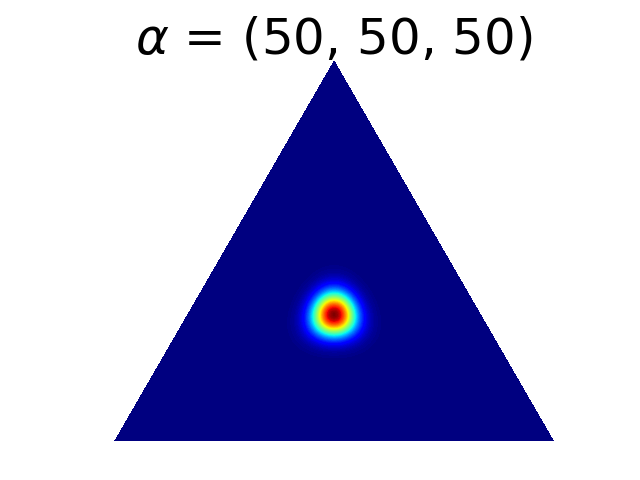}
         \caption{Data uncertainty}
         \label{fig:y equals x}
     \end{subfigure}
     \begin{subfigure}[b]{0.15\textwidth}
         \centering
         \includegraphics[width=\textwidth]{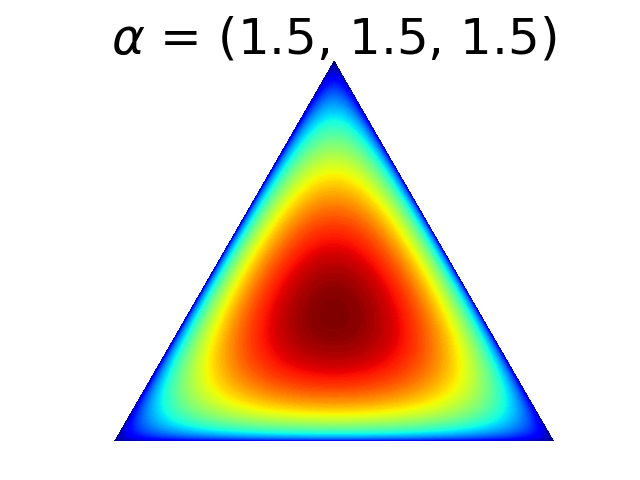}
         \caption{Out-of-dist.}
         \label{fig:y equals x}
     \end{subfigure}
    \caption{The desired behaviors of a Dirichlet distribution over categorical distributions. The visualizations show three Dirichlet distributions over three classes. Each node of the simplex represents one class. In (a) the sharp Dirichlet distribution with its expectation close to the upper node represents a certain prediction of a categorical distribution. In (b) the sharp Dirichlet distribution in the center of the simplex represents high data uncertainty but low distributional uncertainty. In (c) the flat Dirichlet distribution indicates high distributional uncertainty. }
    \label{fig:simplex_uncertainty}
\end{figure}

Several other classification approaches \cite{evidential.neural.networks,prior.network,malinin2019reverse,nandy2020towards} followed a similar idea of taking the logit magnitude into account, but make use of the Dirichlet distribution. The Dirichlet distribution is the conjugate prior of the categorical distribution and hence can be interpreted as a distribution over categorical distributions. The density of the Dirichlet distribution is defined by 
\begin{equation*}
    \text{Dir}\left(\mu\vert\alpha\right) = \frac{\Gamma(\alpha_0)}{\prod_{c=1}^{K}\Gamma(\alpha_c)}\prod_{c=1}^{K}\mu_c^{\alpha_c-1}, 
    \quad \alpha_c > 0,~\alpha_0=\sum_{c=1}^K\alpha_c \quad,
\end{equation*}
where $\Gamma$ is the gamma function, $\alpha_1, ..., \alpha_K$ are called the concentration parameters, and the scalar $\alpha_0$ is the precision of the distribution. In practice, the concentrations $\alpha_1,...,\alpha_K$ are derived by applying a strictly positive transformation, as for example the exponential function, to the logit values. As visualized in Figure \ref{fig:simplex_uncertainty}, a higher concentration value leads to a sharper Dirichlet distribution. \\
The set of all class probabilities of a categorical distribution over $k$ classes is equivalent to a $k-1$-dimensional standard or probability simplex. Each node of this simplex represents a probability vector with the full probability mass on one class and each convex combination of the nodes represents a categorical distribution with the probability mass distributed over multiple classes. Malinin et al. \cite{prior.network} argued that a high model uncertainty should lead to a lower precision value and therefore to a flat distribution over the whole simplex, since the network is not familiar with the data. In contrast to this, data uncertainty should be represented by a sharper but also centered distribution, since the network can handle the data, but cannot give a clear class preference. In Figure \ref{fig:simplex_uncertainty} the different desired behaviors are shown. 
The Dirichlet distribution is utilized in several approaches as \textit{Dirichlet Prior Networks} \cite{dirichlet.networks,prior.network} and \textit{Evidential Neural Networks} \cite{joo2020being,evidential.neural.networks}. Both of these network types output the parameters of a Dirichlet distribution from which the categorical distribution describing the class probabilities can be derived. The general idea of prior networks \cite{prior.network} is already described above and is visualized in Figure \ref{fig:simplex_uncertainty}. 
Prior networks are trained in a multi task way with the goal of minimizing the expected Kullback-Leibler (KL) divergence between the predictions of in-distribution data and a sharp Dirichlet distribution and between a flat Dirichlet distribution and the predictions of out-of-distribution data \cite{prior.network}. Besides the main motivation of a better separation between in-distribution and OOD samples, these approaches also improve the separation between the confidence of correct and incorrect predictions, as was shown by \cite{tsiligkaridis2020failure}. As a follow up, \cite{malinin2019reverse} discussed that for the case that the data uncertainty is high, the forward definition of the KL-divergence can lead to an undesirable multi-model target distribution. In order to avoid this, they reformulated the loss using the reverse KL-divergence. The experiments showed improved results in the uncertainty estimation as well as for the adversarial robustness. 
Zhao et al. \cite{DBLP:journals/corr/abs-1910-04819} extended the Dirichlet network approach by a new loss function that aims at minimizing an upper bound on the expected error based on the $\mathcal{L}_\infty$-norm, i.e. optimizing an expected worst-case upper bound. \cite{mixture.of.dirichlet} argued that using a mixture of Dirichlet distributions gives much more flexibility in approximating the posterior distribution. Therefore, an approach where the network predicts the parameters for a mixture of $K$ Dirichlet distributions was suggested. For this, the network logits represent the parameters for $M$ Dirichlet distributions and additionally $M$ weights $\omega_i, i=1,..,M$ with the constraint $\sum_{i=1}^M\omega_i=1$ are optimized. Nandy et al. \cite{nandy2020towards} analytically showed that for in-domain samples with high data uncertainty, the Dirichlet distribution predicted for a false prediction is often flatter than for a correct prediction. They argued that this makes it harder to differentiate between in- and out-of-distribution predictions and suggested a regularization term towards maximizing the gap between in- and out-of-distribution samples. \\
Evidential neural networks \cite{evidential.neural.networks} also optimize the parameterization of a single Dirichlet network. The loss formulation is derived by using subjective logic and interpret the logits as multinomial opinions or beliefs, as introduced in Evidence or Dempster-Shafer theory \cite{dempster1968generalization}. Evidential neural networks set the total amount of evidence in relation with the number of classes and conclude a value of uncertainty from this, i.e. receiving an additional "I don't know class". The loss is formulated as expected value of a basic loss, as for example categorical cross entropy, with respect to a Dirichlet distribution parameterized by the logits. Additionally, a regularization term is added, encouraging the network to not consider features that provide evidence for multiple classes at the same time, as for example a circle is for 6 and 8. Due to this, the networks do not differentiate between data uncertainty and model uncertainty, but learn whether they can give a certain prediction or not.  \cite{regularized.evidential.networks} extended this idea by differentiating between acuity and dissonance in the collected evidence in order to better separate in- and out-of-distribution samples. For that, two explicit data sets containing overlapping classes and out-of-distribution samples are needed to learn a regularization term. Amini et al. \cite{alex2019} transferred the idea of evidential neural networks from classification tasks to regression tasks by learning the parameters of an evidential normal inverse gamma distribution over an underlying Normal distribution. Charpentier et al. \cite{charpentier2020posterior} avoided the need of OOD data for the training process by using normalizing flows to learn a distribution over a latent space for each class. A new input sample is projected onto this latent space and a Dirichlet distribution is parameterized based on the class wise densities of the received latent point. 

Beside the Dirichlet distribution based approaches described above, several other internal approaches exist. In \cite{liang2017enhancing}, a relatively simple approach based on small pertubations on the training input data and the temperature scaling calibration is presented leading to an efficient differentiation of in- and out-of-distritbuion samples. Mo$\dot{\text{z}}$ejko et al. \cite{inhibited.softmax} made use of the inhibited softmax function. It contains an artificial and constant logit that makes the absolute magnitude of the single logits more determining in the softmax output. Van Amersfoort et al. \cite{kernel.network} showed that Radial Basis Function (RBF) networks can be used to achieve competitive results in accuracy and very good results regarding uncertainty estimation. RBF networks learn a linear transformation on the logits and classify inputs based on the distance between the transformed logits and the learned class centroids. In \cite{kernel.network}, a scaled exponentiated $L_2$ distance was used. The data uncertainty can be directly derived from the distances between the centroids. By including penalties on the Jacobian matrix in the loss function, the network was trained to be more sensitive to changes in the input space. As a result, the method reached good performance on out-of-distribution detection. In several tests, the approach was compared to a five members deep ensemble \cite{deep.ensembles} and it was shown that this single network approach performs at least equivalently well on detecting out-of-distribution samples and improves the true-positive rate. \\
For regression tasks, Oala et al. \cite{interval.neural.networks.uncertainty.score} introduced an uncertainty score based on the lower and an upper bound output of an interval neural network. The interval neural network has the same structure as the underlying deterministic neural network and is initialized with the deterministic network's weights. In contrast to Gaussian representations of uncertainty given by a standard deviation, this approach can give non-symmetric values of uncertainty. Furthermore, the approach is found to be more robust in the presence of noise. Tagasovska and Lopez-Paz \cite{tagasovska2019single} presented an approach to estimate data and model uncertainty. A simultaneous quantile regression loss function was introduced in order to generate well-calibrated prediction intervals for the data uncertainty.  The model uncertainty is quantified based on a mapping from the training data to zero, based on so called Orthonormal Certificates. The aim was that out-of-distribution samples, where the model is uncertain, are mapped to a non-zero value and thus can be recognized. Kawashima et al. \cite{kawashima2020aleatoric} introduced a method which computes virtual residuals in the training samples of a regression task based on a cross-validation like pre-training step. With original training data expanded by the information of these residuals, the actual predictor is trained to give a prediction and a value of certainty. The experiments indicated that the virtual residuals represent a promising tool in order to avoid overconfident network predictions. 
\subsubsection{External Uncertainty Quantification Approaches}~\\
External uncertainty quantification approaches do not affect the models' predictions, since the evaluation of the uncertainty is separated from the underlying prediction task. Furthermore, several external approaches can be applied to already trained networks at the same time without affecting each other. Raghu et al. \cite{second.opinion.medical} argued that when both tasks, the prediction and the uncertainty quantification, are done by one single method, the uncertainty estimation is biased by the actual prediction task. Therefore, they recommended a "direct uncertainty prediction" and suggested to train two neural networks, one for the actual prediction task and a second one for the prediction of the uncertainty on the first network's predictions. Similarly, Ramalho and Miranda \cite{density.estimation.in.representation.space} introduced an additional neural network for uncertainty estimation. But in contrast to \cite{second.opinion.medical}, the representation space of the training data is considered and the density around a given test sample is evaluated. The additional neural network uses this training data density in order to predict whether the main network's estimate is expected to be correct or false. Hsu et al. \cite{hsu2020generalized} detected out-of-distribution examples in classification tasks at test-time by predicting total probabilities for each class, additional to the categorical distribution given by the softmax output. The class wise total probability is predicted by applying the sigmoid function to the network's logits. Based on these total probabilities, OOD examples can be identified as those with low class probabilities for all classes.  \\
In contrast to this, Oberdiek et al. \cite{oberdiek2018classification} took the sensitivity of the model, i.e. the model's slope, into account by using gradient metrics for the uncertainty quantification in classification tasks. Lee et al. \cite{lee2020gradients} applied a similar idea but made use of back propagated gradients. In their work they presented state of the art results on out-of-distribution and corrupted input detection. 
\subsubsection{Summing Up Single Deterministic Methods}
Compared to many other principles, single deterministic methods are computational efficient in training and evaluation. For training, only one network has to be trained and often the approaches can even be applied on pre-trained networks. Depending on the actual approach, only a single or at most two forward passes have to be fulfilled for evaluation. The underlying networks could contain more complex loss functions, which slows down the training process \cite{evidential.neural.networks} or external components that have to be trained and evaluated additionally \cite{second.opinion.medical}. But in general, this is still more efficient than the number of predictions needed for ensembles based methods (Section \ref{sec:ensemble.methods}), Bayesian methods (Section \ref{sec:bayesian_methods}), and test-time data augmentation methods (Section \ref{sec:test_time_augmentation}).
A drawback of single deterministic neural network approaches is the fact that they rely on a single opinion and can therefore become very sensitive to the underlying network architecture, training procedure, and the training data.
\begin{table*}[t]
\centering
\caption{Overview over the properties different types of Bayesian neural networks approaches. The properties are stated relatively among the approaches. The properties can not used as comparison to other uncertainty methods as ensembles, single deterministic models, and test-time augmentation methods. For a comparison of Bayesian methods to these methods see Table \ref{tab:types_compare}}
\begin{tabular}{>{\raggedright\arraybackslash}p{3.1cm}p{4cm}p{4cm}p{4cm}}

& \textbf{Variational Inference} 
& \textbf{Sampling  Approaches}
& \textbf{Laplace  Approximation} 
\\ \hline

\addlinespace[2ex]
\textbf{Description} 
&Variational inference approaches approximate the (in general intractable) posterior distribution by optimizing over a family of tractable distributions. Achieved by minimizing the KL divergence. 
&Representation of the target random variable from which realizations can be sampled. Such methods are based on Markov Chain Monte Carlo and further extensions.
&Simplifies the target distribution by approximating the log-posterior distribution and then, based on this approximation, deriving a normal distribution on the network weights. 
\\ 

\addlinespace[2ex]
\textbf{Analytic expression} 
&Yes 
&No 
&Yes 
\\

\addlinespace[2ex]
\textbf{Can be applied to pretrained networks} 
&No 
&No 
&Yes 
\\

\addlinespace[2ex]
\textbf{Deterministic?} 
&Yes  
&No  
&Yes 
\\

\addlinespace[2ex]
\textbf{Unbiased?} 
&No 
&Yes 
&No 
\\

\addlinespace[2ex]
\textbf{Optimum} 
&Local optimum 
&Hard to mix between modes 
&Local optimum 
\\

\addlinespace[2ex]
\textbf{Convergence} 
& Easy to assess convergence 
& Hard to assess convergence 
& Easy to assess convergence 
\\ 

\addlinespace[2ex]
\textbf{Computational effort at training time} 
&Medium - Convergence may be slowed down by regularization.  Additional parameters for uncertainty representation.
&High - M forward passes based on sampled parameters, otherwise intractable
&Low - Training of one deterministic model and Laplace approximation 
\\ 

\addlinespace[2ex]
\textbf{Computational effort at inference} 
&High - M forward passes based on sampled parameters, otherwise intractable
&High - M forward passes based on sampled parameters, otherwise intractable 
&High - M forward passes based on sampled parameters, otherwise intractable 
\\ \hline

\end{tabular}
\label{tab:bayesian_methods}
\end{table*}
\subsection{\textbf{Bayesian Neural Networks}}\label{sec:bayesian_methods}
Bayesian Neural Networks (BNNs) \cite{denker1987large, tishby1989consistent, buntine1991bayesian} have the ability to combine the scalability, expressiveness, and predictive performance of neural networks with the Bayesian learning as opposed to learning via the maximum likelihood principles. This is achieved by inferring the probability distribution over the network parameters $\theta=(w_1,...,w_K)$. More specifically, given a training input-target pair $(x, y)$ the posterior distribution over the space of parameters $p(\theta|x,y)$ is modelled by assuming a prior distribution over the parameters $p(\theta)$ and applying Bayes theorem:
\begin{equation}
\label{1.b.eq1}
p(\theta\vert x,y) = \frac{p(y\vert x,\theta)p(\theta)}{p(y|x)} \propto p(y\vert x,\theta)p(\theta).
\end{equation}
Here, the normalization constant in \eqref{1.b.eq1} is called the model evidence $p(y|x)$ which is defined as
\begin{equation}
p(y | x) = \int p(y\vert x, \theta)p(\theta)d\theta.
\end{equation}
Once the posterior distribution over the weights have been estimated, the prediction of an output $y^*$ for a new input data $x^*$ can be obtained by Bayesian Model Averaging or Full Bayesian Analysis that involves marginalizing the likelihood $p(y|x,\theta)$ with the posterior distribution:
\begin{equation}
\label{eq:bayesian_marginalization}
p(y^*|x^*, x, y) = \int p(y^*|x^*,\theta) p(\theta|x,y)d\theta.
\end{equation}
This Bayesian way of prediction is a direct application of the law of total probability and endows the ability to compute the principled predictive uncertainty. The integral of \eqref{eq:bayesian_marginalization} is intractable for the most common prior posterior pairs and approximation techniques are therefore typically applied. The most widespread approximation, the \textit{Monte Carlo Approximation}, follows the law of large numbers and approximates the expected value by the mean of $N$ deterministic networks, $f_{\theta_1}, ...,f_{\theta_N}$, parameterized by $N$ samples, $\theta_1, \theta_2, ..., \theta_N$, from the posterior distribution of the weights, i.e.
\begin{equation}
   	y^*\quad \approx \quad \frac{1}{N}\sum_{i=1}^{N} y_i^* \quad = \quad \frac{1}{N}\sum_{i=1}^{N} f_{\theta_i}(x^*).
\end{equation} 
Wilson and Izmailov \cite{wilson2020bayesian} argue that a key advantage of BNNs lie in this marginalization step, which particularly can improve both the accuracy and calibration of modern deep neural networks. We note that the use-cases of BNNs are not limited for uncertainty estimation but open up the possibility to bridge the powerful Bayesian tool-boxes within deep learning. Notable examples include Bayesian model selection \cite{mackay1992bayesian, sato2001online, corduneanu2001variational, ghosh2017model}, model compression \cite{louizos2017bayesian, federici2017improved, achterhold2018variational}, active learning \cite{mackay1992information, gal2017deep, kirsch2019batchbald}, continual learning \cite{nguyen2017variational, ebrahimi2019uncertainty, farquhar2019unifying, li2020continual}, theoretic advances in Bayesian learning \cite{khan2019approximate} and beyond.\\
While the formulation is rather simple, there exist several challenges. For example, no closed form solution exists for the posterior inference as conjugate priors do not typically exist for complex models such as neural networks \cite{bishop2006pattern}. Hence, approximate Bayesian inference techniques are often needed to compute the posterior probabilities. Yet, directly using approximate Bayesian inference techniques have been proven to be difficult as the size of the data and number of parameters are too large for the use-cases of deep neural networks. In other words, the integrals of above equations are not computationally tractable as the size of the data and number of parameters grows. Moreover, specifying a meaningful prior for deep neural networks is another challenge that is less understood.

In this survey, we classify the BNNs into three different types based on how the posterior distribution is inferred to approximate Bayesian inference:
\begin{itemize}
  \setlength\itemsep{0.5em}
    \item \textit{Variational inference} \cite{hinton1993keeping, barber1998ensemble}\\
    Variational inference approaches approximate the (in general intractable) posterior distribution by optimizing over a family of tractable distributions. 
    \item \textit{Sampling approaches} \cite{neal1992bayesian}\\
    Sampling approaches deliver a representation of the target random variable from which realizations can be sampled. Such methods are based on Markov Chain Monte Carlo and further extensions. 
    \item \textit{Laplace approximation} \cite{denker1991transforming, mackay1992practical} \\
    Laplace approximation simplifies the target distribution by approximating the log-posterior distribution and then, based on this approximation, deriving a normal distribution over the network weights.
\end{itemize}
While limiting our scope to these three categories, we also acknowledge several advances in related domains of BNN research. Some examples are (i) approximate inference techniques such as alpha divergence \cite{hernandez2016black, li2017dropout, minka2005divergence}, expectation propagation \cite{minka2001expectation, zhao2020probabilistic}, assumed density filtering \cite{hernandez2015probabilistic} etc, (ii) probabilistic programming to exploit modern Graphical Processing Units (GPUs) \cite{tran2016edward, tran2016deep, bingham2019pyro, cabanas2019inferpy}, (iii) different types of priors \cite{ito2005bayesian, sun2018functional}, (iv) advancements in theoretical understandings of BNNs \cite{depeweg2017sensitivity, khan2019approximate, farquhar2020try}, (iv) uncertainty propagation techniques to speed up the marginalization procedures \cite{postels2019sampling} and (v) computations of aleatoric uncertainty \cite{gast2018lightweight, depeweg2018decomposition, depeweg2017uncertainty}.\\
\subsubsection{Variational Inference}
The goal of variational inference is to infer the posterior probabilities $p(\theta|x,y)$ using a pre-specified family of distributions $q(\theta)$. Here, these so-called variational family $q(\theta)$ is defined as a parametric distribution. An example is the Multivariate Normal distribution where its parameters are the mean and the covariance matrix. The main idea of variational inference is to find the settings of these parameters that make $q(\theta)$ to be close to the posterior of interest $p(\theta|x,y)$. This measure of closeness between the probability distributions are given by the Kullback-Leibler (KL) divergence
\begin{equation}\label{eq:kl}
    \text{KL}(q||p) = \mathbb{E}_q\left [ \text{log} \frac{q(\theta)}{p(\theta|x,y)} \right ].
\end{equation}
Due to the posterior $p(\theta\vert x, y)$ the KL-divergence in \eqref{eq:kl} can not be minimized directly. Instead, the evidence lower bound (ELBO), a function that is equal to the KL divergence up to a constant, is optimized. For a given prior distribution on the parameters $p(\theta)$, the ELBO is given by
\begin{equation}
     L = \mathbb{E}_q\left[\log\frac{p(y|x,\theta)}{q(\theta)}\right]
\end{equation}
and for the KL divergence 
\begin{equation}
    \text{KL}(q||p) = -L + \log p(y\vert x)
\end{equation}
holds.

Variational methods for BNNs have been pioneered by Hinton and Van Camp \cite{hinton1993keeping} where the authors derived a diagonal Gaussian approximation to the posterior distribution of neural networks (couched in information theory - a minimum description length). Another notable extension in 1990s has been proposed by Barber and Bishop \cite{barber1998ensemble}, in which the full covariance matrix was chosen as the variational family, and the authors demonstrated how the ELBO can be optimized for neural networks. Several modern approaches can be viewed as extensions of these early works \cite{hinton1993keeping, barber1998ensemble} with a focus on how to scale the variational inference to modern neural networks. \\
An evident direction with the current methods are the use of stochastic variational inference (or Monte-Carlo variational inference), where the optimization of ELBO is performed using mini-batch of data. One of the first connections to stochastic variational inference has been proposed by Graves et al. \cite{graves2011practical} with Gaussian priors. In 2015, Blundell et al. \cite{blundell2015weight} introduced Bayes By Backprop, a further extension of stochastic variational inference  \cite{graves2011practical} to non-Gaussian priors and demonstrated how the stochastic gradients can be made unbiased. Notable, Kingma et al. \cite{kingma2015variational} introduced the local reparameterization trick to reduce the variance of the stochastic gradients. One of the key concepts is to reformulate the loss function of neural network as the ELBO. As a result the intractable posterior distribution is indirectly optimized and variational inference is compatible to back-propagation with certain modifications to the training procedure. These extensions widely focus on the fragility of stochastic variational inference that arises due to sensitivity to initialization, prior definition and variance of the gradients. These limitations have been addressed recently by Wu et al. \cite{wu2018deterministic}, where a hierarchical prior was used and the moments of the variational distribution are approximated deterministically. \\
Above works commonly assumed mean-field approximations as the variational family, neglecting the correlations between the parameters. In order to make more expressive variational distributions feasible for deep neural networks, several works proposed to infer using the matrix normal distribution \cite{louizos2016structured, zhang2018noisy, sun2017learning} or more expressive variants \cite{bae2018eigenvalue, mishkin2018slang} where the covariance matrix is decomposed into the Kronecker products of smaller matrices or in a low rank form plus a positive diagonal matrix. A notable contribution towards expressive posterior distributions has been the use of normalizing flows \cite{rezende2015variational, louizos2017multiplicative} - a hierarchical probability distribution where a sequence of invertible transformations are applied so that a simple initial density function is transformed into a more complex distribution. Interestingly, Farquhar et al. \cite{farquhar2020try} argue that mean-field approximation is not a restrictive assumption, and the layer-wise weight correlations may not be as important as capturing the depth-wise correlations. While the claim of Farquhar et al. \cite{farquhar2020try} may remain to be an open question, the mean-field approximations have an advantage on smaller computational complexities \cite{farquhar2020try}. For example, Osawa et al. \cite{osawa2019practical} demonstrated that variational inference can be scaled up to ImageNet size data-sets and architectures using multiple GPUs and proposed practical tricks such as data augmentation, momentum initialization and learning rate scheduling. \\
One of the successes in variational methods have been accomplished by casting existing stochastic elements of deep learning as variational inference. A widely known example is Monte Carlo Dropout (MC Dropout) where the dropout layers are formulated as Bernoulli distributed random variables, and training a neural network with dropout layers can be approximated as performing variational inference \cite{gal2015bayesian, gal2016dropout, gal2017concrete}. A main advantage of MC dropout is that the predictive uncertainty can be computed by activating dropout not only during training, but also at test time. In this way, once the neural network is trained with dropout layers, the implementation efforts can be kept minimum and the practitioners do not need expert knowledge to reason about uncertainty - certain criteria that the authors are attributing to its success  \cite{gal2016dropout}. The practical values of this method has been demonstrated also in several works \cite{eaton2018towards,loquercio2020general,marcmohsin2020uncertainty} and resulted in different extensions (evaluating the usage of different dropout masks for example for convolutional layers \cite{tassi2019bayesian} or by changing the representations of the predictive uncertainty into model and data uncertainties \cite{kendall2017uncertainties}). Approaches that build upon the similar idea but randomly drop incoming activations of a node, instead of dropping an activation for all following nodes, were also proposed within the literature \cite{mobiny2019dropconnect} and called drop connect. This was found to be more robust on the uncertainty representation, even though it was shown that a combination of both can lead to higher accuracy and robustness in the test predictions \cite{mcclure2016robustly}. Lastly, connections of variation inference to Adam \cite{khan2018fast}, RMS Prop \cite{khan2017vprop} and batch normalization \cite{atanov2018uncertainty} have been further suggested in the literature.\\
\subsubsection{Sampling Methods}
Sampling methods, or also often called Monte Carlo methods, are another family of Bayesian inference algorithms that represent uncertainty without a parametric model. Specifically, sampling methods use a set of hypotheses (or samples) drawn from the distribution and offer an advantage that the representation itself is not restricted by the type of distribution (e.g. can be multi-modal or non-Gaussian) - hence probability distributions are obtained non-parametrically. Popular algorithms within this domain are particle filtering, rejection sampling, importance sampling and Markov Chain Monte Carlo sampling (MCMC) \cite{bishop2006pattern}. \\ 
In case of neural networks, MCMC is often used since alternatives such as rejection and importance sampling are known to be inefficient for such high dimensional problems. The main idea of MCMC is to sample from arbitrary distributions by transition in state space where this transition is governed by a record of the current state and the proposal distribution that aims to estimate the target distribution (e.g. the true posterior). To explain this, let us start defining the Markov Chain: a Markov Chain is a distribution over random variables $x_1, \cdots, x_T$ which follows the state transition rule:
\begin{align}
p(x_1,\cdots, x_T) = p(x_1)\prod_{t=2}^{T}p(x_t|x_{t-1}),
\end{align}
i.e. the next state only depends on the current state and not on any other former state. 
In order to draw samples from the true posterior, MCMC sampling methods first generate samples in an iterative and the Markov Chain fashion. Then, at each iteration, the algorithm decides to either accept or reject the samples where the probability of acceptance is determined by certain rules. In this way, as more and more samples are produced, their values can approximate the desired distribution. \\
Hamiltonian Monte Carlo or Hybrid Monte Carlo (HMC) \cite{duane1987hybrid} is an important variant of MCMC sampling method (pioneered by Neals \cite{neal1992bayesian, neal1994improved, neal1995bayesian, neal2011mcmc} for neural networks), and is often known to be the gold standards of Bayesian inference \cite{neal2011mcmc, dubey2016variance, li2017dropout}. The algorithm works as follows: (i) start by initializing a set of parameters $\theta$ (either randomly or in a user-specific manner). Then, for a given number of total iterations, (ii) instead of a random walk, a momentum vector - an auxiliary variable $\rho$ is sampled, and the current value of parameters $\theta$ is updated via the Hamiltonian dynamics:
\begin{align}
H(\rho, \theta) = -\text{log} p(\rho, \theta) = -\text{log} p(\rho|\theta) - \text{log} p(\theta).
\end{align}
Defining the potential energy ($V(\theta) = - log p(\theta)$ and the kinetic energy $T(\rho|\theta) = -\text{log} p(\rho|\theta)$, the update steps via Hamilton's equations are governed by,
\begin{align}
\frac{d\theta}{dt} &= \frac{\partial H}{\partial \rho} = \frac{\partial T}{\partial \rho} \ \text{and}\\
\frac{d\rho}{dt} &= - \frac{\partial H}{\partial \theta} = -\frac{\partial T}{\partial \theta} - \frac{\partial V}{\partial \theta}.
\end{align}
The so-called leapfrog integrator is used as a solver \cite{leimkuhler2004simulating}. (iii) For each step, a Metropolis acceptance criterion is applied to either reject or accept the samples (similar to MCMC). Unfortunately, HMC requires the processing of the entire data-set per iteration, which is computationally too expensive when the data-set size grows to million to even billions. Hence, many modern algorithms focus on how to perform the computations in a mini-batch fashion stochastically. In this context, for the first time, Welling and Teh \cite{welling2011bayesian} proposed to combine Stochastic Gradient Descent (SGD) with Langevin dynamics (a form of MCMC \cite{rossky1978brownian, roberts2002langevin, neal2011mcmc}) in order to obtain a scalable approximation to MCMC algorithm based on mini-batch SGD \cite{kushner2003stochastic, goodfellow2016deep}. The work demonstrated that performing Bayesian inference on Deep Neural Networks can be as simple as running a noisy SGD. This method does not include the momentum term of HMC via using the first order Langevin dynamics and opened up a new research area on Stochastic Gradient Markov Chain Monte Carlo (SG-MCMC).  \\
Consequently, several extensions are available which include the use of 2nd order information such as preconditioning and optimizing with the Fisher Information Matrix (FIM) \cite{ma2015complete, marceau2017natural, nado2018stochastic}, the Hessian \cite{simsekli2016stochastic, zhang2011quasi, fu2016quasi}, adapting preconditioning diagonal matrix \cite{li2016preconditioned}, generating samples from non-isotropic target densities using Fisher scoring \cite{ahn2012bayesian}, and samplers in the Riemannian manifold \cite{patterson2013stochastic} using the first order Langevin dynamics and Levy diffusion noise and momentum \cite{ye2018stochastic}. Within these methods, the so-called parameter dependent diffusion matrices are incorporated with an intention to offset the stochastic perturbation of the gradient. To do so, the "thermostat" ideas \cite{ding2014bayesian, shang2015covariance, leimkuhler2016adaptive} are proposed so that a prescribed constant temperature distribution is maintained with the parameter dependent noise. Ahn et al. \cite{ahn2014distributed} devised a distributed computing system for SG-MCMC to exploit the modern computing routines, while Wang et al. \cite{wang2018adversarial} showed that Generative Adversarial Models (GANs) can be used to distill the samples for improved memory efficiency, instead of distillation for enhancing the run-time capabilities of computing predictive uncertainty \cite{balan2015bayesian}. Lastly, other recent trends are techniques that reduce the variance \cite{dubey2016variance, zou2018stochastic} and bias \cite{durmus2016stochastic, durmus2019high} arising from stochastic gradients. \\
Concurrently, there have been solid advances in theory of SG-MCMC methods and their applications in practice. Sato and Nakagawa \cite{sato2014approximation}, for the first time, showed that the SGLD algorithm with constant step size weakly converges; Chen et al. \cite{chen2015convergence} showed that faster convergence rates and more accurate invariant measures can be observed for SG-MCMCs with higher order integrators rather than a 1st order Euler integrator while Teh et al. \cite{teh2016consistency} studied the consistency and fluctuation properties of the SGLD. As a result, verifiable conditions obeying a central limit theorem for which the algorithm is consistent, and how its asymptotic bias-variance decomposition depends on step-size sequences have been discovered. A more detailed review of the SG-MCMC with a focus on supporting theoretical results can be found in Nemeth and Fearnhead \cite{nemeth2019stochastic}. Practically, SG-MCMC techniques have been applied to shape classification and uncertainty quantification \cite{li2016learning}, empirically study and validate the effects of tempered posteriors (or called cold-posteriors) \cite{wenzel2020good} and train a deep neural network in order to generalize and avoid over-fitting \cite{ye2017langevin, chandra2019langevin}.\\
\subsubsection{Laplace Approximation}
The goal of the Laplace Approximation is to estimate the posterior distribution over the parameters of neural networks $p(\theta\mid x,y)$ around a local mode of the loss surface with a Multivariate Normal distribution. The Laplace Approximation to the posterior can be obtained by taking the second-order Taylor series expansion of the log posterior over the weights around the MAP estimate $\hat \theta$ given some data $(x, y)$. If we assume a Gaussian prior with a scalar precision value $\tau>0$, then this corresponds to the commonly used $L_2$-regularization, and the Taylor series expansion results in
\begin{align*}
\log p(\theta\mid x,y) & \approx \log p(\hat \theta \mid x,y) \\ & + \frac{1}{2}(\theta -\hat \theta)^T(H + \tau I)(\theta -\hat \theta),
\end{align*}
where the first-order term vanishes because the gradient of the log posterior $\delta\theta=\nabla \log p(\theta \mid x,y)$ is zero at the maximum $\hat \theta$. Taking the exponential on both sides and approximating integrals by reverse engineering densities, the weight posterior is approximately a Gaussian with the mean $\hat \theta$ and the covariance matrix $(H+\tau I)^{-1}$ where $H$ is the Hessian of $\log p(\theta\mid x,y) $. This means that the model uncertainty is represented by the Hessian $H$ resulting in a Multivariate Normal distribution:
\begin{align}
    p(\theta\mid x,y) \sim \mathcal{N}\left(\hat \theta, (H+\tau I)^{-1}\right).
\end{align}
In contrast to the two other methods described, the Laplace approximation can be applied on already trained networks, and is generally applicable when using standard loss functions such as MSE or cross entropy and piece-wise linear activations (e.g RELU). Mackay \cite{mackay1992practical} and Denker et al. \cite{denker1991transforming} have pioneered the Laplace approximation for neural networks in 1990s, and several modern methods provide an extension to deep neural networks \cite{botev2017practical,martens2015optimizing,ritter2018scalable,lee2020estimating}. \\
The core of the Laplace Approximation is the estimation of the Hessian. Unfortunately, due to the enormous number of parameters in modern neural networks, the Hessian matrices cannot be computed in a feasible way as opposed to relative smaller networks in Mackay \cite{mackay1992practical} and Denker et al. \cite{denker1991transforming}. Consequently, several different ways for approximating $H$ have been proposed in the literature. A brief review is as follows. Instead of diagonal approximations  (e.g. \cite{Yann1989}, \cite{Salimans2016}), several researchers have been focusing on including the off-diagonal elements (e.g. \cite{Liu1998}, \cite{hennig2013} and \cite{Roux2010AFN}). Amongst them, layer-wise Kronecker Factor approximation of \cite{Grosse16}, \cite{martens2015optimizing}, \cite{botev2017practical} and \cite{chen2018bda} have demonstrated a notable scalability \cite{Ba2017}. A recent extension can be found in \cite{George2018} where the authors propose to re-scale the eigen-values of the Kronecker factored matrices so that the diagonal variance in its eigenbasis is accurate. The work presents an interesting idea as one can prove that in terms of a Frobenius norm, the proposed approximation is more accurate than that of \cite{martens2015optimizing}. However, as this approximation is harmed by inaccurate estimates of eigenvectors, Lee et al. \cite{lee2020estimating} proposed to further correct the diagonal elements in the parameter space. \\
Existing works obtain Laplace Approximation using various approximation of the Hessian in the line of fidelity-complexity trade-offs. For several works, an approximation using the diagonal of the Fisher information matrix or Gauss Newton matrix, leading to independently distributed model weights, have been utilized in order to prune weights \cite{lecun_braindemage} or perform continual learning in order to avoid catastrophic forgetting \cite{kirkpatrick2017overcoming}. In Ritter et al. \cite{ritter2018scalable}, the Kronecker factorization of the approximate block-diagonal Hessian \cite{martens2015optimizing, botev2017practical} have been applied to obtain scalable Laplace Approximation for neural networks. With this, the weights among different layers are still assumed to be independently distributed, but not the correlations within the same layer. Recently, building upon the current understandings of neural network's loss landscape that many eigenvalues of the Hessian tend to be zero, \cite{lee2020estimating} developed a low rank approximation that leads to sparse representations of the layers' co-variance matrices. Furthermore, Lee et al. \cite{lee2020estimating} demonstrated that the Laplace Approximation can be scaled to ImageNet size data-sets and architectures, and further showed that with the proposed sparsification technique, the memory complexity of modelling correlations can be made similar to the diagonal approximation. Lastly, Kristiadi et al. \cite{kristiadi2020being} proposed a simple procedure to compute the last-layer Gaussian approximation (neglecting the model uncertainty in all other layers of neural networks), and showed that even such a minimalist solution can mitigate overconfidence predictions of ReLU networks.\\
Recent efforts have extended the Laplace Approximation beyond the Hessian approximation. To tackle the widely known assumption that the Laplace Approximation is for the bell shaped true posterior and thus resulting in under-fitting behavior \cite{ritter2018scalable}, Humt et al. \cite{humt2020bayesian} proposed to use Bayesian Optimization and showed that hyperparameters of the Laplace Approximation can be efficiently optimized with increaed calibration performance. Another work in this domain is by Kristiadi et al. \cite{kristiadi2020learnable}, who proposed uncertainty units - a new type of hidden units that changes the geometry of the loss landscape so that more accurate inference is possible. While Shinde et al. \cite{shinde2020learning} demonstrated practical effectiveness of the Laplace Approximation to the autonomous driving applications, Feng et al. \cite{feng2019introspective} showed the possibility to (i) incorporate contextual information and (ii) domain adaptation in a semi-supervised manner within the context of image classification. This is achieved by designing unary potentials within a Conditional Random Field. Several real-time methods also exist that do not require multiple forward passes to compute the predictive uncertainty. So-called linearized Laplace Approximation has been proposed in \cite{foong2019between, immer2020improving} using the ideas of Mackay \cite{mackay1992information} and have been extended with Laplace bridge for classification \cite{hobbhahn2020fast}. Within this framework, Daxberger et al. \cite{daxberger2020expressive} proposed inferring the sub-networks to increase the expressivity of covariance propagation while remaining computationally tractable.\\
\subsubsection{Sum Up Bayesian Methods} Bayesian methods for deep learning have emerged as a strong research domain by combining principled Bayesian learning for deep neural networks. A review of current BNNs has been provided with a focus on mostly, how the posterior $p(\theta\vert x,y)$ is inferred. As an observation, many of the recent breakthroughs have been achieved by performing approximate Bayesian inference in a mini-batch fashion (stochastically) or investigating relatively simple but scalable techniques such as MC-dropout or Laplace Approximation. As a result, several works demonstrated that the posterior inference in large scale settings are now possible \cite{maddox2019simple, osawa2019practical, lee2020estimating}, and the field has several practical approximation tools to compute more expressive and accurate posteriors since the revival of BNNs beyond the pioneers \cite{hinton1993keeping, barber1998ensemble, neal1992bayesian, denker1991transforming, mackay1992practical}. There are also emerging challenges on new frontiers beyond accurate inference techniques. Some examples are: (i) how to specify meaningful priors? \cite{ito2005bayesian, sun2018functional}, (ii) how to efficiently marginalize over the parameters for fast predictive uncertainty? \cite{balan2015bayesian, postels2019sampling, hobbhahn2020fast} (iii) infrastructures such as new benchmarks, evaluation protocols and software tools  \cite{mukhoti2018importance, tran2016deep, bingham2019pyro, filos2019systematic}, and (iv) towards better understandings on the current methodologies and their potential applications \cite{farquhar2020try, wenzel2020good, mukhoti2018evaluating, feng2019introspective}.
\subsection{\textbf{Ensemble Methods}}\label{sec:ensemble.methods}
\subsubsection{Principles of Ensemble Methods}
Ensembles derive a prediction based on the predictions received from multiple so-called ensemble members.
They target at a better generalization by making use of synergy effects among the different models, arguing that a group of decision makers tend to make better decisions than a single decision maker \cite{ensemble.survey,hansen1990neural}. For an ensemble $f:X \rightarrow Y$ with members $f_i:X \rightarrow Y$ for $i \in {1,2,...,M}$, this could be for example implemented by simply averaging over the members' predictions, 
\begin{equation*}
    f(x):=\frac{1}{M} \sum_{i=1}^M f_i(x)~.
\end{equation*}
Based on this intuitive idea, several works applying ensemble methods to different kinds of practical tasks and approaches, as for example bioinformatics \cite{cao2020ensemble,nannia2020ensemble,wei2017novel}, remote sensing \cite{lv2017remote,dai2019semisupervised,marushko2020methods}, or reinforcement learning \cite{kurutach2018model,rajeswaran2016epopt} can be found in the literature. Besides the improvement in the accuracy, ensembles give an intuitive way of representing the model uncertainty on a prediction by evaluating the variety among the member's predictions. \\
Compared to Bayesian and single deterministic network approaches, ensemble methods have two major differences. First, the general idea behind ensembles is relatively clear and there are not many groundbreaking differences in the application of different types of ensemble methods and their application in different fields. Hence, this section focuses on different strategies to train an ensemble and some variations that target on making ensemble methods more efficient. Second, ensemble methods were originally not introduced to explicitly handle and quantify uncertainties in neural networks. Although the derivation of uncertainty from ensemble predictions is obvious, since they actually aim at reducing the model uncertainty, ensembles were first introduced and discussed in order to improve the accuracy on a prediction \cite{hansen1990neural}. Therefore, many works on ensemble methods do not explicitly take the uncertainty into account. Notwithstanding this, ensembles have been found to be well suited for uncertainty estimations in neural networks \cite{deep.ensembles}.\\
\subsubsection{Single- and Multi-Mode Evaluation}
One main point where ensemble methods differ from the other methods presented in this paper is the number of local optima that are considered, i.e. the differentiation into \textit{single-mode} and \textit{multi-mode} evaluation. \\
In order to create synergies and marginalise false predictions of single members, the members of an ensemble have to behave differently in case of an uncertain outcome. The mapping defined by a neural network is highly non-linear and hence the optimized loss function contains many local optima to which a training algorithm could converge to. Deterministic neural networks converge to one single local optimum in the solution space \cite{fort2019deep}. Other approaches, e.g. BNNs, still converge to one single optimum, but additionally take the uncertainty on this local optimum into account \cite{fort2019deep}. This means, that neighbouring points within a certain region around the solution also affect the loss and also influence the prediction of a test sample. Since these methods focus on single regions, the evaluation is called \textit{single-mode} evaluation. In contrast to this, ensemble methods consist of several networks, which should converge to different local optima. This leads to a so called multi-mode evaluation \cite{fort2019deep}. 
\begin{figure}[t]
\resizebox{0.5\textwidth}{!}{
   \includegraphics[trim=75 450 50 50, clip]{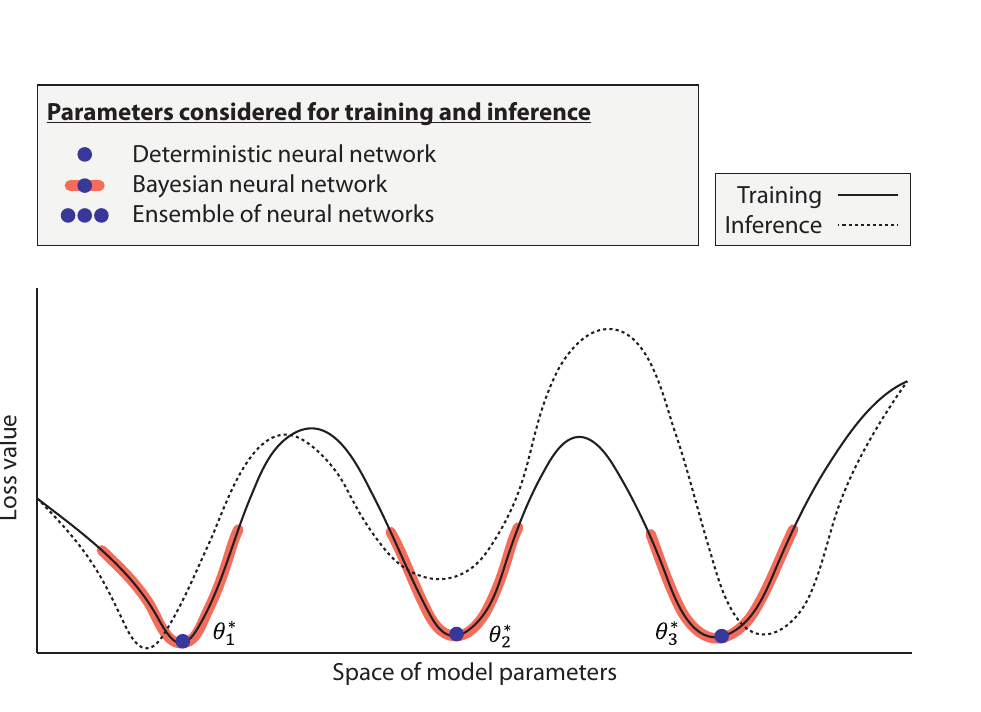}}
    \caption{A visualization of the different evaluation behaviours of deterministic neural networks, Bayesian neural networks and the ensemble of deterministic neural networks. The $x$-axis indicates the network parameters $\theta$ and the $y$-axis represents the loss value. While the deterministic network learns the parameters based on a pointwise estimation, the Bayesian neural network also takes the surrounding of the single point into account. The ensemble of deterministic methods optimizes pointwise but learns several different parameter settings.} 
\label{fig:loss_landscape}
\end{figure}

In Figure \ref{fig:loss_landscape}, the considered parameters of a single-mode deterministic, single-mode probabilistic (Bayesian) and multi-mode ensemble approach are visualized. The goal of multi-mode evaluation is that different local optima could lead to models with different strengths and weaknesses in the predictions such that a combination of several such models brings synergy effects improving the overall performance. \\
\subsubsection{Bringing Variety into Ensembles}
One of the most crucial points when applying ensemble methods is to maximize the variety in the behaviour among the single networks \cite{renda2019comparing,deep.ensembles}. In order to increase the variety, several different approaches can be applied:
\begin{itemize}
  \setlength\itemsep{0.5em}
    \item \textbf{Random Initialization and Data Shuffle} \\
    Due to the very non-linear loss landscape, different initializations of a neural network lead in general to different training results. Since the training is realized on mini-batches, the order of the training data points also affects the final result. 
    \item \textbf{Bagging and Boosting}\\
    Bagging (\textbf{B}ootstrap \textbf{agg}regat\textbf{ing}) and Boosting are two strategies that vary the distribution of the used training data sets by sampling new sets of training samples from the original set. Bagging is sampling from the training data uniformly and with replacement \cite{bishop2006pattern}. Thanks to the replacement process, ensemble members can see single samples several times in the training set while missing some other training samples. For boosting, the members are trained one after another and the probability of sampling a sample for the next training set is based on the performance of the already trained ensemble \cite{bishop2006pattern}. 
    \item \textbf{Data Augmentation}\\
    Augmenting the input data randomly for each ensemble member leads to models trained on different data points and therefore in general to a larger variety among the different members. 
    \item \textbf{Ensemble of different Network Architecture}\\
    The combination of different network architectures leads to different loss landscapes and can therefore also increase the diversity in the resulting predictions \cite{herron2020ensembles}.
    
\end{itemize}
In several works, it has been shown that the variety induced by random initialization works sufficiently and that bagging could even lead to a weaker performance \cite{lee2015m,deep.ensembles}. Livieris et al. \cite{livieris2020ensemble} evaluated different bagging and boosting strategies for ensembles of weight constrained neural networks. Interestingly, it is found that bagging performs better for a small number of ensemble members while boosting performs better for a large number.
Nanni et al. \cite{augmentation_ensemble} evaluated ensembles based on different types of image augmentation for bioimage classification tasks and compared those to each other. Guo and Gould \cite{guo2015deep} used augmentation methods within in an ensemble approach for object detection. Both works stated that the ensemble approach using augmentations improves the resulting accuracy. In contrast to this, \cite{rahaman2020uncertainty,wen2020combining} stated with respect to uncertainty quantification that image augmentation can harm the calibration of an ensemble and post-processing calibration methods have to be slightly adapted when using ensemble methods. Other ways of inducing variety for specific tasks have been also introduced. For instance, in \cite{Kim_2018_ECCV}, the members are trained with different attention masks in order to focus on different parts of the input data. Other approaches focused on the training process and introduced learning rate schedulers that are designed to discover several local optima within one training process \cite{huang2017snapshot,yang2020autoensemble}. Following, an ensemble can be built based on local optima found within one single training run. It is important to note that if not explicitly stated, the works and approaches presented so far targeted on improvements in the predictive accuracy and did not explicitly consider uncertainty quantification. \\
\subsubsection{Ensemble Methods and Uncertainty Quantification}
Besides the improvement in the accuracy, ensembles are widely used for modelling uncertainty on predictions of complex models, as for example in climate prediction \cite{leutbecher2008ensemble,parker2013ensemble}. Accordingly, ensembles are also used for quantifying the uncertainty on a deep neural network's prediction, and over the last years they became more and more popular for such tasks \cite{deep.ensembles,renda2019comparing}. Lakshminarayanan et al. \cite{deep.ensembles} are often referenced as a base work on uncertainty estimations derived from ensembles of neural networks and as a reference for the competitiveness of deep ensembles. They introduced an ensemble training pipeline to quantify predictive uncertainty within DNNs. In order to handle data and model uncertainty, the member networks are designed with two heads, representing the prediction and a predicted value of data uncertainty on the prediction. The approach is evaluated with respect to accuracy, calibration, and out-of-distribution detection for classification and regression tasks. In all tests, the method performs at least equally well as the BNN approaches used for comparison, namely Monte Carlo Dropout and Probabilistic Backpropagation. Lakshminarayanan et al. \cite{deep.ensembles} also showed that shuffling the training data and a random initialization of the training process induces a sufficient variety in the models in order to predict the uncertainty for the given architectures and data sets. Furthermore, bagging is even found to worsen the predictive uncertainty estimation, extending the findings of Lee et al. \cite{lee2015m}, who found bagging to worsen the predictive accuracy of ensemble methods on the investigated tasks. Gustafsson et al. \cite{gustafsson2020evaluating} introduced a framework for the comparison of uncertainty quantification methods with a specific focus on real life applications. Based on this framework, they compared ensembles and Monte Carlo dropout and found ensembles to be more reliable and applicable to real life applications. These findings endorse the results reported by Beluch et al. \cite{beluch2018power} who found ensemble methods to deliver more accurate and better calibrated predictions on active learning tasks than Monte Carlo Dropout. Ovadia et al. \cite{ovadia2019can} evaluated different uncertainty quantification methods based on test sets affected by distribution shifts. The excessive evaluation contains a variety of model types and data modalities. As a take away, the authors stated that already for a relatively small ensemble size of five, deep ensembles seem to perform best and are more robust to data set shifts than the compared methods. Vyas et al. \cite{vyas2018out} presented an ensemble method for the improved detection of out-of-distribution samples. For each member, a subset of the training data is considered as out-of-distribution. For the training process, a loss, seeking a minimum margin greater zero between the average entropy of the in-domain and the out-of-distribution subsets is introduced and leads to a significant improvement in the out-of-distribution detection. \\
\subsubsection{Making Ensemble Methods more Efficient}
Compared to single model methods, ensemble methods come along with a significantly increased computational effort and memory consumption \cite{ensemble.survey,distribution.distillation}. When deploying an ensemble for a real life application the available memory and computational power are often limited. Such limitations could easily become a bottleneck \cite{kocic2019end} and could become critical for applications with limited reaction time. Reducing the number of models leads to less memory and computational power consumption.
\textit{Pruning approaches} reduce the complexity of ensembles by pruning over the members and reducing the redundancy among them. For that, several approaches based on different diversity measures are developed to remove single members without strongly affecting the performance \cite{guo2018margin,cavalcanti2016combining,martinez2008analysis}. 

Distillation is another approach where the number of networks is reduced to one single model. It is the procedure of teaching a single network to represent the knowledge of a group of neural networks \cite{bucilua2006model}. First works on the distillation of neural networks were motivated by restrictions when deploying large scale classification problems \cite{bucilua2006model}. The original classification problem is separated into several sub-problems focusing on single blocks of classes that are difficult to differentiate. Several smaller trainer networks are trained on the sub-problems and then teach one student network to separate all classes at the same time. In contrast to this, \textit{Ensemble distillation approaches} capture the behaviour of an ensemble by one single network. First works on ensemble distillation used the average of the softmax outputs of the ensemble members in order to teach a student network the derived predictive uncertainty \cite{hinton2015distilling}. Englesson and Azizpour \cite{englesson2019efficient} justify the resulting predictive distributions of this approach and additionally cover the handling of out-of-distribution samples. When averaging over the members' outputs, the model uncertainty, which is represented in the variety of ensemble outputs, gets lost. To overcome this drawback, researchers applied the idea of learning higher order distributions, i.e. distributions over a distribution, instead of directly predicting the output \cite{distribution.distillation.general.framework,distribution.distillation}. The members are then distillated based on the divergence from the average distribution. The idea is closely related to the prior networks \cite{prior.network} and the evidential neural networks \cite{evidential.neural.networks}, which are described in Section \ref{sec:deterministic_methods}. \cite{distribution.distillation} modelled ensemble members and the distilled network as prior networks predicting the parameters of a Dirichlet distribution. The distillation then seeks to minimize the KL divergence between the averaged Dirichlet distributions of the ensemble members and the output of the distilled network. Lindqvist et al. \cite{distribution.distillation.general.framework} generalized this idea to any other parameterizable distribution. With that, the method is also applicable to regression problems, for example by predicting a mean and standard deviation to describe a normal distribution. Within several tests, the distillation models generated by these approaches are able to distinguish between data uncertainty and model uncertainty. Although distillation methods cannot completely capture the behaviour of an underlying ensemble, it has been shown that they are capable of delivering good and for some experiments even comparable results \cite{distribution.distillation.general.framework,distribution.distillation,reich2020ensemble}. \\
Other approaches, as \textit{sub-ensembles} \cite{sub.ensembles} and \textit{batch-ensembles} \cite{batch.ensembles} seek to reduce the computation effort and memory consumption by sharing parts among the single members. It is important to note that the possibility of using different model architectures for the ensemble members could get lost when parts of the ensembles are shared. Also, the training of the models cannot be run in a completely independent manner. Therefore, the actual time needed for training does not necessarily decrease in the same way as the computational effort does. \\
Sub-ensembles \cite{sub.ensembles} divide a neural network architecture into two sub-networks. The trunk network for the extraction of general information from the input data, and the task network that uses these information to fulfill the actual task. In order to train a sub-ensemble, first, the weights of each member's trunk network are fixed based on the resulting parameters of one single model's training process. Following, the parameters of each ensemble members' task network are trained independently from the other members. As a result, the members are built with a common trunk and an individual task sub-network. Since the training and the evaluation of the trunk network have to be done only once, the number of computations needed for training and testing decreases by the factor $\frac{M \cdot N_{\text{task}}  + N_{\text{trunk}}}{M\cdot N}$, where $N_{\text{task}}$, $N_{\text{trunk}}$, and $N$ stand for the number of variables in the task networks, the trunk network, and the complete network. Valdenegro-Toro \cite{sub.ensembles} further underlined the usage of a shared trunk network by arguing that the trunk network is in general computational more costly than the task network. 
In contrast to this, batch-ensembles \cite{batch.ensembles} connect the member networks with each other at every layer. The ensemble members' weights are described as a Hadamard product of one shared weight matrix $W \in \R^{n\times m}$ and $M$ individual rank one matrices $F_i \in \R^{n \times m}$, each linked with one of the $M$ ensemble members. The rank one matrices can be written as a multiplication $F_i=r_is_i^\text{T}$ of two vectors $s\in \R^{n}$ and $r\in \R^{m}$ and hence the matrix $F_i$ can be described by $n+m$ parameters. With this approach, each additional ensemble member increases the number of parameters only by the factor $\frac{n+m}{M\cdot(n+m)+n\cdot m} + 1$ instead of $\frac{M+1}{M}=1 + \frac{1}{M}$. On the one hand, with this approach, the members are not independent anymore such that all the members have to be trained in parallel. On the other hand, the authors also showed that the parallelization can be realized similar to the optimization on mini-batches and on a single unit.\\
\subsubsection{Sum Up Ensemble Methods}
Ensemble methods are very easy to apply, since no complex implementation or major modification of the standard deterministic model have to be realized. Furthermore, ensemble members are trained independently from each other, which makes the training easily parallelizable. Also, trained ensembles can be extended easily, but the needed memory and the computational effort increases linearly with the number of members for training and evaluation. The main challenge when working with ensemble methods is the need of introducing diversity among the ensemble members. For accuracy, uncertainty quantification, and out-of-distribution detection, random initialization, data shuffling, and augmentations have been found to be sufficient for many applications and tasks \cite{deep.ensembles,augmentation_ensemble}. Since these methods may be applied anyway, they do not need much additional effort. The independence of the single ensemble members leads to a linear increase in the required memory and computation power with each additional member. This holds for the training as well as for testing. This limits the deployment of ensemble methods in many practical applications where the computation power or memory is limited, the application is time-critical, or very large networks with high inference time are included \cite{distribution.distillation}.\\
Many aspects of ensemble approaches are only investigated with respect to the performance on the predictive accuracy but do not take predictive uncertainty into account. This also holds for the comparison of different training strategies for a broad range of problems and data sets. 
Especially since the overconfidence from single members can be transferred to the whole ensemble, strategies that encourage the members to deliver different false predictions instead of all delivering the same false prediction should be further investigated. For a better understanding of ensemble behavior, further evaluations of the loss landscape, as done by Fort et al. \cite{fort2019deep}, could offer interesting insights.

\subsection{\textbf{Test Time Augmentation}}\label{sec:test_time_augmentation}
Inspired by ensemble methods and adversarial examples \cite{Ayhan.2018}, the test time data augmentation is one of the simpler predictive uncertainty estimation techniques. The basic method is to create multiple test samples from each test sample by applying data augmentation techniques on it and then test all those samples to compute a predictive distribution in order to measure uncertainty. The idea behind this method is that the augmented test samples allow the exploration of different views and is therefore capable of capturing the uncertainty. Mostly, this technique of test time data augmentations has been used in medical image processing \cite{wang2018automatic,wang2019aleatoric,Ayhan.2018,moshkov2020test}. One of the reasons for this is that the field of medical image processing already makes heavy use of data augmentations while using deep learning \cite{ronneberger2015u}, so it is quite easy to just apply those same augmentations during test time to calculate the uncertainties. Another reason is that collecting medical images is costly, thus forcing practitioners to rely on data augmentation techniques. 
Moshkov et al. \cite{moshkov2020test} used the test time augmentation technique for cell segmentation tasks. For that, they created multiple variations of the test data before feeding it to a trained UNet or Mask R-CNN architecture. Following, they used a majority voting to create the final output segmentation mask and discuss the policies of applying different augmentation techniques and how they affect the final predictive results of the deep networks.\\
Overall, test time augmentation is an easy method for estimating uncertainties because it keeps the underlying model unchanged, requires no additional data, and is simple to put into practice with off-the-shelf libraries. Nonetheless, it needs to be kept in mind that during applying this technique, one should only apply valid augmentations to the data, meaning that the augmentations should not generate data from outside the target distribution. According to \cite{shanmugam2020and}, test time augmentation can change many correct predictions into incorrect predictions (and vice versa) due to many factors such as the nature of the problem at hand, the size of training data, the deep neural network architecture, and the type of augmentation. To limit the impact of these factors, Shanmugam et al. \cite{shanmugam2020and} proposed a learning-based method for test time augmentation that takes these factors into consideration. In particular, the proposed method learns a function that aggregates the predictions from each augmentation of a test sample. Similar to \cite{shanmugam2020and}, Molchanov et al. \cite{molchanov2020greedy} proposed a method, named “greedy Policy Search”, for constructing a test-time augmentation policy by choosing augmentations to be include in a fixed-length policy. Similarly, Kim et al. \cite{kim2020learning} proposed a method for learning a loss predictor from the training data for instance-aware test-time augmentation selection. The predictor selects test-time augmentations with the lowest predicted loss for a given sample.

Although learnable test time augmentation techniques \cite{shanmugam2020and, molchanov2020greedy, kim2020learning} help to select valid augmentations, one of the major open question is to find out the effect on uncertainty due to different kinds of augmentations. It can for example happen that a simple augmentation like reflection is not able to capture much of the uncertainty while some domain specialized stretching and shearing captures more uncertainty. It is also important to find out how many augmentations are needed to correctly quantify uncertainties in a given task. This is particularly important in applications like earth observation, where inference might be needed on global scale with limited resources.
\subsection{Neural Network Uncertainty Quantification Approaches for Real Life Applications}
In order to use the presented methods on real life tasks, several different considerations have to be taken into account. The memory and computational power is often restricted while many real world tasks my be time-critical \cite{kocic2019end}. An overview over the main properties is given in Table \ref{tab:types_compare}. \\
The presented applications all come along with advantages and disadvantages, depending on the properties a user is interested in. While ensemble methods and test-time augmentation methods are relatively easy to apply, Bayesian approaches deliver a clear description of the uncertainty on the models parameters and also deliver a deeper theoretical basis. The computational effort and memory consumption is a common restriction on real life applications, where single deterministic network approaches perform best, but distillation of ensembles or efficient Bayesian methods can also be taken into consideration. Within the different types of Bayesian approaches, the performance, the computational effort, and the implementation effort still vary strongly. Laplace approximations are relatively easy to apply and compared to sampling approaches much less computational effort is needed. Furthermore, there often already exist pretrained networks for an application. In this case, Laplace Approximation and external deterministic single network approaches can in general be applied to already trained networks. \\~\\
Another important aspect that has to be taken into account for uncertainty quantification in real life applications is the source and type of uncertainty. For real life applications, out-of-distribution detection forms the maybe most important challenge in order to avoid unexpected decisions of the network and to be aware of adversarial attacks. Especially since many motivations of uncertainty quantification are given by risk minimization, methods that deliver risk averse predictions are an important field to evaluate.
Many works already demonstrated the capability of detecting out-of-distribution samples on several tasks and built a strong fundamental tool set for the deployment in real life applications \cite{yu2019unsupervised,vyas2018out,ren2019likelihood,gustafsson2020evaluating}. However, in real life, the tasks are much more difficult than finding out-of-distribution samples among data sets (e.g., MNIST or CIFAR data sets etc.) and the main challenge lies in comparing such approaches on several real-world data sets against each other. The work of Gustafsson et al. \cite{gustafsson2020evaluating} forms a first important step towards an evaluation of methods that better suits the demands in real life applications. Interestingly, they found for their tests ensembles to outperform the considered Bayesian approaches. This indicates, that the multi-mode evaluation given by ensembles is a powerful property for real life applications. Nevertheless Bayesian approaches have delivered strong results as well and furthermore come along with a strong theoretical foundation \cite{lee2020estimating,hobbhahn2020fast,eggenreich2020variational,gal2017deep}. As a way to go, the combination of efficient ensemble strategies and Bayesian approaches could combine the variability in the model parameters while still considering several modes for a prediction. 
Also, single deterministic approaches as the prior networks \cite{prior.network,nandy2020towards,evidential.neural.networks,regularized.evidential.networks} deliver comparable results while consuming significantly less computation power. However, this efficiency often comes along with the problem that separated sets of in- and out-of-distribution samples have to be available for the training process \cite{regularized.evidential.networks,nandy2020towards}. 
In general, the development of new problem and loss formulations as for example given in \cite{nandy2020towards} leads to a better understanding and description of the underlying problem and forms an important field of research.

\section{Uncertainty Measures and Quality}\label{sec:uncertainty_measures}
In Section \ref{sec:uncertainty_quantification_methods}, we presented different methods for modeling and predicting different types of uncertainty in neural networks. In order to evaluate these approaches, measures have to be applied on the derived uncertainties. In the following, we present different measures for quantifying the different predicted types of uncertainty.
In general, the correctness and trustworthiness of these uncertainties is not automatically given. In fact, there are several reasons why evaluating the quality of the uncertainty estimates is a challenging task.
\begin{itemize}
  \setlength\itemsep{0.5em}
    \item First, the quality of the uncertainty estimation depends on the underlying method for estimating uncertainty. This is exemplified in the work undertaken by Yao et al. \cite{yao2019quality}, which shows that different approximates of Bayesian inference (e.g. Gaussian and Laplace approximates) result in different qualities of uncertainty estimates.
    \item Second, there is a lack of ground truth uncertainty estimates \cite{deep.ensembles} and defining ground truth uncertainty estimates is challenging. For instance, if we define the ground truth uncertainty as the uncertainty across human subjects, we still have to answer questions as "How many subjects do we need?" or "How to choose the subjects?".
    \item Third, there is a lack of a unified quantitative evaluation metric \cite{huang2019evaluating}. To be more specific, the uncertainty is defined differently in different machine learning tasks such as classification, segmentation, and regression. For instance, prediction intervals or standard deviations are used to represent uncertainty in regression tasks, while entropy (and other related measures) are used to capture uncertainty in classification and segmentation tasks.
\end{itemize}
\subsection{Evaluating Uncertainty in Classification Tasks}
For classification tasks, the network's softmax output already represents a measure of confidence. But since the raw softmax output is neither very reliable \cite{hendrycks2016baseline} nor can it represent all sources of uncertainty \cite{smith2018understanding}, further approaches and corresponding measures were developed. 
\subsubsection{Measuring Data Uncertainty in Classification Tasks}~\\
Consider a classification task with $K$ different classes and a probability vector network output $p(x)$ for some input sample $x$. In the following $p$ is used for simplification and $p_k$ stands for the $k$-th entry in the vector. In general, the given prediction $p$ represents a categorical distribution, i.e. it assigns a probability to each class to be the correct prediction. Since the prediction is not given as an explicit class but as a probability distribution, (un)certainty estimates can be directly derived from the prediction. In general this pointwise prediction can be seen as estimated data uncertainty \cite{kendall2017uncertainties}. However, as discussed in Section \ref{sec:uncertainty_types_and_sources}, the model's estimation of the data uncertainty is affected by model uncertainty, which has to be taken into account separately. In order to evaluate the amount of predicted data uncertainty, one can for example apply the maximal class probability or the entropy measures:
\begin{align}
    &\text{Maximal probability:} \quad &p_{\text{max}} &=\max\left\{p_k\right\}_{k=1}^K &\\[1em]
    &\text{Entropy:}  &\text{H}(p) &=-\sum_{k=1}^Kp_k\log_2(p_k) &
\end{align}
The maximal probability represents a direct representation of certainty, while entropy describes the average level of information in a random variable. Even though a softmax output should represent the data uncertainty, one cannot tell from a single prediction how large the amount of model uncertainty is that affects this specific prediction as well. 
\subsubsection{Measuring Model Uncertainty in Classification Tasks}~\\
As already discussed in Section \ref{sec:uncertainty_quantification_methods}, a single softmax prediction is not a very reliable way for uncertainty quantification since it is often badly calibrated \cite{smith2018understanding} and does not have any information about the certainty the model itself has on this specific output \cite{smith2018understanding}. An (approximated) posterior distribution $p(\theta \vert D)$ on the learned model parameters can help to receive better uncertainty estimates. With such a posterior distribution, the softmax output itself becomes a random variable and one can evaluate its variation, i.e. uncertainty. For simplicity, we denote $p(y\vert \theta, x)$ also as $p$ and it will be clear from context whether $p$ depends on $\theta$ or not. The most common measures for this are the mutual information (MI), the expected Kullback-Leibler Divergence (EKL), and the predictive variance. Basically, all these measures compute the expected divergence between the (stochastic) softmax output and the expected softmax output
\begin{align}
    \hat{p} = \mathbb{E}_{\theta\sim p(\theta\vert D)}\left[p(y\vert x, \theta\right]~.
\end{align}
The MI uses the entropy to measure the mutual dependence between two variables. In the described case, the difference between the information given in the expected softmax output and the expected information in the softmax output is compared, i.e.
\begin{align}\label{eq:MI}
    \text{MI}\left(\theta, y \vert x, D\right) = \text{H}\left[\hat{p}\right] - \mathbb{E}_{\theta\sim p(\theta\vert D)}\text{H}\left[p(y \vert x, \theta )\right]~.
\end{align}
Smith and Gal \cite{smith2018understanding} pointed out that the MI is minimal when the knowledge about model parameters does not increase the information in the final prediction. Therefore, the MI can be interpreted as a measure of model uncertainty. \\
The Kullback-Leibler divergence measures the divergence between two given probability distributions. The EKL can be used to measure the (expected) divergence among the possible softmax outputs,
\begin{align}\label{eq:EKL}
    \mathbb{E}_{\theta\sim p(\theta \vert D)}\left[KL(\hat{p}~||~p)\right] =\mathbb{E}_{\theta\sim p(\theta \vert D)}\left[\sum_{i=1}^K \hat{p}_i \log\left(\frac{\hat{p}_i}{p_i}\right)\right]~,
\end{align}
which can also be interpreted as a measure of uncertainty on the model's output and therefore represents the model uncertainty. \\
The predictive variance evaluates the variance on the (random) softmax outpus, i.e.
\begin{align}\label{eq:pred_sigma}
    \sigma(p) &= \mathbb{E}_{\theta\sim p(\theta\vert D)} \left[\left(p - \hat{p}  \right)^2\right]~.
\end{align}
As described in Section \ref{sec:uncertainty_quantification_methods}, an analytically described posterior distribution $p(\theta\vert D)$ is only given for a subset of the Bayesian methods. And even for an analytically described distribution, the propagation of the parameter uncertainty into the prediction is in almost all cases intractable and has to be approximated for example with Monte Carlo approximation. Similarly, ensemble methods collect predictions from $M$ neural networks, and test-time data augmentation approaches receive $M$ predictions from $M$ different augmentations applied to the original input sample. For all these cases, we receive a set of $M$ samples, $\left\{p^i\right\}_{i=1}^M$, which can be used to approximate the intractable or even undefined underlying distribution. With these approximations, the measures defined in \eqref{eq:MI}, \eqref{eq:EKL}, and \eqref{eq:pred_sigma} can be applied straight forward and only the expectation has to be replaced by average sums. For example, the expected softmax output becomes 
\begin{align*}
    \hat{p} \approx \frac{1}{M}\sum_{i=1}^M p^i~.
\end{align*}
For the expectations given in \eqref{eq:MI}, \eqref{eq:EKL}, and \eqref{eq:pred_sigma}, the expectation is approximated similarly. 
\subsubsection{Measuring Distributional Uncertainty in Classification Tasks}~\\
Although these uncertainty measures are widely used to capture the variability among several predictions derived from Bayesian neural networks \cite{kendall2017uncertainties}, ensemble methods \cite{deep.ensembles}, or test-time data augmentation methods \cite{Ayhan.2018}, they cannot capture distributional shifts in the input data or out-of-distribution examples, which could lead to a biased inference process and a falsely stated confidence. If all predictors attribute a high probability mass to the same (false) class label, this induces a low variability among the estimates. Hence, the network seams to be certain about its prediction, while the uncertainty in the prediction itself (given by the softmax probabilities) is also evaluated to be low. To tackle this issue, several approaches described in Section \ref{sec:uncertainty_quantification_methods} take the magnitude of the logits into account, since a larger logit indicates larger evidence for the corresponding class \cite{evidential.neural.networks}. Thus, the methods either interpret the total sum of the (exponentials of) the logits as precision value of a Dirichlet distribution (see description of Dirichlet Priors in Section \ref{sec:deterministic_methods}) \cite{prior.network,malinin2019reverse,nandy2020towards}, or as a collection of evidence that is compared to a defined constant \cite{evidential.neural.networks,inhibited.softmax}. One can also derive a total class probability for each class individually by applying the sigmoid function to each logit \cite{hsu2020generalized}. Based on the class-wise total probabilities, OOD samples might easier be detected, since all classes can have low probability at the same time. Other methods deliver an explicit measure how well new data samples suit into the training data distribution. Based on this, they also give a measure that a sample will be predicted correctly \cite{density.estimation.in.representation.space}. 
\subsubsection{Performance Measure on Complete Data Set}~\\
While the measures described above measure the performance of individual predictions, others evaluate the usage of these measures on a set of samples. Measures of uncertainty can be used to separate between correctly and falsely classified samples or between in-domain and out-of-distribution samples \cite{hendrycks2016baseline}. For that, the samples are split into two sets, for example in-domain and out-of-distribution or correctly classified and falsely classified. The two most common approaches are the \textit{Receiver Operating Characteristic} (ROC) curve and the \textit{Precision-Recall} (PR) curve. Both methods generate curves based on different thresholds of the underlying measure. For each considered threshold, the ROC curve plots the true positive rate against the false positive rate\footnote{The true positive rate is the number of samples, which are correctly predicted as positive divided by the total number of true samples. The false positive rate is the number of samples falsely predicted as positive divided by the total number of negative samples (see also \cite{davis2006relationship})}, and the PR curve plots the precision against the recall\footnote{The precision is equal to the number of samples that are correctly classified as positive, divided by the total number of positive samples. The recall is equal to the number of samples correctly predicted as positive divided by the total number of positive samples (see also \cite{davis2006relationship})}. While the ROC and PR curves give a visual idea of how well the underlying measures are suited to separate the two considered test cases, they do not give a qualitative measure. To reach this, the area under the curve (AUC) can be evaluated. Roughly speaking, the AUC gives a probability value that a randomly chosen positive sample leads to a higher measure than a randomly chosen negative example. For example, the maximum softmax values measure ranks of correctly classified examples higher than falsely classified examples. Hendrycks and Gimpel \cite{hendrycks2016baseline} showed for several application fields that correct predictions have in general a higher predicted certainty in the softmax value than false predictions. Especially for the evaluation of in-domain and out-of-distribution examples, the \textit{Area Under Receiver Operating Curve} (AUROC) and the \textit{Area Under Precision Recall Curce} (AUPRC) are commonly used \cite{nandy2020towards,prior.network,malinin2019reverse}. The clear weakness of these evaluations is the fact that the performance is evaluated and the optimal threshold is computed based on a given test data set. A distribution shift from the test set distribution can ruin the whole performance and make the derived thresholds impractical.
\subsection{Evaluating Uncertainty in Regression Tasks}
\subsubsection{Measuring Data Uncertainty in Regression Predictions}
In contrast to classification tasks, where the network typically outputs a probability distritution over the possible classes, regression tasks only predict a pointwise estimation without any hint of data uncertainty. As already described in Section \ref{sec:uncertainty_quantification_methods}, a common approach to overcome this is to let the network predict the parameters of a probability distribution, for example a mean vector and a standard deviation for a normally distributed uncertainty \cite{deep.ensembles,kendall2017uncertainties}. Doing so, a measure of data uncertainty is directly given. The prediction of the standard deviation allows an analytical description that the (unknown) true value is within a specific region. The interval that covers the true value with a probability of $\alpha$ (under the assumption that the predicted distribution is correct) is given by  
\begin{equation}
    \left[\hat{y}-\frac{1}{2}\Phi^{-1}(\alpha)\cdot\sigma;\quad \hat{y}+\frac{1}{2}\Phi^{-1}(\alpha)\cdot\sigma\right]
\end{equation}
where $\Phi^{-1}$ is the quantile function, the inverse of the cumulative probability function. For a given probability value $\alpha$ the quantile function gives a boundary, such that $100\cdot\alpha\%$ of a standard normal distribution's probability mass is on values smaller than $\Phi^{-1}(\alpha)$. Quantiles assume some probability distribution and interpret the given prediction as the expected value of the distribution. \\
In contrast to this, other approaches \cite{pearce2018highquality,su2018tight} directly predict a so called prediction interval (PI) 
\begin{align}
    PI(x) = \left[B_l, B_u\right]
\end{align}
in which the prediction is assumed to lay. Such intervals induce an uncertainty as a uniform distribution without giving a concrete prediction. The certainty of such approaches can, as the name indicates, be directly measured by the size of the predicted interval. The \textit{Mean Prediction Interval Width} (MPIW) can be used to evaluate the average certainty of the model \cite{pearce2018highquality,su2018tight}. In order to evaluate the correctness of the predicted intervals the \textit{Prediction Interval Coverage Probability} (PICP) can be applied \cite{pearce2018highquality,su2018tight}. The PCIP represents the percentage of test predictions that fall into a prediction interval and is defined as 
\begin{equation}\label{eq:picp}
    \text{PICP}=\frac{c}{n}~,
\end{equation}
where $n$ is the total number of predictions and $c$ the number of ground truth values that are actually captured by the predicted intervals.
\subsubsection{Measuring Model Uncertainty in Regression Predictions}~\\
In Section \ref{sec:uncertainty_types_and_sources}, it is described, that model uncertainty is mainly caused by the model's architecture, the training process, and underrepresented areas in the training data. Hence, there is no real difference in the causes and effects of model uncertainty between regression and classification tasks such that model uncertainty in regression tasks can be measured equivalently as already described for classification tasks, i.e. in most cases by approximating an average prediction and measuring the divergence among the single predictions \cite{kendall2017uncertainties}.

\subsection{Evaluating Uncertainty in Segmentation Tasks}
The evaluation of uncertainties in segmentation tasks is very similar to the evaluation for classification problems. The uncertainty is estimated in segmentation tasks using approximates of Bayesian inference \cite{nair2020exploring,roy2019bayesian,labonte2019we,eaton2018towards,mcclure2019knowing,soleimany2019image,soberanis2020uncertainty,Seebock_2020} or test-time data augmentation techniques \cite{wang2019aleatoric}. In the context of segmentation, the uncertainty in pixel wise segmentation is measured using confidence intervals \cite{labonte2019we,eaton2018towards}, the predictive variance \cite{soleimany2019image,Seebock_2020}, the predictive entropy \cite{roy2019bayesian,wang2019aleatoric,mcclure2019knowing,soberanis2020uncertainty} or the mutual information \cite{nair2020exploring}. The uncertainty in structure (volume) estimation is obtained by averaging over all pixel-wise uncertainty estimates \cite{Seebock_2020,mcclure2019knowing}. The quality of volume uncertainties is assessed by evaluating the coefficient of variation, the average Dice score or the intersection over union \cite{roy2019bayesian,wang2019aleatoric}. These metrics measure the agreement in area overlap between multiple estimates in a pair-wise fashion.
Ideally, a false segmentation should result in an increase in pixel-wise and structure uncertainty. To evaluate whether this is the case, Nair et al. \cite{nair2020exploring} evaluated the pixel-level true positive rate and false detection rate as well as the ROC curves for the retained pixels at different uncertainty thresholds. Similar to \cite{nair2020exploring}, McClure et al. \cite{mcclure2019knowing} also analyzed the area under the ROC curve.

\section{Calibration}\label{sec:calibration}
A predictor is called well-calibrated if the derived predictive confidence represents a good approximation of the actual probability of correctness \cite{guo2017calibration}. Therefore, in order to make use of uncertainty quantification methods, one has to be sure that the network is well calibrated. Formally, for classification tasks a neural network $f_\theta$ is calibrated \cite{kuleshov2018accurate} if it holds that 
\begin{align}\label{eq:calibration_classification}
    \forall p \in [0,1]:\quad \sum_{i=1}^N \sum_{k=1}^K\frac{y_{i,k}\cdot\mathbb{I}\{f_\theta(x_i)_k=p\}}{\mathbb{I}\{f_\theta(x_i)_k=p\}} \xrightarrow[]{N \to \infty} p~.
\end{align}
Here, $\mathbb{I}\{\cdot\}$ is the indicator function that is either 1 if the condition is true or 0 if it is false and $y_{i,k}$ is the $k$-th entry in the one-hot encoded groundtruth vector of a training sample $(x_i,y_i)$. This formulation means that for example $30\%$ of all predictions with a predictive confidence of $70\%$ should actually be false.
For regression tasks the calibration can be defined such that predicted confidence intervals should match the confidence intervals empirically computed from the data set \cite{kuleshov2018accurate}, i.e.
\begin{equation}\label{eq:calibration_regression}
\forall p \in [0,1]:\quad \sum_{i=1}^N\frac{\mathbb{I}\left\{y_i\in \text{conf}_{p}(f_\theta(x_i))\right\}}{N} \xrightarrow[]{N \to \infty} p,
\end{equation}
where $\text{conf}_p$ is the confidence interval that covers $p$ percent of a distribution. 

\begin{figure*}[t]
    \begin{center}
        \begin{subfigure}{0.25\textwidth}\includegraphics[width=\textwidth]{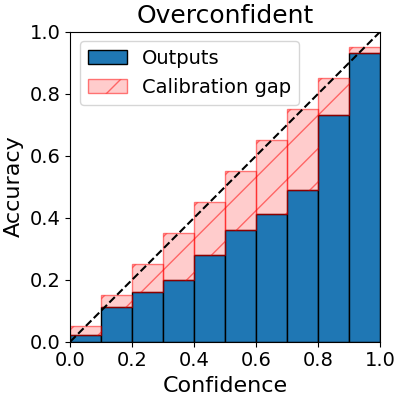}\subcaption[]{Underconfidence}\end{subfigure}
        \begin{subfigure}{0.25\textwidth}\includegraphics[width=\textwidth]{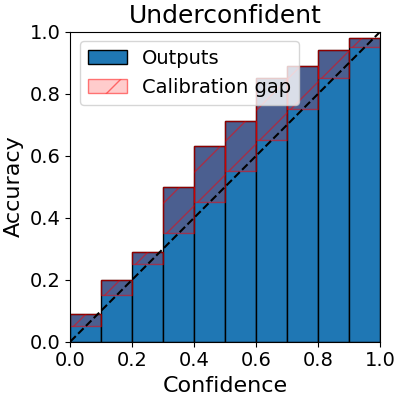}\subcaption[]{Overconfidence}\end{subfigure}
        \begin{subfigure}{0.25\textwidth}\includegraphics[width=\textwidth]{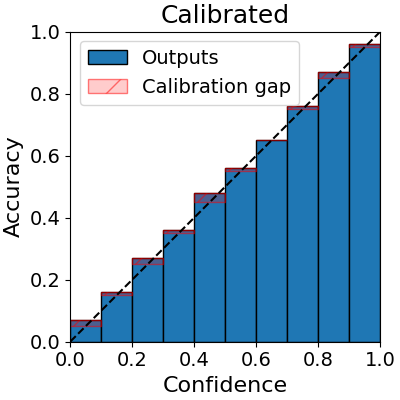}\subcaption[]{Calibrated classifier}\end{subfigure}
    \end{center}
    \caption{(a) Reliability diagram showing an overconfident classifier: The bin-wise accuracy is smaller than the corresponding confidence. (b) Reliability diagram of an underconfident classifier: The bin-wise accuracy is larger than the corresponding confidence. (c) Reliability diagram of a well calibrated classifier: The confidence fits the actual accuracy for the single bins.}
    \label{fig:reliability_diagram}
\end{figure*}

A DNN is called under-confident if the left hand side of \eqref{eq:calibration_classification} and \eqref{eq:calibration_regression} are larger than $p$. Equivalently, it is under-confident if the terms are smaller than $p$. The calibration property of a DNN can be visualized using a \textit{reliability diagram}, as shown in Figure \ref{fig:reliability_diagram}.

In general, calibration errors are caused by factors related to model uncertainty \cite{guo2017calibration}. This is intuitively clear, since as discussed in Section \ref{sec:uncertainty_types_and_sources}, data uncertainty represents the underlying uncertainty that an input $x$ and a target $y$ represent the same real world information. Following, correctly predicted data uncertainty would lead to a perfectly calibrated neural network.
In practice, several works pointed out that deeper networks tend to be more overconfident than shallower ones \cite{guo2017calibration,seo2019learning,li2020improving}. \\
Several methods for uncertainty estimation presented in Section \ref{sec:uncertainty_quantification_methods} also improve the networks calibration \cite{deep.ensembles,gal2016dropout}. This is clear, since these methods quantify model and data uncertainty separately and aim at reducing the model uncertainty on the predictions.
Besides the methods that improve the calibration by reducing the model uncertainty, a large and growing body of literature has investigated methods for explicitly reducing calibration errors. These methods are presented in the following, followed by measures to quantify the calibration error. It is important to note that these methods do not reduce the model uncertainty, but propagate the model uncertainty onto the representation of the data uncertainty. For example, if a binary classifier is overfitted and predicts all samples of a test set as class A with probability 1, while half of the test samples are actually class B, the recalibration methods might map the network output to 0.5 in order to have a reliable confidence. This probability of 0.5 is not equivalent to the data uncertainty but represents the model uncertainty propagated onto the predicted data uncertainty.

\subsection{Calibration Methods}
Calibration methods can be classified into three main groups according to the step when they are applied: 
\begin{itemize}
  \setlength\itemsep{0.5em}
    \item \textit{Regularization methods applied during the training phase} \cite{szegedy2016rethinking, pereyra2017regularizing, lee2017training, muller2019does, venkatesh2019heteroscedastic} \\
    These methods modify the objective, optimization and/or regularization procedure in order to build DNNs that are inherently calibrated.
    \item \textit{Post-processing methods applied after the training process of the DNN} \cite{guo2017calibration, wenger2020non}\\  
     These methods require a held-out calibration data set to adjust the prediction scores for recalibration. They only work under the assumption that the distribution of the left-out validation set is equivalent to the distribution, on which inference is done.
     Hence, also the size of the validation data set can influence the calibration result. 
    \item \textit{Neural network uncertainty estimation methods}\\
    Approaches, as presented in Section \ref{sec:uncertainty_quantification_methods}, that reduce the amount of model uncertainty on a neural network's confidence prediction, also lead to a better calibrated predictor. This is because the remaining predicted data uncertainty better represents the actual uncertainty on the prediction. Such methods are based for example on Bayesian methods \cite{izmailov2019subspace, foong2019between, zhang2019confidence, laves2019well, wilson2020bayesian} or deep ensembles \cite{deep.ensembles, mehrtash2020confidence}.
\end{itemize}
In the following, we present the three types of calibration methods in more detail. 
\begin{figure*}[h]
    \centering
    \resizebox{\textwidth}{!}{%
        \begin{tikzpicture}[framed, every annotation/.style = {draw,
            fill = white, font = \large}]
        \path[mindmap,concept color=black!40,text=black,
        every node/.style={concept,circular drop shadow},
        root/.style    = {concept color=black!40,
            font=\Large\bfseries,text width=10em},
        level 1 concept/.append style={font=\normalsize\bfseries, clockwise from=-15,
            sibling angle=75,text width=7em,
            level distance=15em,inner sep=0pt},
        level 2 concept/.append style={font=\small\bfseries,level distance=9em},
        ]
        
        node[root] {Neural Network Calibration Methods} [clockwise from=-15]
        
        child[concept color=orange] {
            node[concept] {Regularization Methods}
            [clockwise from=90]
            child { node[concept] (Objective Function Modification)
                {Objective Function Modification\textsuperscript{11}}}
            child { node[concept] (Data Augmentation)
                {Data Augmentation\textsuperscript{10}} }
            child { node[concept] (Label Smoothing)
                {Label Smoothing\textsuperscript{9}}}
            child { node[concept] (Exposure to OOD Examples)
                {Exposure to OOD examples\textsuperscript{8}}}
        }
        child[concept color=green!40!black] {
            node[concept] {Uncertainty Estimation Approaches}
            [clockwise from=-60]
            child { node[concept] (Bayesian Neural Networks)
                {Bayesian Neural Networks\textsuperscript{6}}}
            child { node[concept] (Ensemble of Neural Networks)
                {Ensemble of Neural Networks\textsuperscript{5}}}
        }
        child[concept color=blue!60] {
            node {Post-Processing Methods} [clockwise from=-90]
            child { node (temeprature_scaling){Temperature Scaling\textsuperscript{4}} }
            child { node (histogram_binning){Histogram Binning\textsuperscript{3}} }
            child { node (gaussian_process) {Gaussian Processes\textsuperscript{2}} }
            child { node (ensemble_of_post_processing_models) {Ensemble of Post-Processing Models\textsuperscript{1}} }
        };    
        \node[draw,text width=18cm, align=left] at (0,-10.5){
            \textsuperscript{1}~\cite{zhang2020mix}
            ~~~\textsuperscript{2}~\cite{wenger2020non}
            ~~~\textsuperscript{3}~\cite{zadrozny2001obtaining}
            ~~~\textsuperscript{4}\cite{guo2017calibration, liang2017enhancing, laves2019well}
            ~~~\textsuperscript{5}~\cite{deep.ensembles, mehrtash2020confidence}
            ~~~\textsuperscript{6}~\cite{izmailov2019subspace,foong2019between,zhang2019confidence,laves2019well,wilson2020bayesian}
            \textsuperscript{8}~\cite{hendrycks2018deep}
            ~~~\textsuperscript{9}~\cite{szegedy2016rethinking,muller2019does}
            ~~~\textsuperscript{10}~\cite{thulasidasan2019mixup, maronas2020improving, rahaman2020uncertainty, patel2019manifold}
            ~~~\textsuperscript{11}~\cite{pereyra2017regularizing, seo2019learning, lee2017training, venkatesh2019heteroscedastic, maronas2020improving}
        };
        \end{tikzpicture}
    }
    \caption{Visualization of the different types of uncertainty calibration methods presented in this paper.}
    \label{fig:calibration_diagram}
\end{figure*}
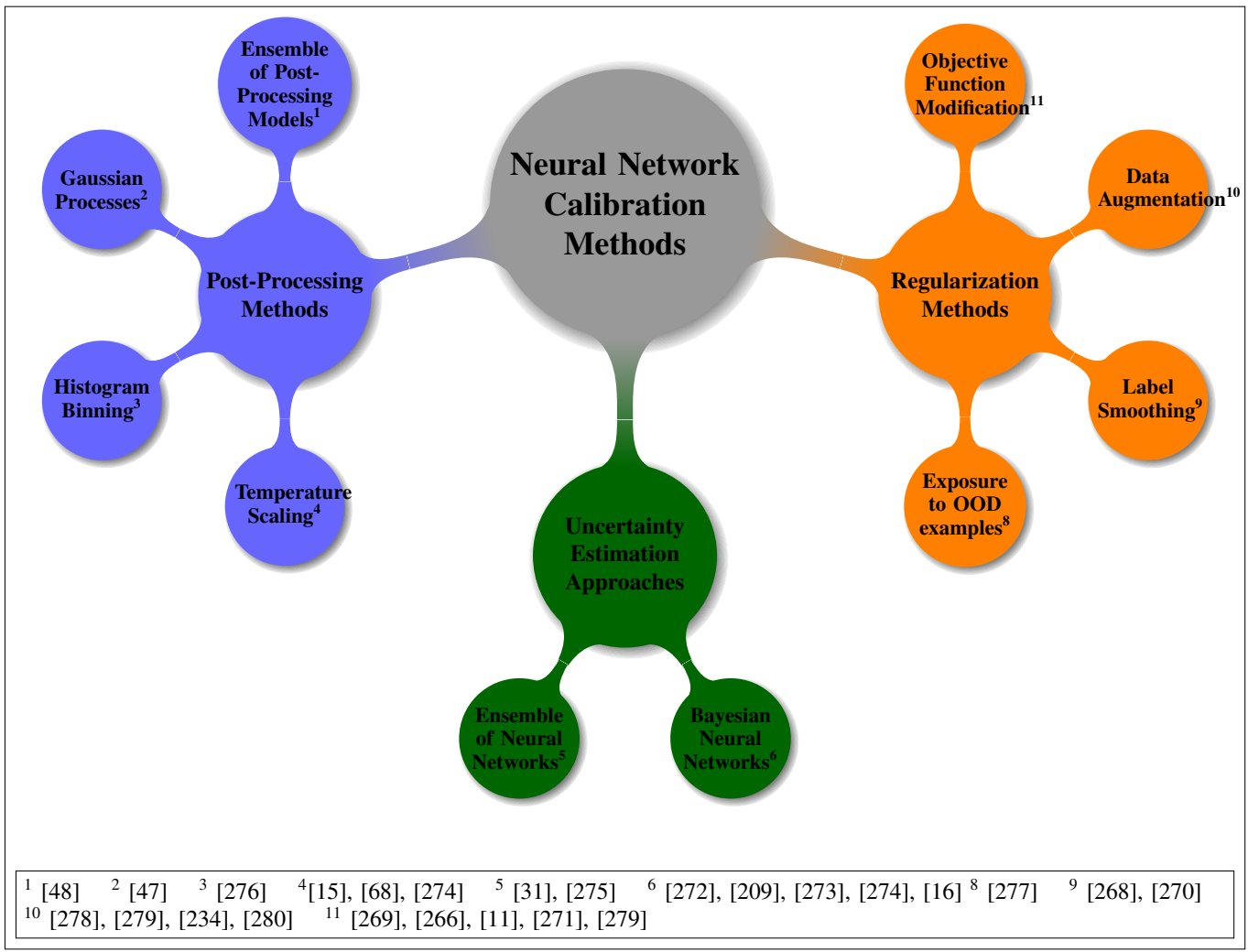

\subsubsection{Regularization Methods}
Regularization methods for calibrating confidences manipulate the training of DNNs by modifying the objective function or by augmenting the training data set. The goal and idea of regularization methods is very similar to the methods presented in Section \ref{sec:deterministic_methods} where the methods mainly quantify model and data uncertainty separately within a single forward pass. However, the methods in Section \ref{sec:deterministic_methods} quantify the model and data uncertainty, while these calibration methods are regularized in order to minimize the model uncertainty. Following, at inference, the model uncertainty cannot be obtained anymore. This is the main motivation for us to separate the approaches presented below from the approaches presented in Section \ref{sec:deterministic_methods}. \\
One popular regularization based calibration method is label smoothing \cite{szegedy2016rethinking}. For label smoothing, the labels of the training examples are modified by taking a small portion $\alpha$ of the true class' probability mass and assign it uniformly to the false classes. For hard, non-smoothed labels, the optimum cannot be reached in practice, as the gradient of the output with respect to the logit vector $z$,
\begin{align}
    \begin{split}
    \nabla_z \text{CE}(y, \hat y(z)) &= \text{softmax}(z) - y \\
    &= \frac{\exp(z)}{\sum_{i=1}^K \exp(z_i)}-y~,
    \end{split}
\end{align}
can only converge to zero with increasing distance between the true and false classes' logits. As a result, the logits of the correct class are much larger than the logits for the incorrect classes and the logits of the incorrect classes can be very different to each other. Label-smoothing avoids this and while it generally leads to a higher training loss, the calibration error decreases and the accuracy often increases as well \cite{muller2019does}. \\
Seo et al. \cite{seo2019learning} extended the idea of label smoothing and directly aimed at reducing the model uncertainty. For this, they sampled $T$ forward passes of a stochastic neural network already at training time. Based on the $T$ forward passes of a training sample $(x_i,y_i)$, a normalized model variance $\alpha_i$ is derived as the mean of the Bhattacharyya coefficients \cite{comaniciu2000real} between the $T$ individual predictions $\hat y_1,...,\hat y_T$ and the average prediction $\bar y = \frac{1}{T}\sum_{t=1}^T\hat y_t$,
\begin{align}
    \begin{split}
        \alpha_i &= \frac{1}{T}\sum_{t=1}^T BC(\bar y_i, \hat y_{i,t})  \\
        &=\frac{1}{T}\sum_{t=1}^T \sum_{k=1}^K \sqrt{\bar y_{i,k} \cdot \hat y_{i,t,k}}~.
    \end{split}
\end{align}
Based on this $\alpha_i$, Seo et al. \cite{seo2019learning} introduced the variance-weighted confidence-integrated loss function that is a convex combination of two contradictive loss functions,
\begin{align}
    \begin{split}
        L^{\text{VWCI}}(\theta)=-\sum_{i=1}^N(1-\alpha_i)L_{\text{GT}}^{(i)}(\theta) + \alpha_i L_{\text{U}}^{(i)}(\theta)~,
    \end{split}
\end{align}
where $L_\text{GT}^{(i)}$ is the mean cross-entropy computed for the training sample $x_i$ with given ground-truth $y_i$. $L_\text{U}$ represents the mean KL-divergence between a uniform target probability vector and the computed prediction. The adaptive smoothing parameter ${\alpha}_i$ pushes predictions of training samples with high model uncertainty (given by high variances) towards a uniform distribution while increasing the prediction scores of samples with low model uncertainty. As a result, variances in the predictions of a single sample are reduced and the network can then be applied with a single forward pass at inference.

Pereyra et al. \cite{pereyra2017regularizing} combated the overconfidence issue by adding the negative entropy to the standard loss function and therefore a penalty that increases with the network's predicted confidence. This results in the entropy-based objective function $L^H$, which is defined as
\begin{equation}
L^H(\theta) = -\frac{1}{N} \sum_{i=1}^{N} y_i \log \hat{y}_i - \alpha_i H(\hat{y}_i)~,
\end{equation}
where $H(\hat{y}_i)$ is the entropy of the output and $\alpha_i$ a parameter that controls the strength of the entropy-based confidence penalty. The parameter $\alpha_i$ is computed equivalently as for the VWCI loss. 

Instead of regularizing the training process by modifying the objective function, Thulasidasan et al. \cite{thulasidasan2019mixup} regularized it by using a data-agnostic data augmentation technique named mixup \cite{zhang2017mixup}. In mixup training, the network is not only trained on the training data, but also on virtual training samples $(\tilde x, \tilde y)$ generated by a convex combination of two random training pairs $(x_i,y_i)$ and $(x_j,y_j)$, i.e. 
\begin{equation}
\tilde{x} = \lambda x_i + (1 - \lambda) x_j
\end{equation}
\begin{equation}
\tilde{y} = \lambda y_i + (1 - \lambda) y_j~.
\end{equation}
According to \cite{thulasidasan2019mixup}, the label smoothing resulting from mixup training can be viewed as a form of entropy-based regularization resulting in inherent calibration of networks trained with mixup. Maro$\tilde{\text{n}}$as et al. \cite{maronas2020improving} see mixup training among the most popular data augmentation regularization techniques due to its ability to improve the calibration as well as the accuracy. However, they argued that in mixup training the data uncertainty in mixed inputs affects the calibration and therefore mixup does not necessarily improve the calibration. They also underlined this claim empirically. Similarly, Rahaman and Thiery \cite{rahaman2020uncertainty} experimentally showed that the distributional-shift induced by data augmentation techniques such as mixup training can negatively affect the confidence calibration. Based on this observation, Maro$\tilde{\text{n}}$as et al. \cite{maronas2020improving} proposed a new objective function that explicitly takes the calibration performance on the unmixed input samples into account. Inspired by the expected calibration error (ECE, see Section \ref{sec:calibration_quality}) Naeini et al. \cite{naeini2015obtaining} measured the calibration performance on the unmixed samples for each batch $b$ by the differentiable squared differences between the batch accuracy and the mean confidence on the batch samples. The total loss is given as a weighted combination of the original loss on mixed and unmixed samples and the calibration measure evaluated only on the unmixed samples: 
\begin{equation}
L^{ECE}(\theta) = \frac{1}{B} \sum_{b \in B} L^b(\theta) + \beta ECE_b~,
\end{equation}
where $L^b(\theta)$ is the original unregularized loss using training and mixed samples included in batch $b$ and $\beta$ is a hyperparameter controlling the relative importance given to the batchwise expected calibration error $ECE_b$. By adding the batchwise calibration error for each batch $b \in B$ to the standard loss function, the miscalibration induced by mixup training is regularized. 

In the context of data augmentation, Patel et al. \cite{patel2019manifold} improved the calibration of uncertainty estimates by using on-manifold data augmentation. While mixup training combines training samples, on-manifold adversarial training generate out-of-domain samples using adversarial attack. They experimentally showed that on-manifold adversarial training outperforms mixup training in improving the calibration. Similar to \cite{patel2019manifold}, Hendrycks et al. \cite{hendrycks2018deep} showed that exposing classifiers to out-of-distribution examples at training can help to improve the calibration.

\subsubsection{Post-Processing Methods}
Post-processing (or post-hoc) methods are applied after the training process and aim at learning a re-calibration function. For this, a subset of the training data is held-out during the training process and used as a calibration set. The re-calibration function is applied to the network's outputs (e.g. the logit vector) and yields an improved calibration learned on the left-out calibration set. Zhang et al. \cite{zhang2020mix} discussed three requirements that should be satisfied by post-hoc calibration methods. They should 
\begin{enumerate}
    \item preserve the accuracy, i.e. should not affect the predictors performance.
    \item be data efficient, i.e. only a small fraction of the training data set should be left out for the calibration.
    \item be able to approximate the correct re-calibration map as long as there is enough data available for calibration. 
\end{enumerate}
Furthermore, they pointed out that none of the existing approaches fulfills all three requirements. \\
For classification tasks, the most basic but still very efficient way of post-hoc calibration is temperature scaling \cite{guo2017calibration}. For temperature scaling, the temperature $T>0$ of the softmax function
\begin{equation}\label{eq:temp_scaling}
    \text{softmax}(z_i) = \frac{\exp^{z_i/T}}{\sum_{j=1}^K\exp^{z_j/T}}~,
\end{equation}
is optimized. For $T=1$ the function remains the regular softmax function. For $T>1$ the output changes such that its entropy increases, i.e. the predicted confidence decreases. For $T\in(0,1)$ the entropy decreases and following, the predicted confidence increases. As already mentioned above, a perfect calibrated neural network outputs MAP estimates. Since the learned transformation can only affect the uncertainty, the log-likelihood based losses as cross-entropy do not have to be replaced by a special calibration loss. While the data efficiency and the preservation of the accuracy is given, the expressiveness of basic temperature scaling is limited \cite{zhang2020mix}. To overcome this, Zhang et al. \cite{zhang2020mix} investigated an ensemble of several temperature scaling models. Doing so, they achieved better calibrated predictions, while preserving the classification accuracy and improving the data efficiency and the expressive power. 
Kull et al. \cite{kullbeyond} were motivated by non-neural network calibration methods, where the calibration is performed class-wise as a one-vs-all binary calibration. They showed that this approach can be interpreted as learning a linear transformation of the predicted log-likelihoods followed by a softmax function. This again is equivalent to train a dense layer on the log-probabilities and hence the method is also very easy to implement and apply. Obviously, the original predictions are not guaranteed to be preserved. 

Analogous to temperature scaling for classification networks, Levi et al. \cite{levi2019evaluating} introduced standard deviation scaling (std-scaling) for regression networks. As the name indicates, the method is trained to rescale the predicted standard deviations of a given network. Equivalently to the motivation of optimizing temperature scaling with the cross-entropy loss, std-scaling can be trained using the Gaussian log-likelihood function as loss, which is in general also used for the training of regression networks, which also give a prediction for the data uncertainty.

Wenger et al. \cite{wenger2020non} proposed a Gaussian process (GP) based method, which can be used to calibrate any multi-class classifier that outputs confidence values and presented their methodology by calibrating neural networks. The main idea behind their work is to learn the calibration map by a Gaussian process that is trained on the networks confidence predictions and the corresponding ground-truths in the left out calibration set. For this approach, the preservation of the predictions is also not assured. 

\subsubsection{Calibration with Uncertainty Estimation Approaches}
As already discussed above, removing the model uncertainty and receiving an accurate estimation of the data uncertainty leads to a well calibrated predictor. Following several works based on deep ensembles \cite{deep.ensembles,mehrtash2020confidence} and BNNs,  \cite{izmailov2019subspace, foong2019between,kristiadi2020being} also compared their performance to other methods based on the resulting calibration. Lakshminarayanan et al. \cite{deep.ensembles} and Mehrtash et al. \cite{mehrtash2020confidence} reported an improved calibration by applying deep ensembles compared to single networks. However, Rahaman and Thiery\cite{rahaman2020uncertainty} showed that for specific configurations as the usage of mixup-regularization, deep ensembles can even increase the calibration error. On the other side they showed that applying temperature scaling on the averaged predictions can give a significant improvement on the calibration. \\
For the Bayesian approaches, \cite{kristiadi2020being} showed that restricting the Bayesian approximation to the weights of the last fully connected layer of a DNN is already enough to improve the calibration significantly. Zhang et al. \cite{zhang2019confidence} and Laves et al. \cite{laves2019well} showed that confidence estimates computed with MC dropout can be poorly calibrated. To overcome this, Zhang et al. \cite{zhang2019confidence} proposed structured dropout, which consists of dropping channel, blocks or layers, to promote model diversity and reduce calibration errors. \\

\subsection{Evaluating Calibration Quality}\label{sec:calibration_quality}
Evaluating calibration consists of measuring the statistical consistency between the predictive distributions and the observations \cite{vaicenavicius2019evaluating}. 
For classification tasks, several calibration measures are based on binning. For that, the predictions are ordered by the predicted confidence $\hat p_i$ and grouped into $M$ bins $b_1,...,b_M$. Following, the calibration of the single bins is evaluated by setting the average bin confidence 
\begin{equation}\label{eq:conf}
    \text{conf}(b_m)=\frac{1}{\vert b_m \vert} \sum_{s\in b_m}\hat{p}_s
\end{equation}
 in relation to the average bin accuracy 
 \begin{equation}\label{eq:acc}
    \text{acc}(b_m) = \frac{1}{\vert b_m \vert} \sum_{s \in b_m} \mathbbm{1}(\hat{y}_s=y_s)~,
\end{equation}
where  $\hat{y}_s$, $y_s$ and $\hat{p}_s$ refer to the predicted and true class label of a sample $s$. As noted in \cite{guo2017calibration}, confidences are well-calibrated when for each bin $\text{acc}(b_m)=\text{conf}(b_m)$.
For a visual evaluation of a model's calibration, the reliability diagram introduced by \cite{degroot1983comparison} is widely used. For a reliability diagram, the $\text{conf}(b_m)$ is plotted against $\text{acc}(b_m)$. For a well-calibrated model, the plot should be close to the diagonal, as visualized in Figure \ref{fig:reliability_diagram}. The basic reliability diagram visualization does not distinguish between different classes. In order to do so and hence to improve the interpretability of the calibration error, Vaicenavicius et al. \cite{vaicenavicius2019evaluating} used an alternative visualization named multidimensional reliability diagram. 

For a quantitative evaluation of a model's calibration, different calibration measures can be considered. \\
The \textit{Expected Calibration Error} (ECE) is a widely used binning based calibration measure \cite{naeini2015obtaining,guo2017calibration, laves2019well, mehrtash2020confidence, thulasidasan2019mixup, wenger2020non}. For the ECE, $M$ equally-spaced bins $b_1,...,b_M$ are considered, where $b_m$ denotes the set of indices of samples whose confidences fall into the interval $I_m=]\frac{m-1}{M},\frac{m}{M}]$. The ECE is then computed as the weighted average of the bin-wise calibration errors, i.e.
\begin{equation}\label{eq:ece}
    \text{ECE} = \sum_{m=1}^{M}\frac{\vert b_m \vert}{N}\vert \text{acc}(b_m)-\text{conf}(b_m)\vert~.
\end{equation}

For the ECE, only the predicted confidence score (top-label) is considered. In contrast to this, the \textit{Static Calibration Error} (SCE) \cite{nixon2019measuring, ghandeharioun2019characterizing} considers the predictions of all classes (all-labels). For each class, the SCE computes the calibration error within the bins and then averages across all the bins, i.e.
\begin{equation}\label{eq:sce}
   \text{SCE} = \frac{1}{K} \sum_{k=1}^{K} \sum_{m=1}^{M} \frac{\vert b_{m_k} \vert}{N} \vert \text{conf}(b_{m_k})-\text{acc}(b_{m_k}) \vert~.
\end{equation}
Here $conf(b_{m_k})$ and $acc(b_{m_k})$ are the confidence and accuracy of bin $b_m$ for class label $k$, respectively. Nixon et al. \cite{nixon2019measuring} empirically showed that all-labels calibration measures such as the SCE are more effective in assessing the calibration error than the top-label calibration measures as the ECE.

In contrast to the ECE and SCE, which group predictions into $M$ equally-spaced bins (what in general leads to different numbers of evaluation samples per bin), the adaptive calibration error \cite{nixon2019measuring,ghandeharioun2019characterizing} adaptively groups predictions into $R$ bins with different width but equal number of predictions. With this adaptive bin size, the \textit{adaptive Expected Calibration Error} (aECE)
\begin{equation}\label{eq:a_ece}
    \text{aECE} = \frac{1}{R}\sum_{r=1}^{R} \vert \text{conf}(b_r) - \text{acc}(b_r) \vert~,
\end{equation}
and the \textit{adaptive Static Calibration Error} (aSCE)
\begin{equation}\label{eq:a_sce}
    \text{aSCE} = \frac{1}{K R} \sum_{k=1}^{K} \sum_{r=1}^{R} \vert \text{conf}(b_{r_k})-\text{acc}(b_{r_k}) \vert
\end{equation}
are defined as extensions of the ECE and the SCE.\\
As has been empirically shown in \cite{patel2019manifold} and \cite{nixon2019measuring}, the adaptive binning calibration measures $\text{aECE}$ and $\text{aSCE}$ are more robust to the number of bins than the corresponding equal-width binning calibration measures $\text{ECE}$ and $\text{SCE}$.

It is important to make clear that in a multi-class setting, the calibration measures can suffer from imbalance in the test data. Even when then calibration is computed classwise, the computed errors are weighted by the number of samples in the classes. Following, larger classes can shadow the bad calibration on small classes, comparable to accuracy values in classification tasks \cite{pulgar2017impact}.

\section{Data Sets and Baselines}\label{sec:data_sets_and_baselines} 
\begin{table*}[ht!]
	\centering
	\caption{Overview of frequently compared benchmark approaches, tasks and their data sets among existing works organized according to the taxonomy of this paper.}
	
	\begin{tabular}{>{\raggedright\arraybackslash}p{3.1cm}p{4.5cm}p{3.5cm}p{4cm}}
		
		& \textbf{Tasks}
		& \textbf{Task index: Data sets}
		& \textbf{Baselines}
		\\ \hline
		
		\addlinespace[2ex]
		\textbf{Bayesian Neural Networks}  
		& 1. Regression\textsuperscript{1,3,4,7,8,11,12,15-17}
		\newline 2. Calibration\textsuperscript{6,10,13,14}
		\newline 3. OOD Detection\textsuperscript{4,6,8,10,12,13}
		\newline 4. Adversarial Attacks\textsuperscript{4,8,12}
		\newline 5. Active Learning\textsuperscript{7,12,14} 
		\newline 6. Continual Learning\textsuperscript{10}
		\newline 7. Reinforcement Learning (Intrinsic Motivation, Contextual Bandits)\textsuperscript{1,14,15}
		& 1: UCI
		\newline~~2,~~3,~~4:~~(not)MNIST,
		\newline~~CIFAR10/100,~~SHVN,
		\newline~~ImageNet
		\newline 5: UCI
		\newline 6: Permuted MNIST
		& Softmax\textsuperscript{44},
		\newline MCdropout\textsuperscript{\hyperlink{gal2016dropout}{1}}, ~~DeepEnsemble\textsuperscript{\hyperlink{2}{2}},
		\newline BBB\textsuperscript{3},~~NormalizingFlow\textsuperscript{4},
		\newline PBP\textsuperscript{7},~~SWAG\textsuperscript{6},
		\newline KFAC\textsuperscript{8},~~DVI\textsuperscript{11}
		\newline HMC\textsuperscript{9}, ~~
		\newline VOGN\textsuperscript{10},~~INF\textsuperscript{12}
		\\ 
		
		\addlinespace[2ex]
		\multicolumn{4}{p\linewidth}{
					\textsuperscript{\hypertarget{gal2016dropout}{1}}\cite{gal2016dropout}~
					\textsuperscript{3}\cite{blundell2015weight}~
					\textsuperscript{4}\cite{louizos2017multiplicative}~
					\textsuperscript{5}\cite{graves2011practical}~
					\textsuperscript{6}\cite{maddox2019simple}~
					\textsuperscript{7}\cite{hernandez2015probabilistic}~
					\textsuperscript{8}\cite{ritter2018scalable}~
					\textsuperscript{9}\cite{neal1995bayesian}~
					\textsuperscript{10}\cite{osawa2019practical}~
					\textsuperscript{11}\cite{wu2018deterministic}~
					\textsuperscript{12}\cite{lee2020estimating}~
					\textsuperscript{13}\cite{wen2020combining}~
					\textsuperscript{14}\cite{zhang2018noisy}~
					\textsuperscript{15}\cite{sun2018functional}~
					\textsuperscript{16}\cite{sun2017learning}~
					\textsuperscript{17}\cite{izmailov2019subspace}~
				}
				\\
		\addlinespace[2ex]
		\hline
		
		\addlinespace[2ex]
		\textbf{Ensembles}
		& 1. Regression\textsuperscript{2,20}
		\newline 2. Calibration\textsuperscript{2,20,24-32}
		\newline 3. OOD Detection\textsuperscript{2,20,25,27-30,32-34}
		\newline 4. Active Learning\textsuperscript{31}
		& 1: Toy\textsuperscript{7}, UCI
		\newline 2, 3: Toy, (not)MNIST, SVHN, LSUN, CIFAR10/100,
		\newline (Tiny)ImageNet, Diabetic~~Retinopathy
		\newline 4: MNIST
		& Softmax\textsuperscript{44},~~MFVI\textsuperscript{5},~~SGLD\textsuperscript{55},
		\newline MCdropout\textsuperscript{\hyperlink{1}{1}},~~DeepEnsemble\textsuperscript{2},
		\newline BBB\textsuperscript{3},~~PBP\textsuperscript{7},~~NormalizingFlow\textsuperscript{4},
		\newline TemperatureScaling\textsuperscript{38,54}
		
		\\ 
		\addlinespace[2ex]
		\multicolumn{4}{p\linewidth}{
			\textsuperscript{\hypertarget{2}{2}}\cite{deep.ensembles}~
			\textsuperscript{20}\cite{distribution.distillation.general.framework}~
			\textsuperscript{24}\cite{rahaman2020uncertainty}~
			\textsuperscript{25}\cite{achrack2020multi}~
			\textsuperscript{26}\cite{huang2017snapshot}~
			\textsuperscript{27}\cite{malinin2019reverse}~
			\textsuperscript{28}\cite{sub.ensembles}~
			\textsuperscript{29}\cite{wen2020combining}~
			\textsuperscript{30}\cite{batch.ensembles}~
			\textsuperscript{31}\cite{beluch2018power}~
			\textsuperscript{32}\cite{ovadia2019can}~
			\textsuperscript{33}\cite{vyas2018out}~
			\textsuperscript{34}\cite{englesson2019efficient}~
		}
		\\
		\addlinespace[2ex]
		\hline
		
		\addlinespace[2ex]
		\textbf{Single Deterministic Models}
		& 1. Regression\textsuperscript{21-23}
		\newline 2. Calibration\textsuperscript{21-23,39,40,41,43}
		\newline 3. OOD Detection\textsuperscript{21,22,38,39,41-53}
		\newline 4. Adversarial Attacks\textsuperscript{21,41,48}
		& 1. Toy\textsuperscript{7}, UCI, NYU Depth
		\newline~~2,~~3:~~(E/Fashion/not)MNIST,
		\newline~~Toy,~~CIFAR10/100,~~SVHN,
		\newline~~LSUN,~~(Tiny)ImageNet,~~IMDB,
		\newline~~Diabetic~~Retinopathy,~~Omniglot
		\newline 4: MNIST, CIFAR10, NYU Depth, Omniglot
		& Softmax\textsuperscript{44},~~GAN\textsuperscript{27},~~Dirichlet\textsuperscript{48},~~BBB\textsuperscript{3},
		\newline~~MCdropout\textsuperscript{\hyperlink{1}{1}},~~DeepEnsemble\textsuperscript{\hyperlink{2}{2},57},
		\newline~~Mahalanobis\textsuperscript{56},~~TemperatureScaling\textsuperscript{38,54}
		\newline~~NormalizingFlow\textsuperscript{4}
		\\
		\addlinespace[2ex]
		\multicolumn{4}{p\linewidth}{
			\textsuperscript{21}\cite{alex2019}~
			\textsuperscript{22}\cite{tagasovska2019single}~
			\textsuperscript{23}\cite{kawashima2020aleatoric}~
			\textsuperscript{38}\cite{liang2017enhancing}
			\textsuperscript{39}\cite{wu2018deterministic}
			\textsuperscript{40}\cite{tsiligkaridis2020failure}
			\textsuperscript{41}\cite{dirichlet.networks}
			\textsuperscript{42}\cite{hein2019relu}
			\textsuperscript{43}\cite{vasudevan2019towards}
			\textsuperscript{44}\cite{hendrycks2016baseline}
			\textsuperscript{45}\cite{nandy2020towards}
			\textsuperscript{46}\cite{inhibited.softmax}
			\textsuperscript{47}\cite{lee2020gradients}
			\textsuperscript{48}\cite{evidential.neural.networks}
			\textsuperscript{49}\cite{malinin2019reverse}
			\textsuperscript{50}\cite{kernel.network}
			\textsuperscript{51}\cite{density.estimation.in.representation.space}
			\textsuperscript{52}\cite{second.opinion.medical}
			\textsuperscript{53}\cite{oberdiek2018classification}
		}
		\\
		\addlinespace[2ex]
		\hline
		
		\addlinespace[2ex]
		\textbf{Test-Time Data Augmentation}
		& 1. Semantic Segmentation\textsuperscript{36, 37}
		\newline 2. Calibration\textsuperscript{35}
		\newline 3. OOD Detection\textsuperscript{35-37}
		& 1, 2, 3: Medical data, Diabetic Retinopathy
		& Softmax\textsuperscript{44},~~MCdropout\textsuperscript{\hyperlink{1}{1}}
		\\
		
		\addlinespace[2ex]
		\multicolumn{4}{p\linewidth}{
			\textsuperscript{35}\cite{Ayhan.2018}~
			\textsuperscript{36}\cite{wang2018automatic}~
			\textsuperscript{37}\cite{wang2019aleatoric}~
		}
		\\
		\addlinespace[2ex]
		\hline		
		\\
		\multicolumn{4}{p\linewidth}{
			\textsuperscript{54}\cite{guo2017calibration}
			\textsuperscript{55}\cite{welling2011bayesian}
			\textsuperscript{56}\cite{lee2018simple}
			\textsuperscript{57}\cite{pearce2018highquality}
		}
		
	\end{tabular}
	\label{tab:datasets_baselines}
\end{table*}

In this section, we collect commonly used tasks and data sets for evaluating uncertainty estimation among existing works. Besides, a variety of baseline approaches commonly used as comparison against the methods proposed by the researchers are also presented. By providing a review on the relevant information of these experiments, we hope that both researchers and practitioners can benefit from it. While the former can gain a basic understanding of recent benchmarks tasks, data sets and baselines so that they can design appropriate experiments to validate their ideas more efficiently, the latter might use the provided information to select more relevant approaches to start based on a concise overview on the tasks and data sets on which the approach has been validated.

In the following, we will introduce the data sets and baselines summarized in table \ref{tab:datasets_baselines} according to the taxonomy used throughout this review.

The structure of the table is designed to organize the main contents of this section concisely, hoping to provide a clear overview of the relevant works. We group the approaches of each category into one of four blocks and extract the most commonly used tasks, data sets and provided baselines for each column respectively. The corresponding literature is listed at the bottom of each block to facilitate lookup. Note that we focus on methodological comparison here, but not the choice of architecture for different methods which has an impact on performance as well. Due to the space limitation and visual density, we only show the most important elements (task, data set, baselines) ranked according to the frequency of use in the literature we have researched.

The main results are as follows. One of the most frequent tasks for evaluating uncertainty estimation methods are regression tasks, where samples close and far away from the training distribution are studied. Furthermore, the calibration of uncertainty estimates in the case of classification problems is very often investigated. 
Further noteworthy tasks are out-of-distribution (OOD) detection and robustness against adversarial attacks. In the medical domain, calibration of semantic segmentation results is the predominant use case.

The choice of data sets is mostly consistent among all reviewed works. For regression, toy data sets are employed for visualization of uncertainty intervals while the UCI data sets are studied in light of (negative) log-likelihood comparison. The most common data sets for calibration and OOD detection are MNIST, CIFAR10 and 100 as well as SVHN while ImageNet and its tiny variant are also studied frequently. These form distinct pairs when OOD detection is studied where models trained on CIFAR variants are evaluated on SVHN and visa versa while MNIST is paired with variants of itself like notMNIST and FashionMNIST. Classification data sets are also commonly distorted and corrupted to study the effects on calibration, blurring the line between OOD detection and adversarial attacks.

Finally, the most commonly used baselines by far are Monte Carlo (MC) Dropout and deep ensembles while the softmax output of deterministic models is almost always employed as a kind of surrogate baseline. It is interesting to note that inside each approach--BNNs, Ensembles, Single Deterministic Models and Input Augmentation--some baselines are preferred over others.
BNNs are most frequently compared against variational inference methods like Bayes' by Backprop (BBB) or Probabilistic Backpropagation (PBP) while for Single Deterministic Models it is more common to compare them against distance-based methods in the case of OOD detection.
Overall, BNN methods show a more diverse set of tasks considered while being less frequently evaluated on large data sets like ImageNet.

To further facilitate access for practitioners, we provide web-links to the authors' official implementations (marked by a star) of all common baselines as identified in the baselines column. Where no official implementation is provided, we instead link to the highest ranked implementations found on \href{https://github.com/}{GitHub} at the time of this survey. The list can be also found within \href{https://github.com/JakobCode/UncertaintyInNeuralNetworks\_Resources}{our GitHub repository on available implementations}\footnote{\href{https://github.com/JakobCode/UncertaintyInNeuralNetworks\_Resources}{https://github.com/JakobCode/UncertaintyInNeuralNetworks\_Resources}}. The relevant baselines are Softmax\textsuperscript{*} (\href{https://github.com/hendrycks/error-detection}{TensorFlow}, \href{https://github.com/hendrycks/outlier-exposure}{PyTorch}), MCdropout (\href{https://github.com/yaringal/DropoutUncertaintyExps}{TensorFlow\textsuperscript{*}}; PyTorch: \href{https://github.com/cpark321/uncertainty-deep-learning}{1}, \href{https://github.com/JavierAntoran/Bayesian-Neural-Networks}{2}), DeepEnsembles (TensorFlow: \href{https://github.com/vvanirudh/deep-ensembles-uncertainty}{1}, \href{https://github.com/Kyushik/Predictive-Uncertainty-Estimation-using-Deep-Ensemble}{2}, \href{https://github.com/axelbrando/Mixture-Density-Networks-for-distribution-and-uncertainty-estimation}{3}; PyTorch: \href{https://github.com/bayesgroup/pytorch-ensembles}{1}, \href{https://github.com/cpark321/uncertainty-deep-learning}{2}), BBB (PyTorch: \href{https://github.com/ThirstyScholar/bayes-by-backprop}{1}, \href{https://github.com/nitarshan/bayes-by-backprop}{2}, \href{https://github.com/cpark321/uncertainty-deep-learning}{3}, \href{https://github.com/JavierAntoran/Bayesian-Neural-Networks}{4}), NormalizingFlow (\href{https://github.com/AMLab-Amsterdam/MNF_VBNN}{TensorFlow}, \href{https://github.com/janosh/torch-mnf}{PyTorch}), \href{https://github.com/HIPS/Probabilistic-Backpropagation}{PBP}, SWAG (\href{https://github.com/wjmaddox/swa_gaussian}{1\textsuperscript{*}}, \href{https://github.com/bayesgroup/pytorch-ensembles}{2}), KFAC (PyTorch: \href{https://github.com/DLR-RM/curvature}{1}, \href{https://github.com/bayesgroup/pytorch-ensembles}{2}, \href{https://github.com/JavierAntoran/Bayesian-Neural-Networks}{3}; \href{https://github.com/tensorflow/kfac}{TensorFlow}), DVI (\href{https://github.com/Microsoft/deterministic-variational-inference}{TensorFlow\textsuperscript{*}}, \href{https://github.com/markovalexander/DVI}{PyTorch}), \href{https://github.com/JavierAntoran/Bayesian-Neural-Networks}{HMC}, \href{https://github.com/team-approx-bayes/dl-with-bayes}{VOGN\textsuperscript{*}}, \href{https://github.com/DLR-RM/curvature}{INF\textsuperscript{*}}, \href{https://github.com/ctallec/pyvarinf}{MFVI}, \href{https://github.com/JavierAntoran/Bayesian-Neural-Networks}{SGLD}, TemperatureScaling (\href{https://github.com/gpleiss/temperature_scaling}{1\textsuperscript{*}}, \href{https://github.com/facebookresearch/odin}{2}, \href{https://github.com/cpark321/uncertainty-deep-learning}{3}), \href{https://github.com/KaosEngineer/PriorNetworks}{GAN\textsuperscript{*}}, \href{https://github.com/dougbrion/pytorch-classification-uncertainty}{Dirichlet} and \href{https://github.com/pokaxpoka/deep_Mahalanobis_detector}{Mahalanobis\textsuperscript{*}}.

\section{Applications of Uncertainty Estimates}\label{sec:application_fields}
From a practical point of view, the main motivation for quantifying uncertainties in DNNs is to be able to classify the received predictions and to make more confident decisions. This section gives a brief overview and examples of the aforementioned motivations. In the first part, we discuss how uncertainty is used within active learning and reinforcement learning. Subsequently, we discuss the interest of the communities working on domain fields like medical image analysis, robotics, and earth observation. These application fields are used representatively for the large number of domains where the uncertainty quantification plays an important role. The challenges and concepts could (and should) be transferred to any application domain of interest. 

\subsubsection{Active Learning}
\begin{figure}
    \centering
    \includegraphics[width=0.38\textwidth]{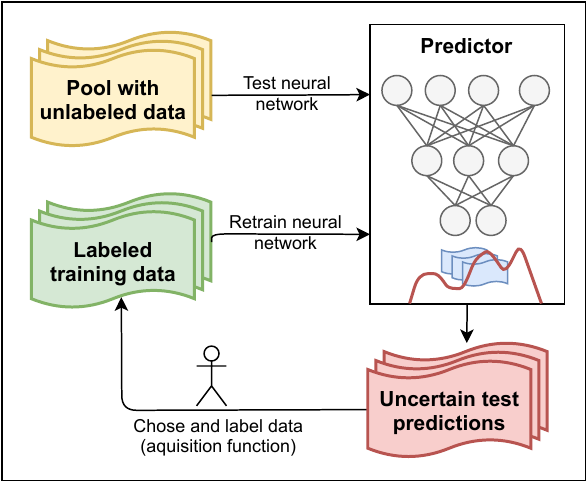}
    \caption{The active learning framework: The acquisition function evaluates the uncertainties on the network's test predictions in order to select unlabelled data. The selected data are labelled and added to the pool of labelled data, which is used to train and improve the performance of the predictor.}
    \label{fig:active_learning}
\end{figure}

The process of collecting labeled data for supervised training of a DNN can be laborious, time-consuming, and costly. To reduce the annotation effort, the active learning framework shown in Figure \ref{fig:active_learning} trains the DNN sequentially on different labelled data sets increasing in size over time \cite{iuzzolino2020automation}. In particular, given a small labelled data set and a large unlabeled data set, a deep neural network trained in the setting of active learning learns from the small labeled data set and decides based on the acquisition function, which samples to select from the pool of unlabeled data. The selected data are added to the training data set and a new DNN is trained on the updated training data set. This process is then repeated with the training set increasing in size over time. Uncertainty sampling is one most popular criteria used in acquisition functions \cite{settles2009active} where predictive uncertainty determines which training samples have the highest uncertainty and should be labelled next. Uncertainty based active learning strategies for deep learning applications have been successfully used in several works \cite{gal2017deep,chitta2018large,pop2018deep,zeng2018relevance,nguyen2019epistemic}. 

\subsubsection{Reinforcement Learning}
\begin{figure}
    \centering
    \includegraphics[width=0.38\textwidth]{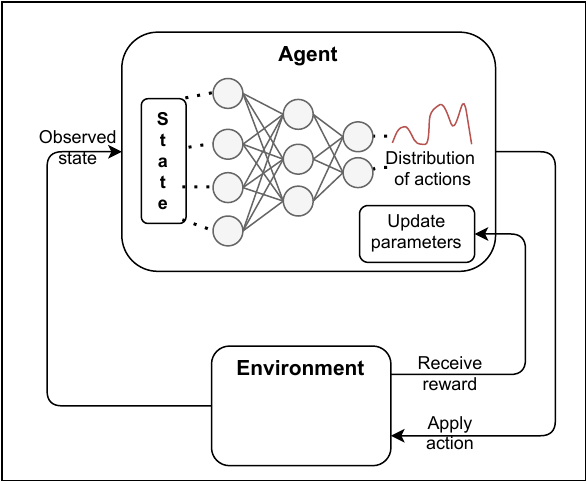}
    \caption{The reinforcement learning framework: The agent interacts with the environment by executing a specific action influencing the next state of the agent. The agent observes a reward representing the cost associated with the executed action. The agent chooses actions based on a policy learned by a deep neural network. However, the predicted uncertainty associated with the action predicted by the deep neural network can help the agent to decide weather to execute the predicted action or not.}
    \label{fig:reinforcement_learning}
\end{figure}

The general framework of deep reinforcement learning is shown in Figure \ref{fig:reinforcement_learning}. In the context of reinforcement learning, uncertainty estimates can be used to solve the exploration-exploitation dilemma. It says that uncertainty estimates can be used to effectively balance the exploration of unknown environments with the exploitation of existing knowledge extracted from known environments. For example, if a robot interacts with an unknown environment, the robot can safely avoid catastrophic failures by reasoning about its uncertainty. To estimate the uncertainty in this framework, Huang et al. \cite{huang2019bootstrap} used an ensemble of bootstrapped models (models trained on different data sets sampled with replacement from the original data set), while Gal and Ghahramani \cite{gal2016dropout} approximated Bayesian inference via dropout sampling. Inspired by \cite{gal2016dropout} and \cite{huang2019bootstrap}, Kahn et al. \cite{kahn2017uncertainty} and Lötjens et al. \cite{lotjens2019safe} used a mixture of deep Bayesian networks performing dropout sampling on an ensemble of bootstrapped models. For further reading, Ghavamzadeh et al. \cite{ghavamzadeh2016bayesian} presented a survey of Bayesian reinforcement learning. 

\subsection{Uncertainty in Real-World Applications}
With increasing usage of deep learning approaches within many different fields, quantifying and handling uncertainties has become more and more important. On one hand, uncertainty quantification plays an important role in risk minimization, which is needed in many application fields. On the other hand, many fields offer only challenging data sources, which are hard to control and verify. This makes the generation of trust-worthy ground truth a very challenging task. In the following, three different fields where uncertainty plays an important role are presented, namely Autonomous Driving, medical image analysis, and earth observation.

\subsubsection{Medical Analysis}
Since the size, shape, and location of many diseases vary largely across patients, the estimation of the predictive uncertainty is crucial in analyzing medical images in applications such as lesion detection \cite{nair2020exploring,seebock2019exploiting}, lung node segmentation \cite{hu2019supervised}, brain tumor segmentation \cite{eaton2018towards,wang2018automatic,wang2019aleatoric,roy2019bayesian,mcclure2019knowing}, parasite segmentation in images of liver stage malaria \cite{soleimany2019image}, recognition of abnormalities on chest radiographs \cite{ghesu2019quantifying}, and age estimation \cite{eggenreich2020variational}. Here, uncertainty estimates in particular improve the interpretability of decisions of DNNs \cite{ayhan2020expert}. They are essential to understand the reliability of segmentation results, to detect false segmented areas and to guide human experts in the task of refinement \cite{wang2019aleatoric}. Well-calibrated and reliable uncertainty estimates allow clinical experts to properly judge whether an automated diagnosis can be trusted \cite{ayhan2020expert}. Uncertainty was estimated in medical image segmentation based on Monte Carlo dropout \cite{eaton2018towards,hu2019supervised,nair2020exploring,roy2019bayesian,seebock2019exploiting,soberanis2020uncertainty,labonte2019we,reinhold2020validating,eggenreich2020variational}, spike-and-slab dropout \cite{mcclure2019knowing}, and spatial dropout \cite{soleimany2019image}. Wang et al. \cite{wang2018automatic,wang2019aleatoric} used test time data augmentation to estimate the data-dependent uncertainty in medical image segmentation.

\subsubsection{Robotics}
Robots are active agents that perceive, decide, plan, and act in the real-world – all based on their incomplete knowledge about the world. As a result, mistakes of the robots not only cause failures of their own mission, but can endanger human lives, e.g. in case of surgical robotics, self-driving cars, space robotics, etc. Hence, the robotics application of deep learning poses unique research challenges that significantly differ from those often addressed in computer vision and other off-line settings \cite{sunderhauf2018limits}. For example, the assumption that the testing condition come from the same distribution as training is often invalid in many settings of robotics, resulting in deterioration of the performance of DNNs in uncontrolled and detrimental conditions. This raises the questions how we can quantify the uncertainty in a DNN’s predictions in order to avoid catastrophic failures. Answering such questions are important in robotics, as it might be a lofty goal to expect data-driven approaches (in many aspects from control to perception) to always be accurate. Instead, reasoning about uncertainty can help in leveraging the recent advances in deep learning for robotics. 

Reasoning about uncertainties and the use of probabilistic representations, as oppose to relying on a single, most-likely estimate, have been central to many domains of robotics research, even before the advent of deep learning \cite{thrun2002probabilistic}. In robot perception, several uncertainty-aware methods have been proposed in the past, starting from localization methods \cite{fox1998markov, fox2000probabilistic, thrun2001robust} to simultaneous localization and mapping (SLAM) frameworks \cite{durrant2006simultaneous, bailey2006simultaneous, montemerlo2002fastslam, kaess2010bayes}. As a result, many probabilistic methods such as factor graphs \cite{dellaert2017factor, loeliger2004introduction} are now the work-horse of advanced consumer products such as robotic vacuum cleaners and unmanned aerial vehicles. In case of planning and control, estimation problems are widely treated as Bayesian sequential learning problems, and sequential decision making frameworks such as POMDPs \cite{silver2010monte, ross2008online} assume a probabilistic treatment of the underlying planning problems. With probabilistic representations, many reinforcement learning algorithms are backed up by stability guarantees for safe interactions in the real-world \cite{richards2018lyapunov, berkenkamp2016safe, berkenkamp2017safe}. Lastly, there have been also several advances starting from reasoning (semantics \cite{grimmett2016introspective} to joint reasoning with geometry), embodiment (e.g. active perception \cite{bajcsy1988active}) to learning (e.g. active learning \cite{triebel2016driven, narr2016stream, cohn1996active} and identifying unknown objects \cite{nguyen2015deep, wong2020identifying, boerdijk2020s}). 

Similarly, with the advent of deep learning, many researchers proposed new methods to quantify the uncertainty in deep learning as well as on how to further exploit such information. As oppose to many generic approaches, we summarize task-specific methods and their application in practice as followings. Notably, \cite{richter2017safe} proposed to perform novelty detection using auto-encoders, where the reconstructed outputs of auto-encoders was used to decide how much one can trust the network’s predictions. Peretroukhin et al. \cite{peretroukhin2020smooth} developed a SO(3) representation and uncertainty estimation framework for the problem of rotational learning problems with uncertainty. \cite{lutjens2019safe, kahn2017uncertainty, kahn2018self, stulp2011learning} demonstrated uncertainty-aware, real world application of a reinforcement learning algorithm for robotics, while \cite{tchuiev2018inference, feldman2018bayesian} proposed to leverage spatial information, on top of MC-dropout. \cite{shinde2020learning, yang2020d3vo, wang2017deepvo} developed deep learning based localization systems along with uncertainty estimates. Other approaches also learn on the robots' past experiences of failures or detect inconsistencies of the predictors \cite{guruau2016fit, daftry2016introspective}. In summary, the robotics community has been both, the users and the developers of the uncertainty estimation frameworks targeted to a specific problems. 

Yet, robotics pose several unique challenges to uncertainty estimation methods for DNNs. These are for example, (i) how to limit the computational burden and build real-time capable methods that can be executed on the robots with limited computational capacities (e.g. aerial, space robots, etc); (ii) how to leverage spatial and temporal information, as robots sense sequentially instead of having a batch of training data for uncertainty estimates; (iii) whether robots can select the most uncertainty samples and update its learner online; (iv) Whether robots can purposefully manipulate the scene when uncertain. Most of these challenges arise due to the properties of robots that they are physically situated systems. 

\subsubsection{Earth Observation(EO)}
\begin{figure*}[t]
\resizebox{\textwidth}{!}{
   \includegraphics{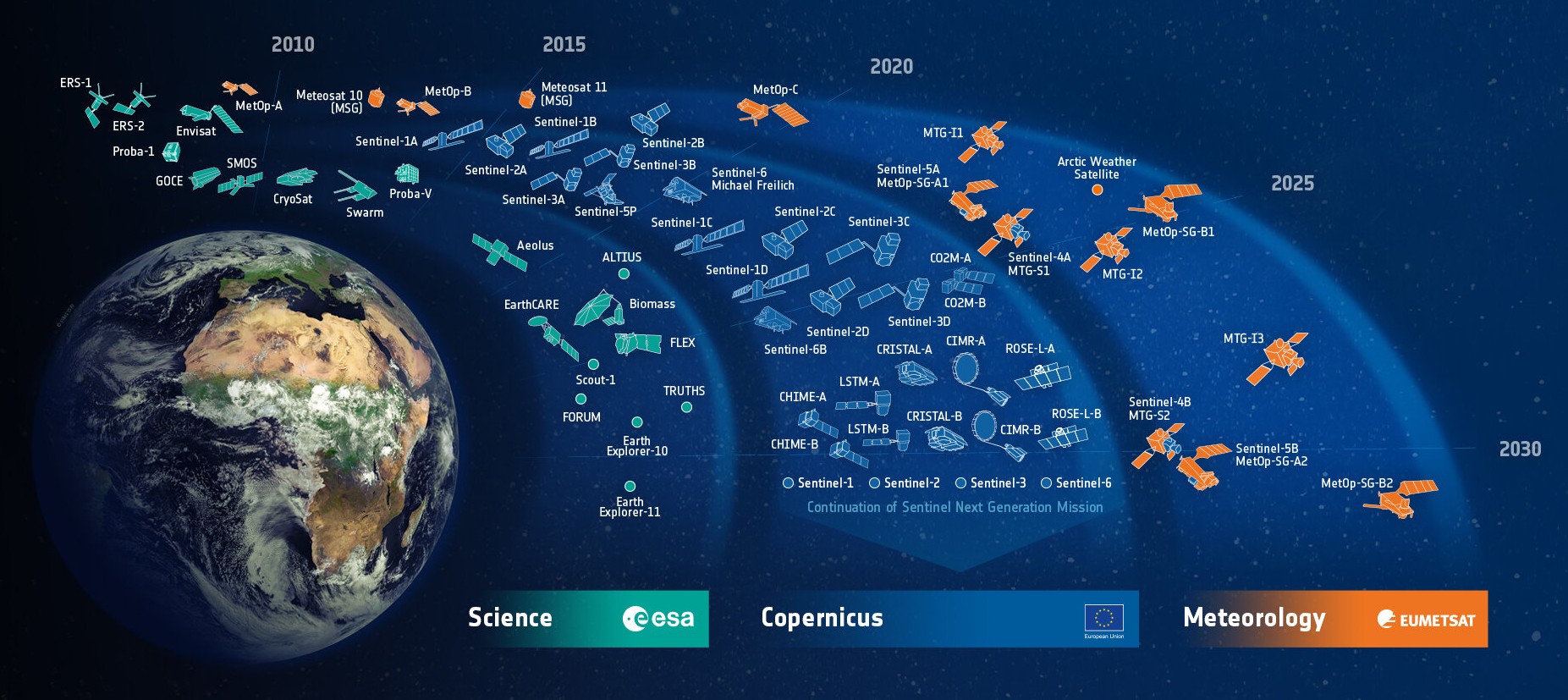}}
    \caption{European Space Agency (ESA) Developed Earth Observation Missions \cite{esa_satellites}.}
    \label{fig:ESA} 
\end{figure*}

Earth Observation (EO) systems are increasingly used to make critical decisions related to urban planning \cite{netzband2007applied}, resource management \cite{giardino2010application}, disaster response \cite{van2000remote}, and many more. Right now, there are hundreds of EO satellites in space, owned by different space agencies and private companies. Figure \ref{fig:ESA} shows the satellites owned by the European Space Agency (ESA). Like in many other domains, deep learning has shown great initial success in the field of EO over the past few years \cite{zhu2017deep}. These early successes consisted of taking the latest developments of deep learning in computer vision and applying them to small curated earth observation data sets \cite{zhu2017deep}. At the same time, the underlying data is very challenging. Even though the amount of data is huge, so is the variability in the data. This variability is caused by different sensor types, spatial changes (e.g. different regions and resolutions), and temporal changes (e.g. changing light conditions, weather conditions, seasons). Besides the challenge of efficient uncertainty quantification methods for such large amounts of data, several other challenges that can be tackled with uncertainty quantification exist in the field of EO. All in all, the sensitivity of many EO applications together with the nature of EO systems and the challenging EO data make the quantification of uncertainties very important in this field. Despite hundreds of publications in the last years on DL for EO, the range of literature on measuring uncertainties of these systems is relatively small.

Furthermore, due to the large variation in the data, a data sample received at test time is often not covered by the training data distribution. For example while preparing training data for a local climate zone classification, the human experts might be presented only with images where there is no obstruction and structures are clearly visible. When a model which is trained on this data set is deployed in real world, it might see the images with clouds obstructing the structures or snow giving them a completely different look. Also, the classes in EO data can have a very wide distribution. For example, there are millions of types of houses in the world and no training data can contain the examples for all of them. The question is where the OOD detector will draw the line and declare the following houses as OOD. Hence, OOD detection is important in earth observation and uncertainty measurements play an important part in this \cite{gawlikowski2021out}. 

Another common task in EO, where uncertainties can play an important role, is the data fusion. Optical images normally contain only a few channels like RGB. In contrast to this, EO data can contain optical images with up to hundreds of channels, and a variety of different sensors with different spatial, temporal, and semantic properties. Fusing the information from these different sources and channels propagates the uncertainties from different sources onto the prediction. The challenge lies in developing methods that do not only quantify uncertainties but also the amount of contribution from different channels individually and which learn to focus on the trustworthy data source for a given sample \cite{gawlikowski2020fusion}.

Unlike normal computer vision scenarios where the image acquisition equipment is quite near to the subject, the EO satellites are hundreds of kilometers away from the subject. The sensitivity of sensors, the atmospheric absorption properties, and surface reflectance properties all contribute to uncertainties in the acquired data. Integrating the knowledge of physical EO systems, which also contain information about uncertainty models in those systems, is another major open issue. However, for several applications in EO, measuring uncertainties is not only something good to have but rather an important requirement of the field. E.g., the geo-variables derived from EO data may be assimilated into process models (ocean, hydrological, weather, climate, etc) and the assimilation requires the probability distribution of the estimated variables.

\section{Conclusion and Outlook}
\label{sec:con}
\subsection{Conclusion - How well do the current uncertainty quantification methods work for real world applications?}\label{ssec:conclusion}
Even though many advances on uncertainty quantification in neural networks have been made over the last years, their adoption in practical mission- and safety-critical applications is still limited. There are several reasons for this, which are discussed one-by-one as follows:
\begin{itemize}
  \setlength\itemsep{0.5em}
    \item \textbf{Missing Validation of Existing Methods over Real-World Problems}~\\
    Although DNNs have become the defacto standard in solving numerous computer vision and medical image processing tasks, the majority of existing models are not able to appropriately quantify uncertainty that is inherent to their inferences particularly in real world applications. This is primarily because the baseline models are mostly developed using standard data sets such as Cifar10/100, ImageNet, or well known regression data sets that are specific to a particular use case and are therefore not readily applicable to complex real-world environments, as for example low resolutional satellite data or other data sources affected by noise. Although many researchers from other fields apply uncertainty quantification in their field \cite{marcmohsin2020uncertainty,loquercio2020general,choi2019gaussian}, a broad and structured evaluation of existing methods based on different real world applications is not available yet. Works like \cite{gustafsson2020evaluating} already built first steps towards a real life evaluation.
    \item \textbf{Lack of Standardized Evaluation Protocol}~\\
    Existing methods for evaluating the estimated uncertainty are better suited to compare uncertainty quantification methods based on measurable quantities such as the calibration \cite{nado2021uncertainty} or the performance on out-of-distribution detection \cite{prior.network}. As described in Section \ref{sec:data_sets_and_baselines}, these tests are performed on standardized sets within the machine learning community. Furthermore, the details of these experiments might differ in the experimental setting from paper to paper \cite{mukhoti2018importance}. However, a clear standardized protocol of tests that should be performed on uncertainty quantification methods is still not available. For researchers from other domains it is difficult to directly find state of the art methods for the field they are interested in, not to speak of the hard decision on which sub-field of uncertainty quantification to focus. This makes the direct comparison of the latest approaches difficult and also limits the acceptance and adoption of current existing methods for uncertainty quantification.
    \item \textbf{Inability to Evaluate Uncertainty Associated to a Single Decision}~\\ 
    Existing measures for evaluating the estimated uncertainty (e.g., the expected calibration error) are based on the whole testing data set. This means, that equivalent to classification tasks on unbalanced data sets, the uncertainty associated with single samples or small groups of samples may potentially get biased towards the performance on the rest of the data set. But for practical applications, assessing the reliability of a predicted confidence would give much more possibilities than an aggregated reliability based on some testing data, which are independent from the current situation \cite{kull2014reliability}. Especially for mission- and safety-critical applications, pointwise evaluation measures could be of paramount importance and hence such evaluation approaches are very desirable. 
    \item \textbf{Lack of Ground Truth Uncertainties}~\\
    Current methods are empirically evaluated and the performance is underlined by reasonable and explainable values of uncertainty. A ground truth uncertainty that could be used for validation is in general not available.
    Additionally, even though existing methods are calibrated on given data sets, one cannot simply transfer these results to any other data set since one has to be aware of shifts in the data distribution and that many fields can only cover a tiny portion of the actual data environment. In application fields as EO, the preparation of a huge amount of training data is hard and expensive and hence synthetic data can be used to train a model. For this artificial data, artificial uncertainties in labels and data should be taken into account to receive a better understanding of the uncertainty quantification performance. The gap between the real and synthetic data, or estimated and real uncertainty further limits the adoption of currently existing methods for uncertainty quantification.
    \item \textbf{Explainability Issue:} ~\\
    Existing methods of neural network uncertainty quantification deliver predictions of certainty without any clue about what causes possible uncertainties. Even though those certainty values often look \textit{reasonable} to a human observer, one does not know whether the uncertainties are actually predicted based on the same observations the human observer made. But without being sure about the reasons and motivations of single uncertainty estimations, a proper transfer from one data set to another, and even only a domain shift, are much harder to realize with a guaranteed performance. Regarding safety critical real life applications, the lack of explainability makes the application of the available methods significantly harder. Besides the explainability of neural networks decisions, existing methods for uncertainty quantification are not well understood on a higher level. For instance, explaining the behavior of single deterministic approaches, ensembles or Bayesian methods is a current direction of research and remains difficult to grasp in every detail \cite{fort2019deep}. It is, however, crucial to understand how those methods operate and capture uncertainty to identify pathways for refinement, detect and characterize uncertainty, failures and important shortcomings \cite{fort2019deep}.
\end{itemize}

\subsection{Outlook}

\begin{itemize}
    \item \textbf{Generic Evaluation Framework}\\
    As already discussed above, there are still problems regarding the evaluation of uncertainty methods, as the lack of 'ground truth' uncertainties, the inability to test on single instances, and standardized benchmarking protocols, etc. To cope with such issues, the provision of an evaluation protocol containing various concrete baseline data sets and evaluation metrics that cover all types of uncertainty would undoubtedly help to boost research in uncertainty quantification. Also, the evaluation with regard to risk-averse and worst case scenarios should be considered there. This means, that uncertainty predictions with a very high predicted uncertainty should never fail, as for example for a prediction of a red or green traffic light. Such a general protocol would enable researchers to easily compare different types of methods against an established benchmark as well as on real world data sets. The adoption of such a standard evaluation protocol should be encouraged by conferences and journals.
    \item \textbf{Expert \& Systematic Comparison of Baselines}~\\
    A broad and structured comparison of existing methods for uncertainty estimation on real world applications is not available yet. An evaluation on real world data is even not standard in current machine learning research papers. As a result, given a specific application, it remains unclear which method for uncertainty estimation performs best and whether the latest methods outperform older methods also on real world examples. This is also partly caused by the fact, that researchers from other domains that use uncertainty quantification methods, in general present successful applications of single approaches on a specific problem or a data set by hand. Considering this, there are several points that could be adopted for a better comparison within the different research domains. For instance, domain experts should also compare different approaches against each other and present the weaknesses of single approaches in this domain. Similarly, for a better comparison among several domains, a collection of all the works in the different real world domains could be collected and exchanged on a central platform. Such a platform might also help machine learning researchers in providing an additional source of challenges in the real world and would pave way to broadly highlight weaknesses in the current state of the art approaches. Google's repository on baselines in uncertainties in neural networks \cite{nado2021uncertainty}\footnote{\href{https://github.com/google/uncertainty-baselines}{https://github.com/google/uncertainty-baselines}} could be such a platform and a step towards achieving this goal. 
    
    \item \textbf{Uncertainty Ground Truths} \\
    It remains difficult to validate existing methods due to the lack of uncertainty ground truths. An actual uncertainty ground truth on which methods can be compared in an ImageNet like manner would make the evaluation of predictions on single samples possible. To reach this, the evaluation of the data generation process and occurring sources of uncertainty, as for example the labeling process, might be investigated in more detail.
    
    \item \textbf{Explainability and Physical Models} \\
    Knowing the actual reasons for a false high certainty or a low certainty makes it much easier to engineer the methods for real life applications, which again increases the trust of people into such methods. Recently, Antorán et al. \cite{antoran2020getting} claimed to have published the first work on explainable uncertainty estimation. Uncertainty estimations, in general, form an important step towards explainable artificial intelligence. Explainable uncertainty estimations would give an even deeper understanding of the decision process of a neural network, which, in practical deployment of DNNs, shall incorporate the desired ability to be risk averse while staying applicable in real world (especially safety critical applications). Also, the possibility of improving explainability with physically based arguments offers great potential. While DNNs are very flexible and efficient, they do not directly embed the domain specific expert knowledge that is mostly available and can often be described by mathematical or physical models, as for example earth system science problems \cite{reichstein2019deep}. Such physic guided models offer a variety of possibilities to include explicit knowledge as well as practical uncertainty representations into a deep learning framework \cite{willard2020integrating,de2019deep}. 
    \end{itemize}



\bibliography{main_bib}


\end{document}